\documentclass{article}

\usepackage[final]{neurips_2023}

\usepackage{hyperref}

\usepackage[utf8]{inputenc} % allow utf-8 input
\usepackage[T1]{fontenc}    % use 8-bit T1 fonts
\usepackage{hyperref}       % hyperlinks
\usepackage{url}            % simple URL typesetting
\usepackage{booktabs}       % professional-quality tables
\usepackage{amsfonts}       % blackboard math symbols
\usepackage{nicefrac}       % compact symbols for 1/2, etc.
\usepackage{microtype}      % microtypography
\usepackage{xcolor}         % colors
\usepackage{tabularx, booktabs}
\usepackage{longtable}
\usepackage{listings}

\newcolumntype{Y}{>{\centering\arraybackslash}X}
\newcolumntype{L}{>{\arraybackslash}X}

\usepackage{enumitem}
\usepackage{booktabs}  
\usepackage{graphicx}
\usepackage{caption}
\usepackage{subcaption}
\usepackage{wrapfig}
\usepackage{amssymb}
\usepackage{amsmath}
\usepackage{makecell}
\usepackage{multirow}
\usepackage{placeins}
\usepackage[normalem]{ulem}

\usepackage{wrapfig}
\usepackage{bbm}
\usepackage{amsthm}

\usepackage{lmodern,babel,adjustbox,booktabs,multirow}

\usepackage{amsmath,amsfonts,bm}

\def\eqref#1{equation~\ref{#1}}
\def\1{\bm{1}}

\def\vh{{\bm{h}}}

\def\vp{{\bm{p}}}

\def\vr{{\bm{r}}}

\def\vw{{\bm{w}}}
\def\vx{{\bm{x}}}
\def\vy{{\bm{y}}}
\def\vz{{\bm{z}}}

\def\mB{{\bm{B}}}

\def\mD{{\bm{D}}}

\DeclareMathAlphabet{\mathsfit}{\encodingdefault}{\sfdefault}{m}{sl}
\SetMathAlphabet{\mathsfit}{bold}{\encodingdefault}{\sfdefault}{bx}{n}

\usepackage[linesnumbered,ruled,vlined]{algorithm2e}

\usepackage[toc,page,header]{appendix}
\usepackage{minitoc}

\newcommand{\smallsection}[1]{\textbf{#1.~~~~}}

\newcommand{\ourmodel}{GRIN MoE\xspace}
\newcommand{\sm}{SparseMixer-v2\xspace}

\title{GRIN: \emph{GR}adient-\emph{IN}formed MoE}

\begin{document}

\doparttoc % Tell to minitoc to generate a toc for the parts
\faketableofcontents % Run a fake tableofcontents command for the partocs

\author{
{\normalfont Liyuan Liu\thanks{Correspondence to \{lucliu, jfgao, wzchen\}@microsoft.com} \qquad  Young Jin Kim \qquad  Shuohang Wang \qquad   Chen Liang}\\
\qquad Yelong Shen \qquad Hao Cheng \qquad  Xiaodong Liu \qquad  Masahiro Tanaka \\
Xiaoxia Wu \qquad Wenxiang Hu \qquad Vishrav Chaudhary \\
Zeqi Lin \qquad  Chenruidong Zhang \qquad Jilong Xue \\
Hany Awadalla \qquad Jianfeng Gao\footnotemark[1] \qquad  Weizhu Chen\footnotemark[1] \\
\rule[28pt]{0pt}{0pt}
\hspace{-6.5pt}{Microsoft}
}

\maketitle

\vspace{14pt}
\begin{abstract}
% Mixture-of-Experts (MoE) models are known to scale more effectively than dense models due to their sparse computation through expert routing, 
Mixture-of-Experts (MoE) models scale more effectively than dense models due to sparse computation through expert routing,
selectively activating only a small subset of expert modules.
However, sparse computation challenges traditional training practices, as discrete expert routing hinders standard backpropagation and thus gradient-based optimization, which are the cornerstone of deep learning.
To better pursue the scaling power of MoE, we introduce GRIN (\emph{GR}adient-\emph{IN}formed MoE training), which incorporates sparse gradient estimation for expert routing and configures model parallelism to avoid token dropping.
Applying GRIN to autoregressive language modeling, we develop a top-2 16$\times$3.8B MoE model.
Our model, with only 6.6B activated parameters, outperforms a 7B dense model and matches the performance of a 14B dense model trained on the same data. 
Extensive evaluations across diverse tasks demonstrate the potential of GRIN to significantly enhance MoE efficacy, achieving 79.4 on MMLU, 83.7 on HellaSwag, 74.4 on HumanEval, and 58.9 on MATH.

\end{abstract}

\section{Introduction}

The success of large-scale pre-training highlights the importance of model scalability~\citep{openai2023gpt4,touvron2023llama}. 
%In pursuit of model scaling, 
Mixture-of-Experts (MoE) models have emerged as a promising approach, selectively activating only a small subset of modules on specific inputs 
through an \emph{expert routing} process, thus improving the model scalability by orders of magnitude~\citep{Lepikhin2020GShardSG,fedus2022switch,zoph2022stmoedesigningstabletransferable}.

%Maintaining MoE models to be sparsely activated presents various 
However, the sparse activation mechanism of MoE presents several challenges to model training. 
For example, while the discrete routing function %, being discrete in nature, 
produces non-differentiable outputs, 
backpropagation, the cornerstone of deep learning, is exclusively compatible with differentiable functions~\citep{rosenblatt1957perceptron,Bengio2013EstimatingOP}.
Consequently, backpropagation cannot be directly applied for gradient computation of expert routing.

To fully leverage the scaling potential of MoE, we study gradient estimation for expert routing and configure model parallelism to avoid token dropping in this work. 
Extending \citet{Liu2023BridgingDA,Liu2023SparseMixer}, we propose \sm to estimate gradient for expert routing, which differs from conventional practices that use the gating gradient as a proxy for the routing gradient.
Additionally, we propose a scalable MoE training recipe that uses pipeline parallelism and tensor parallelism instead of expert parallelism, thus eliminating the needs of using a capacity factor and dropping tokens during training. 
%The training recipe is further adapted to accommodate such parallelism designs.

\begin{figure}
\begin{minipage}{\textwidth}
\begin{minipage}[b]{0.62\textwidth}
\centering
\includegraphics[width=\textwidth]{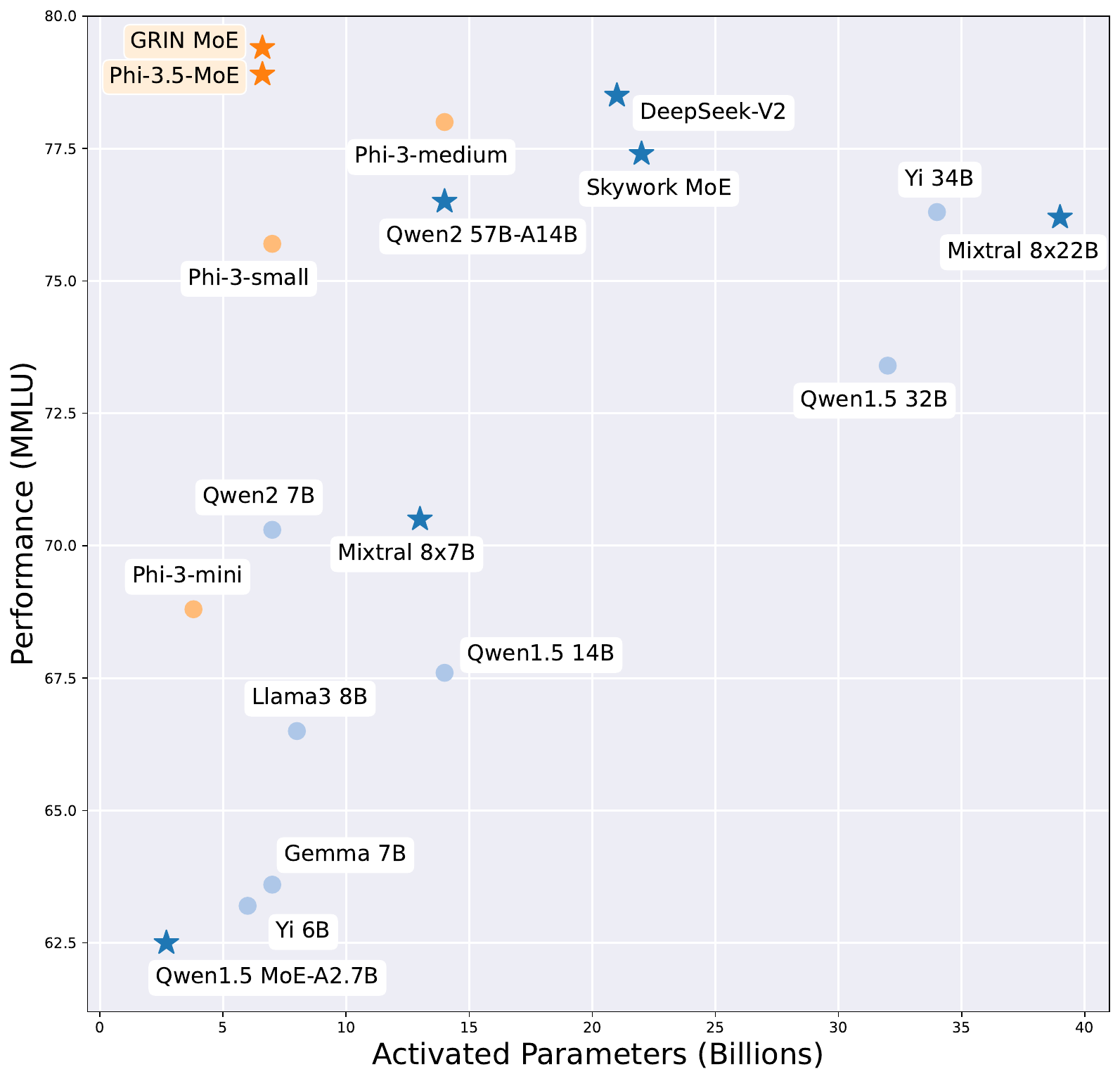}
\captionof{figure}{MMLU accuracy and activated parameters.}
\label{fig:mmlu}
\end{minipage}
\hfill\,
\begin{minipage}[b]{0.34\textwidth}

% \begin{minipage}{0.4\linewidth}
% \begin{table}[t!]
% \footnotesize
% \centering
{
\small
\begin{tabular}{l c}
\toprule
Parameter  & Value\\
\midrule
\multicolumn{2}{c}{General}\\
\midrule
\texttt{vocab\_size} & 32064\\
% \texttt{pretrain\_context\_len} & 8192\\
\texttt{n\_position} & 4096\\
\texttt{n\_layers} & 32\\
\texttt{n\_embed} & 4096\\
\texttt{normalization} & LayerNorm \\

\midrule
\multicolumn{2}{c}{Attention specific}\\
\midrule
\texttt{window\_size} & 2048\\
\texttt{n\_head} & 32\\
\texttt{n\_kv\_head} & 8\\
\texttt{head\_dim} & 128\\
\texttt{rotary\_dim} & 128 \\
\midrule
\multicolumn{2}{c}{Expert specific}\\
\midrule
\texttt{activation} & SwiGLU\\
\texttt{inner\_dim} & 6400\\
\texttt{n\_expert} & 16\\
\texttt{top\_k} & 2\\
\texttt{moe\_mod} & every layer\\

\bottomrule
\end{tabular}
}
\vspace{.2cm}
% \caption{Model architecture parameters.}
\label{tab:model_params}
% \end{table}

\captionof{table}{\ourmodel Specifics.}
% \end{minipage}
\end{minipage}
\end{minipage}
\end{figure}

We have applied our method, \emph{GR}adient-\emph{IN}formed MoE, to autoregressive language modeling and developed a 16x3.8B MoE model, referred to as \ourmodel\footnote{GRIN MoE weights: \url{https://huggingface.co/microsoft/GRIN-MoE}. Note a different version of mid-training and post-training, emphasizing long context and multilingual ability, has been conducted and has been released at \url{https://huggingface.co/microsoft/Phi-3.5-MoE-instruct}.}. 
The model incorporates MoE layers as its feedforward network, conducting top-2 routing among 16 experts at each layer.
Specifically, each expert is implemented as a GLU network and the routing module selectively activates 2 out of 16 expert networks for each token at each layer. 
As a result, the 16$\times$3.8B model has 42B parameters in total and 6.6B activated parameters for inference. 

Our evaluations across a diverse set of tasks reveal that \ourmodel achieves exceptionally good performance, particularly in coding and mathematics tasks. 
For example, our model scores 79.4 on MMLU, 74.4 on HumanEval, 80.3 on MBPP, and 58.9 on MATH. 
\ourmodel also compares favorably with dense models trained on the same data.
For instance, \ourmodel achieves an average score of 79.58 in Table~\ref{table:bbeval}, outperforming 7B dense model (average score of 75.74) and matching the 14B dense model (average score of 78.46) trained on the same data.

In addition, we have conducted an in-depth analysis to shed light on why \ourmodel works well. 
We show that \ourmodel can answer sophisticated questions correctly and quickly pick up on hints, and that \ourmodel exhibits different expert assignment patterns for different tasks and domains, indicating that its experts have developed specialized expertise and that the router can effectively compose these experts.

% \todo{update figure}
% \todo{fix remaining contents}

\section{Model Architecture}
\label{sec:backgrounds}

%\subsection{Architecture}
%In this part, we introduce the architecture of our MoE model, \ourmodel, with necessary backgrounds.
Similar to existing state-of-the-art MoE models, \ourmodel is based on a transformer architecture \citep{vaswani2017attention} with a stack of transformer blocks.

\smallsection{Transformer}
The Transformer network is constructed by stacking Transformer blocks, each consisting of an attention layer and a feedforward layer.
Residual connections and layer normalization are applied to all sub-layers in a Pre-LN manner.

\smallsection{Attention}
Following Mistral \citep{mistral2023report}, we implement the attention layer with grouped-query attention \citep{ainslie-etal-2023-gqa} and sliding window attention \citep{child2019swa}.
Both techniques are computationally efficient and allow \ourmodel to attend information beyond the window size.
RoPE is adopted for the position encoding to enable long context encoding after pretraining\citep{rope-paper}.
%which enables easier length extension after pretraining. 
Our implementation is mostly based on FlashAttention 2 \citep{Dao2023FlashAttention2FA}.

\smallsection{Mixture of Experts}
Different from conventional Transformer models, we construct the feedforward layer as a Mixture-of-Experts layer, employing a router network to sparsely activate selected networks for each input. 

The idea of MoE is originally discussed in 
\citet{Jacobs1991AdaptiveMO} and \citet{Jordan1994HierarchicalMO}, which integrates separate networks together and uses each to handle a separate subset of training cases. 
Recently, many attempts have been made to leverage MoE for scaling large language models~\citep{Shazeer2017OutrageouslyLN, Lepikhin2020GShardSG,Lewis2021BASELS,kim2021scalable,Lepikhin2020GShardSG,fedus2022switch,zoph2022stmoedesigningstabletransferable}.

For each MoE layer, the model picks from a set of distinct feedforward networks for every input tokens, which is determined by a router network.
Particularly, given $n$ expert parameters $\{\vw_0, \cdots, \vw_{n-1}\}$, the output of one MoE module for inference is 
\begin{eqnarray}
    &\sum_{i=0}^{n-1} Gating(\vz)_i \cdot TopK(\vz)_i \cdot Expert(\vx, \vw_i),
    \label{eqn:moe}
\end{eqnarray}
where $\vz = Router(\vx, \vr)$, $\vr$ is the router parameters, $Gating(\cdot)$  is a gating function (usually softmax), and $Expert(\cdot)$ is a FNN.
In our study, we define use a linear network as the router, i.e., $Router(\vx, \vr) = \vx \cdot \vr^T$
As to $TopK(\vz)$, it is the TopK function, i.e., $TopK(\vz)_i := 1$ if $\vz_i$ is among the TopK coordinates of $\vz$ and $TopK(\vz)_i := 0$ otherwise. 

During model training, different MoE algorithms may produce different outputs, as
we will discuss in detail in Section~\ref{sec:training}.
%elaborate more on the model training in Section~\ref{sec:training}. 

\newcommand{\datasetcell}[3]{\makecell{\large #1  \\[-0.5ex] \tiny (#2)    }  }
 
\begin{table}[t]
\caption{Model Performance on Popular Benchmarks}
\vspace{0.2cm}
\centering
% \caption{Results on the GLUE development set. S refers to Switch and S+S refers to Switch+SparseMixer. AVG is the average score across eight tasks. }
% \resizebox{1.03\linewidth}{!}{
% \hskip-.5cm

\resizebox{1.0\linewidth}{!}{
% \addtolength{\tabcolsep}{-0.3em}
%\hskip-2cm
\begin{tabular}{ c|ccccccccccccc } 
% \toprule
& \multicolumn{4}{c}{\large MoE} & \multicolumn{6}{c}{\large Dense} & \multirow{ 2}{*}[-1em]{\makecell{GPT-3.5 \\ \footnotesize v1106 }} & \multirow{ 2}{*}[-1em]{\makecell{GPT-4o \\ \footnotesize 2024}} & \multirow{ 2}{*}[-1em]{\makecell{Gemini \\ \footnotesize 1.5-Flash}}\\
\cmidrule(lr){2-5}
\cmidrule(lr){6-11}
& \ourmodel & Phi-3.5 & \multicolumn{2}{c}{Mixtral} & \multicolumn{2}{c}{Phi-3} & Mistral & Gemma & \multicolumn{2}{c}{Llama3} \\
& 16$\times$3.8B & 16$\times$3.8B & 8$\times$7B & 8$\times$22B  & 14B & 7B & 7B & 7B & 8B & 70B\\ 
% &\makecell{Psi\\ \footnotesize 3.8bx16 }& \makecell{Mixtral\\ \footnotesize 8x7b }  & \makecell{Mixtral\\ \footnotesize 8x22b } & \makecell{DeepSeek\\ \footnotesize V2 } &  \makecell{Phi-3\\ \footnotesize 14b  } &  \makecell{Phi-3\\ \footnotesize 7b  } & \makecell{Mistral\\ \footnotesize 7b } &\makecell{Gemma \\ \footnotesize 7b }&\makecell{Llama-3 \\ \footnotesize 8b }  &\makecell{Llama-3 \\ \footnotesize 70b }&    \\
\midrule 
\datasetcell{MMLU}{5-Shot}{\cite{hendrycks2021measuring} }         & 79.4 & 78.9 & 70.5 & 76.2 & 78.0&75.7 & 61.7 & 63.6 & 66.5 &80.2& 71.4&86.9&79.4  \\[1.5ex]
\datasetcell{HellaSwag}{5-Shot}{\cite{zellers2019hellaswag} }      & 83.7 & 83.8 & 70.4 & 79.0 & 82.4&77.0 & 58.5 & 49.8 & 71.1 &82.6& 78.8&91.7&-  \\[1.5ex]
\datasetcell{ANLI}{7-Shot}{\cite{nie2020adversarial}}              & 60.6 & 59.8 & 55.2 & 65.2 & 55.8&58.1 & 47.1 & 48.7 & 57.3 &68.3& 58.1&75.7&65.6  \\
% \midrule 
\datasetcell{ GSM-8K}{8-Shot; CoT}{\cite{cobbe2021training} }      & 90.4 & 88.7 & 64.7 & 83.8 & 91.0&89.6 & 46.4 & 59.8 & 77.4 &93.5& 78.1&93.8&82.4  \\[1.5ex]
% \datasetcell{Math}{0-Shot; CoT}{\cite{cobbe2021training} }         & 58.9 & 11.1 & 41.8 & 46.2&34.6 & 15.0 & 11.4 & 28.2 &51.2& 45.3 &67.8&38.0 \\
% \midrule
\datasetcell{ MedQA}{2-Shot}{\cite{jin2020disease} }               & 70.4 & 70.5 & 62.2 & 67.9 & 69.9&65.4 & 50.0 & 49.6 & 60.5 &78.5& 63.4&88.9&-  \\[1.5ex]
\datasetcell{ AGIEval}{0-Shot}{\cite{zhong2023agieval} }           & 48.2 & 50.3 & 45.2 & 54.0 & 50.2&45.1 & 35.1 & 42.1 & 42.0 &56.9& 48.4&37.6&45.2  \\[1.5ex]
\datasetcell{ TriviaQA}{5-Shot}{ \cite{joshi2017triviaqa}}         & 73.9 & 71.6 & 78.5 & 82.2 & 73.9&58.1 & 75.2 & 72.3 & 67.7 &84.5& 85.8&66.0&- \\
% \midrule 
\datasetcell{Arc-C}{10-Shot}{\cite{clark2018think} }               & 92.0 & 91.0 & 87.3 & 91.3 & 91.6&90.7 & 78.6 & 78.3 & 82.8 &93.0& 87.4&97.0&88.3 \\[1.5ex]
\datasetcell{Arc-E}{10-Shot}{\cite{clark2018think} }               & 98.0 & 97.1 & 95.6 & 96.9 & 97.7&97.0 & 90.6 & 91.4 & 93.4 &98.2& 96.3&99.0&97.1  \\[1.5ex]
\datasetcell{ PIQA}{5-Shot}{\cite{bisk2019piqa} }                  & 89.0 & 88.6 & 86.0 & 85.0 & 87.9&86.9 & 77.7 & 78.1 & 75.7 &85.3& 86.6&92.9&87.5  \\[1.5ex]
\datasetcell{ SociQA}{5-Shot}{\cite{bisk2019piqa} }                & 79.5 & 78.0 & 75.9 & 78.2 & 80.2&79.2 & 74.6 & 65.5 & 73.9 &81.1& 68.3&81.4&77.8  \\
% \midrule 
\datasetcell{ BigBench-Hard}{3-Shot; CoT}{ }                       & 81.4 & 79.1 & 69.7 & 81.8 & 81.4&79.1 & 57.3 & 59.6 & 51.5 & 80.2 & 68.3& 81.2$^*$ &- \\[1.5ex]
\datasetcell{WinoGrande}{5-Shot}{\cite{sakaguchi2019winogrande} }  & 81.4 & 81.3 & 62.0 & 75.3 & 81.5&81.5 & 54.2 & 55.6 & 65.0 &83.3& 68.8&89.3&74.7  \\[1.5ex]
\datasetcell{OpenBookQA}{10-Shot}{\cite{mihaylov2018suit} }        & 89.8 & 89.6 & 85.8 & 88.6 & 87.4&88.0 & 79.8 & 78.6 & 82.6 &91.8& 86.0&95.2&89.0  \\[1.5ex]
\datasetcell{BoolQ}{2-Shot}{\cite{clark2019boolq} }                & 83.4 & 84.6 & 77.6 & 82.7 & 86.5&84.8 & 72.2 & 66.0 & 80.9 &89.1& 79.1&90.6&85.8  \\[1.5ex] 
{\makecell{CommonSenseQA  \\[-0.5ex] \tiny (10-Shot) }}            & 81.8 & 83.5 & 78.1 & 82.0 & 82.8&80.0 & 72.6 & 76.2 & 79.0 &84.4& 79.6&88.5&84.0  \\[1.5ex]
\datasetcell{TruthfulQA}{10-Shot; MC2}{\cite{lin2022truthfulqa} }  & 74.5 & 77.5 & 60.1 & 67.4 & 75.1&70.2 & 53.0 & 52.1 & 63.2 &81.9& 85.8&85.6&76.6  \\
 
% \midrule 
\datasetcell{ HumanEval}{0-Shot}{\cite{chen2021evaluating} }       & 74.4 & 70.7 & 37.8 & 39.6 & 62.2&61.0 & 28.0 & 34.1 & 60.4 &78.7& 62.2&92.1&64.4 \\[1.5ex]
\datasetcell{ MBPP}{3-Shot}{\cite{austin2021program} }             & 80.3 & 80.8 & 60.2 & 70.7 & 75.2&71.7 & 50.8 & 51.5 & 67.7 &81.3& 77.8&90.4&77.5  \\
\midrule 
% {\large Average} & 78.55 & 66.70 & 74.48 & 76.85 & 73.69 & 58.92 & 59.22 & 67.34 & 81.2 & 73.78 & 84.76 \\
{\large Average} & 79.58 & 79.23 & 69.62 & 76.20 & 78.46 & 75.74 & 61.23 & 61.73 & 69.40 & 82.78 & 75.27 & 85.70
\end{tabular}
}
\label{table:bbeval}
\end{table}

%\subsection{Pre-training and Post-training}

%\smallsection{Pre-training} \ourmodel is pre-trained on 4T tokens as a Causal Language Model. The same training dataset has been used to train Phi3 dense models~\citep{abdin2024phi}. 

%\smallsection{Post-training} 
%Post-training consists of two stages: Supervised Fine-Tuning (SFT) based on the causal language modeling objective, followed by Direct Preference Optimization (DPO;  \citeauthor{rafailov2024direct}, \citeyear{rafailov2024direct}). The model is trained with 24B tokens in SFT, using high-quality data across diverse categories, e.g., math, coding and conversation \citep{abdin2024phi}. The DPO dataset contains 1.4B tokens, including safety and identity preference data that is used to align the model output with Microsoft's Responsible AI principles. We further adopt regularization techniques, such as adding random noise to the input embedding \citep{jain2023neftune} and applying dropout in expert layers \citep{fedus2022switch}, to improve the model's generalization performance.
%It is worth mentioning that another version of mid-training and post-training have been conducted with an emphasize on long context and multilingual ability, which has been released as Phi3.5 MoE\citep{abdin2024phi}. 
\section{GRIN MoE}
\label{sec:training}

%We recognized engineering challenges and training challenges that play pivotal roles in pursuing the scaling power of MoE. 
%Many of these challenges remain as open questions and our study has been heavily inspired by prior state of the art. 
%Accordingly, we summarize our observations and intuitions here, in hope they can further shed insights in MoE training and guide future algorithm developments. 

% \smallsection{Pre-training} \ourmodel is pre-trained on 4T tokens as a Causal Language Model. The same training dataset has been used to train Phi3 dense models~\citep{abdin2024phi}. 

% \smallsection{Post-training} 
% Post-training consists of two stages: Supervised Fine-Tuning (SFT) based on the causal language modeling objective, followed by Direct Preference Optimization (DPO;  \citeauthor{rafailov2024direct}, \citeyear{rafailov2024direct}). The model is trained with 24B tokens in SFT, using high-quality data across diverse categories, e.g., math, coding and conversation \citep{abdin2024phi}. The DPO dataset contains 1.4B tokens, including safety and identity preference data that is used to align the model output with Microsoft's Responsible AI principles. We further adopt regularization techniques, such as adding random noise to the input embedding \citep{jain2023neftune} and applying dropout in expert layers \citep{fedus2022switch}, to improve the model's generalization performance.

In this section, we discuss in detail the two key techniques used in \ourmodel:
%Particularly, the training of \ourmodel differs from conventional MoE training in mostly two aspects:

\begin{itemize}[leftmargin=*]
\item We propose \sm to estimate the gradient related to expert routing, while the conventional MoE training treats expert gating as a proxy for the gradient estimation. 
\item We scale MoE training with neither expert parallelism nor token dropping, while the conventional MoE training employs expert parallelism and deploys token dropping.
\end{itemize}

\subsection{Gradient Estimation for Expert Routing}

\begin{figure}[t]
    \centering
    \vspace{-10pt}
    \includegraphics[width=0.8\linewidth]{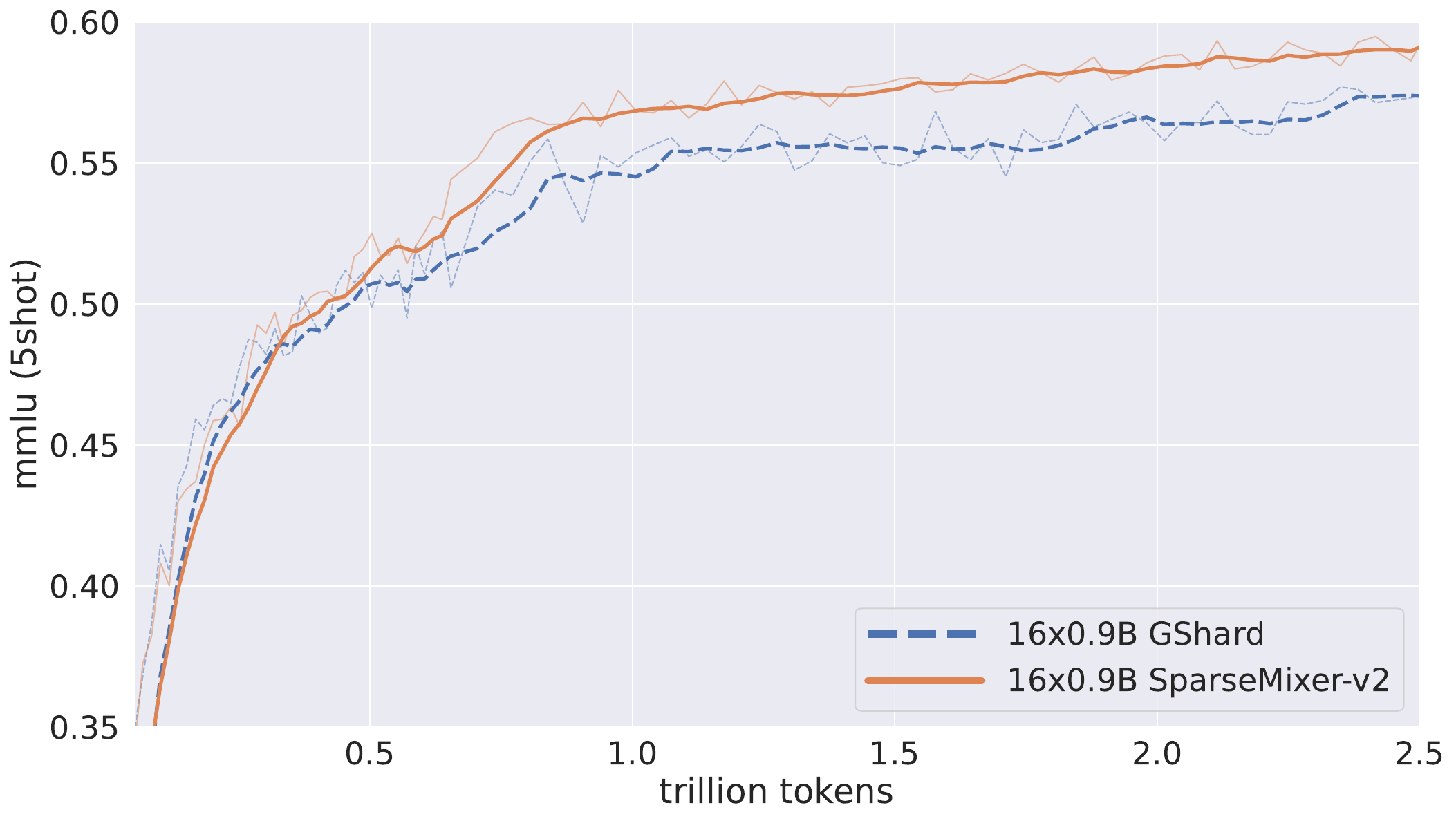}
    \caption{Controlled Comparisons of \sm and GShard on 16$\times$0.9B MoE.}
    \label{fig:msr-controlled}
\end{figure}
Expert routing produces discrete expert assignment, allowing the network to be sparsely activated and thus bringing the great scaling potential. 
At the same time, such a routing process is not differentiable, thus making it infeasible to directly apply the vanilla backpropagation and obtain reliable gradient estimation. 

\smallsection{Conventional MoE Training}
Conventional MoE training uses the router outputs to compute gating results, treating the gating gradients as a proxy of the router gradient. 
Particularly, as in Equation~\ref{eqn:moe}, the output of the MoE module is: 
$\sum_{i=0}^{n-1} Gating_i(\vz) \cdot TopK(\vz)_i \cdot Expert(\vx, \vw_i)$, where  $\vz = Router(\vx, \vr)$.
Conventional MoE training views $TopK(\vz)_i$ as constants and only back-propagates through $Gating(\vz)$ to compute router weight gradients.
In this way, it treats the gating gradient as a proxy of the router gradient, i.e., $\nabla_{ \mbox{\scalebox{.7}{conventional}}} {\scalebox{1.3}{$\bm{r}$}} := \nabla Gating(\vz) \cdot \frac{\partial Gating(\vz)}{\partial \vr}$.

\smallsection{\sm}
Inspired by recent advances in Straight-Through gradient estimators~\citep{Bengio2013EstimatingOP,Liu2023BridgingDA}, we proposed the \sm method, an extension of SparseMixer \citep{Liu2023SparseMixer}, to obtain scalable and reliable gradient estimations in this study.
%in \cite{Liu2023SparseMixer} and further adapt it as the \sm method used in this study.

We briefly introduce the \sm method below, and leave a detailed description to Appendix~\ref{appendix:sparsemixer}. 
We first replace the $TopK(\cdot)$ function as random sampling of discrete variables in model training.  
Then, following \cite{Liu2023BridgingDA} and \cite{Liu2023SparseMixer}, we apply Heun's third order method to approximate the expert routing gradient and construct a modified back-propagation to give a mathematically sound gradient estimation for expert routing. 

\smallsection{Effectiveness of \sm} % SparseMixer}
In \cite{Liu2023SparseMixer},  the effectiveness of SparseMixer is demonstrated on the neural machine translation task and the ELECTRA language model training.
However, it has not been applied to autoregressive language model training at a large scale. 
In the development of \ourmodel, we conducted controlled experiments that showd promising results for \sm. 
The result motivates us to apply this algorithm to training \ourmodel. 

Particularly, we trained two 16$\times$0.9B MoEs with 2.5T tokens.
One of them follows the same recipe used in \ourmodel, and the other replaces \sm with the conventional GShard method. 
As shown in Figure~\ref{fig:msr-controlled}, the performance boost of \sm generalizes to the autoregressive language model training at the 16$\times$0.9B scale: although GShard performs better at the first 0.5T tokens, \sm achieves  stronger performance in the later stage of training. 
It is worth mentioning the similar phenomenon observed in small scale experiments on ELECTRA pretraining that Switch tends to perform better in the beginning while SparseMixer comes from behind in the late stage~\citep{Liu2023SparseMixer}. 
We suspect this is due to the model architecture difference, e.g., it introduces more randomness to training by replacing $TopK(\cdot)$ function with random sampling of discrete variables, which may slow down the training in the begining.
Also, it is worth mentioning that, such extra randomness also makes it harder to compare the training loss to the GShard.
Semi-controlled experiment results of a larger scale are discussed in Section~\ref{sec:vis}.

% \smallsection{Third-order ODE Solver} 
% The original SparseMixer study adopts the mid-point method for gradient estimation. 
% Here, we upgraded it to the third-order Heun's method. 
% In our study, we also tried ODE solvers with higher order and didnt observe further performance gain. 

% \smallsection{Output Scaling and Initialization}
% Another important change is that, while the original SparseMixer study uses the output scaling to stabilize model training, we removed the output scaling and relies on model initialization for training stabilization. 

% Here, we share the early stage results 

\subsection{Implementation and Scaling}
\label{subsec:computational_scaling}

Comparing to conventional models that activate all parameters for all inputs, MoE models have more parameters for the same FLOPs due to their structured sparse computation, significantly impacting computational efficiency.
Conventional MoE training distributes different expert networks across devices (i.e., expert parallelism) and employs strategies like token dropping to facilitate the training process.

As our first step towards pursuing the scalability brought by MoE, we focus on MoE training with a relative small number of experts (i.e., top2 routing over 16 experts). 
Leveraging recent engineering advances, we avoid expert parallelism and eliminate the need for capacity factor or token dropping.
In the end, we are able to achieve over 80\% relative training efficiency improvement, compared to a dense model with the same active parameters, for \ourmodel training.

% without involving capacity factor or token dropping.
% % and rely on other parallelism for MoE scaling. 
% Benefited from recent engineering advances, our training framework reduces the gap between dense model training and MoE model training. 

% the Megablocks~\citep{Gale2023MegaBlocksES} package to be very helpful. 
% Particularly, we find its \verb|grouped_GEMM| kernels and wrappers outperform its \verb|sparse| version and bring substantial performance improvements. 
% \subsubsection{Implementation and Throughput}

\smallsection{MoE Implementation}
For MoE computation without expert parallelism, we find the Megablocks~\citep{Gale2023MegaBlocksES} package to be very helpful. 
Particularly, we find its \verb|grouped_GEMM| kernels and wrappers outperform its \verb|sparse| version, offering substantial performance improvement. 
In addition, we rely on data parallelism, pipeline parallelism, and activation checkpointing in the training of \ourmodel, which lead to the best throughput for our 16$\times$3.8B model. 
% We also implemented tensor parallelism and find it particularly helpful to support a large dictionary or a large number of experts. 

\smallsection{Training Throughput Comparisons of Dense and MoE Models}
To showcase the benefits of MoE training, we compare its training throughput to that of a conventional dense model. 
Hardware details for these studies are in Appendix~\ref{appendix:hardware}. 
It is important to note that, the throughput of the dense model is measured under the same parallelism setting as that of the MoE model, 
% Although this is not the optimal parallelism setting for the dense model, the result is significant enough to %is  and is merely used as a reference to 
% demonstrate the computational efficiency of the MoE training. 
and the comparison here is to study the GPU kernel efficiency of densely activated networks (i.e., Dense) and sparsely activated networks (i.e., MoE). 

\begin{table}[t]
\centering
\caption{Training Throughput of Dense and MoE on 64 H100 gpus.}
\label{table:efficiency}
% \vspace{-0.3cm}
\scalebox{.9}{
\begin{tabular}{l|ccccc|ccc}
\toprule
  & Total Parameters & Active Parameters & Throughput Per GPU & Relative Throughput    \\   
  \midrule
Dense & 1.6B & 1.6B & 34222 & \\
MoE & 10B & 1.6B & 27962 & 81.71\% \\
  \midrule
Dense & 6.6B & 6.6B & 8176 & \\
MoE & 42B & 6.6B & 7077 & 86.56\% \\
\bottomrule
\end{tabular}
}
% \vspace{-.3cm}
\end{table}

As summarized in Table~\ref{table:efficiency}, we compare MoE models of two different sizes to their corresponding dense models with the same number of parameters, measuring their training throughput using the identical hardware.
Despite having over six times as many parameters as the dense model, MoE models achieve more than 80\% relative throughput in this experiment, confirming the significant computational scaling potential of models with GRIN MoE method.

Additionally, our observations indicate that MoE models do not experience more severe or different throughput degradation compared to dense models when scaling up model size. 
Both dense and MoE models show similar slowdown patterns in our experiments. 
For instance, the training throughput of the 6.6B dense model is approximately 4.19 times slower than that of the 1.6B dense model, which has 4 times fewer parameters. Similarly, the training throughput of the 42B MoE model is about 3.96 times slower than that of the 10B MoE model, which has 4.2 times fewer parameters.

\begin{figure}
    \centering
    % \vspace{-10pt}
    \includegraphics[width=0.8\linewidth]{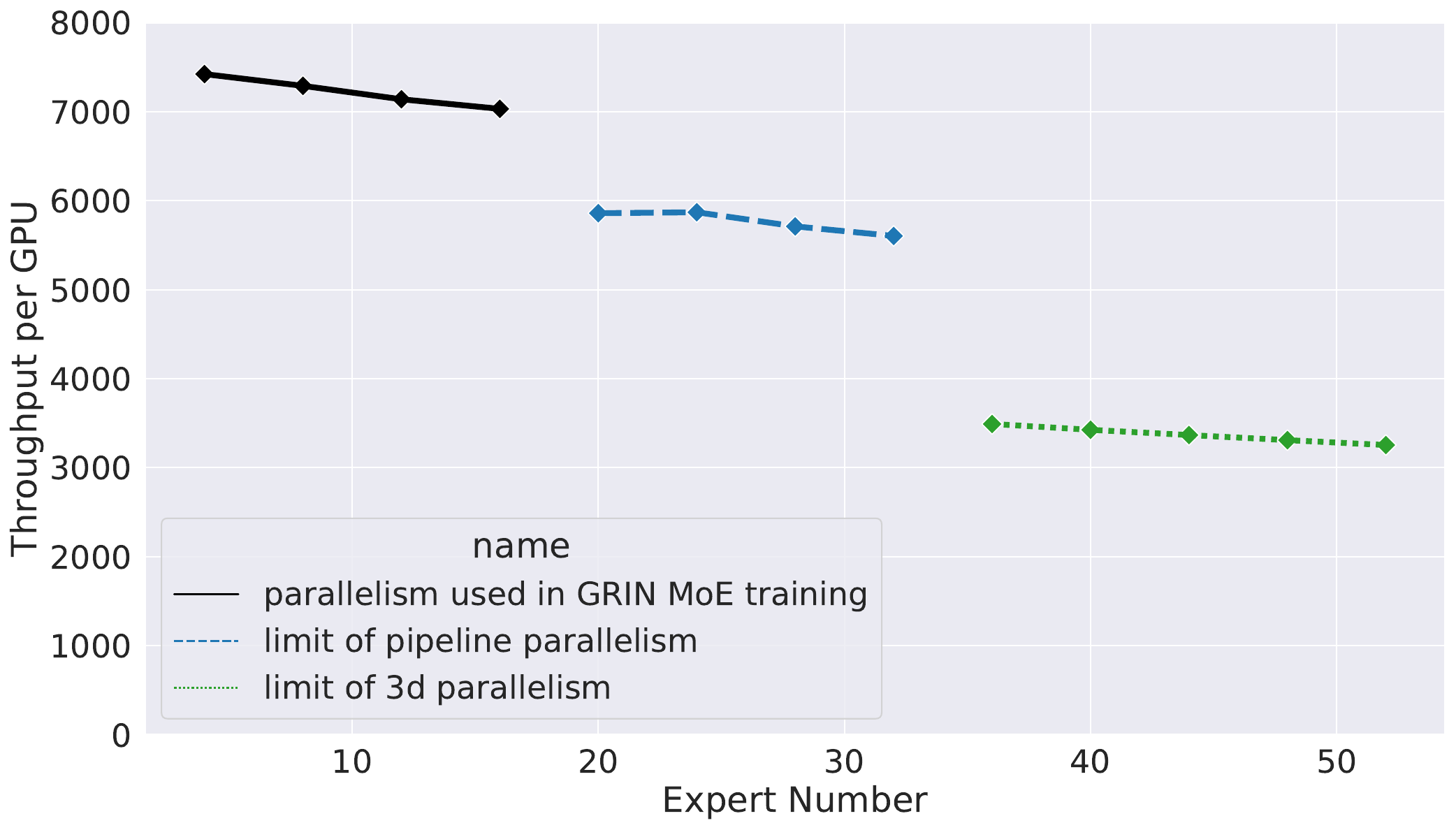}
    \caption{Scaling of Different Parallelism Settings on 64 H100 gpus. The reported throughput for N experts (x-axis) refers to the average training throughput of a 3.8BxN top2 MoE. }
    \label{fig:throughput}
\end{figure}

\smallsection{Scaling Study and Tensor Parallelism}
% Despite the efficiency of the parallelism setting used in the training of \ourmodel, its capacity has a limit. 
% As in Figure~\ref{fig:throughput}, it can only support training with up to 16 experts. 
In this section, we discuss whether it is computationally feasible to train MoE models with a larger number of experts without employing expert parallelism.

First, by relying solely on pipeline parallelism, the maximum number of experts can be extended from 16 to 32 by further partitioning different layers across GPUs. 
However, increasing the number of experts beyond this would result in too many parameters for a single layer, making it difficult to support without partitioning one layer across multiple GPUs.

To address this challenge, conventional MoE training relies on expert parallelism to further partition the model, which introduces the side effects of capacity factor and token dropping. 
In our study, we explore the use of tensor parallelism instead of expert parallelism~\citep{Narayanan2021EfficientLL}. 
Similar to expert parallelism, which has two all-to-all communication overheads in both forward and backward computations, tensor parallelism has two all-reduce communication overheads in forward and backward computations. 
Although all-reduce operations have higher latency than all-to-all operations, we can mitigate these overheads by overlapping communication with computation during backward computation.

As in Figure~\ref{fig:throughput}, the maximum number of supported experts is extended to 52 (with 132B total parameters) by combining pipeline parallelism and tensor parallelism (i.e., 3D parallelism). 
It is worth noting that, since our throughput study hardware setting has only 64 GPUs, we can partition the model into at most 64 stages. 
With 272 H100 GPUs, the limit of this parallelism can be further extended to 200+ experts.

While this demonstrates the feasibility of scaling MoE training without expert parallelism, we observe that using more complex parallelism typically leads to reduced computational throughput.
Correspondingly, one important direction for our future work is to perform MoE training with more experts in a more sparse manner.

\subsection{Global Load Balance Loss Adaptations}
\label{subsec:auxloss}
As previously discussed, our training framework is designed to scale MoE training using tensor parallelism, pipeline parallelism, and data parallelism, but not expert parallelism. 
Correspondingly, there is no need to employ token dropping or capacity factor, allowing the distribution of activated experts to deviate from a uniform distribution.

Therefore, we adapt the load balance loss to regulate the global expert load balance instead of the local one. 
Particularly, the popular load balancing loss is defined as :
\begin{eqnarray}
\alpha \cdot n \cdot \sum_{i=1}^n f_i \cdot E[\texttt{softmax}(\vz)_i],
\nonumber
\end{eqnarray}
where $\alpha$ is a hyper parameter, $n$ is the number of experts, and $f_i$ is the fraction of tokens dispatched to expert $i$~\citep{fedus2022switch}.
Conventionally, $f_i$ is computed at different GPUs locally and thus the load balance loss will regulate the local expert load balance and alleviate token dropping. 
In our study, we modified the load balance auxiliary loss by computing $f_i$ globally (i.e., all-reduced within the data-parallel process group), regulating the expert load to be balanced globally. 
Although this adjustment incurs additional communication overheads, similar to tensor parallelism, 
% these communications can be reduced by overlapping them with other MoE computations. 
these communications can be performed in parallel with computations in an asynchronized way, thus largely reducing the additional latency.

\section{Experiment}

\subsection{Training Setting}

\smallsection{Pre-training} \ourmodel is pre-trained on 4T tokens as a Causal Language Model. The same training dataset has been used to train Phi-3 dense models~\citep{abdin2024phi}.

\smallsection{Post-training} 
Post-training consists of two stages: Supervised Fine-Tuning (SFT) based on the causal language modeling objective, followed by Direct Preference Optimization (DPO;  \citeauthor{rafailov2024direct}, \citeyear{rafailov2024direct}). The model is trained with 24B tokens in SFT, using high-quality data across diverse categories, e.g., math, coding and conversation \citep{abdin2024phi}. The DPO dataset contains 1.4B tokens, including safety and identity preference data that is used to align the model output with Microsoft's Responsible AI principles~\citep{Haider2024Phi3SP}. We further adopt regularization techniques, such as adding random noise to the input embedding \citep{jain2023neftune} and applying dropout in expert layers \citep{fedus2022switch}, to improve the model's generalization performance.
It is worth mentioning that another version of mid-training and post-training have been conducted with an emphasize on long context and multilingual ability, which has been released as Phi-3.5-MoE~\citep{abdin2024phi}. 

\subsection{Evaluation of \ourmodel}

Table~\ref{table:bbeval} summarizes the performance of \ourmodel on popular benchmarks. 
Benchmarks and baseline methods are elaborated in Appendix~\ref{appendix:hardware}.

Since both Phi-3 and \ourmodel models are trained on the same datasets, the effectiveness of our MoE training recipe is easily demonstrated.
We can see that \ourmodel with 6.6B activated parameters performs significantly better than 7B dense model and similar to the 14B dense model. Compared to the 14B dense model, \ourmodel performs better on math, coding, and MMLU tasks.

Comparing GRIN MoE to Phi-3.5-MoE, which has been developed with a different focus (i.e., multilingual capabilities and long context handling), we find that these two models have distinct strengths. 
We observed that GRIN MoE excels in math and reasoning tasks, while Phi-3.5-MoE demonstrates superior performance in question-answering (QA). 
Despite their different strengths, both models yield similar average scores across various benchmarks, which is expected given that both are configured as 16x3.8B MoEs and both are trained with sparse backpropagation. 
Further comparisons are available in Section\ref{subsec:gaokao}. 

%Compare to open-source models, 
Our evaluation also shows that \ourmodel is significantly better than many open-sourced models with a similar number of active parameters, such as Mixtral 8$\times$7B (12.9b activated parameters), Mistral 7B, Gemma 7B, Llama3 8B. And \ourmodel is better than Mixtral 8$\times$22B on most of the tasks. 
Nevertheless, \ourmodel's performance still falls short of Llama3 70B and GPT-4o. 
This gap is expected, given the substantially larger computational and data resources utilized in training these two latter models.

\subsection{Math Ability Case Study}
\label{subsec:gaokao}

Phi-3 data, the training data of \ourmodel, contains a massive amount of synthetic data, greatly boosting model performance on benchmarks. 
Despite its effectiveness, it left doubts on the performance of \ourmodel on real-world tasks. 
Correspondingly, we conduct case studies on math questions of the newly released GAOKAO exam (i.e., Nationwide Unified Examination for Admissions to General Universities and Colleges), which is the annual national undergraduate admission exam in China.
Known for its rigorous security protocols, this exam serves as an ideal "truly held-out" test bed for assessing AI models' ability to answer math questions. 
Note that the training of \ourmodel concludes on June 3, Pacific Standard Time, and the 2024 GAOKAO starts on June 7, China Standard Time.  

\begin{figure}
    \centering
    \vspace{-10pt}
    \includegraphics[width=1.0\linewidth]{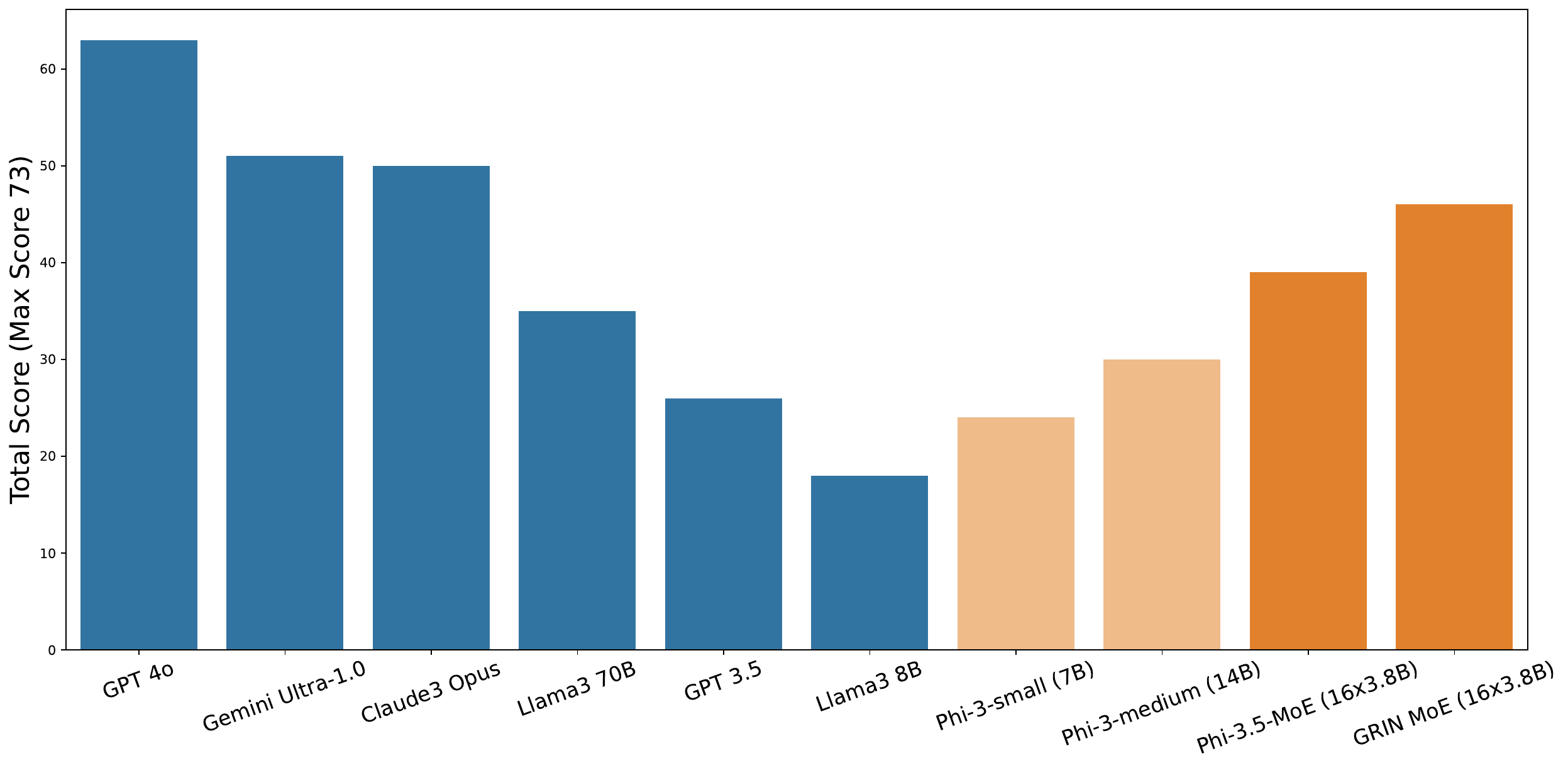}
    \caption{Test Score on Translated 2024 GAOKAO Math-1.}
    \label{fig:gaokao}
\end{figure}

\smallsection{Exam Score}
To assess the capability of various models in answering math questions, we used translated questions as the input\footnote{We used the translated questions from \url{https://github.com/zhengaq/GAOKAO-Math24}.}, scored their responses manually, and visualized the result in Figure~\ref{fig:gaokao} (detailed scoring results and the response of \ourmodel can be find in Appendix~\ref{appendix:gaokao}). 
\ourmodel scored 46 out of 73 points on these questions, outperforming Llama3 70B by 11 points, and is only 6 and 5 points away from the Gemini Ultra-1.0 and Claude3 Opus, respectively. 

These results demonstrate \ourmodel’s strong capacity for mathematical reasoning.
As the chance of data leakage in this case study is slim, the result suggests that the capacity of \ourmodel is likely attributed to the generative distillation approach~\citep{hsieh2023distilling,mukherjee2023orcaprogressivelearningcomplex}, instead of memorization. %data leakage. 

\begin{table}[t]
\centering
\caption{\ourmodel performance on LiveBench-2024-07-25. Models are ranked by their average score (AVG). $^*$Baseline results are referenced from the official benchmark.}
\label{table:livebench}
\scalebox{.8}{
\begin{tabular}{l|cccccc|c}
\toprule
  & Reasoning & Coding & Mathematics & Data Analysis & Language & IF & AVG \\   
  \midrule
Claude-3-haiku$^*$ & 29.3 & 24.5 & 25.7 & 41.5 & 30.1 & 64.0 & 35.9 \\
Mixtral-8x22B-instruct-v0.1$^*$ & 29.3 & 32.0 & 28.3 & 31.7 & 26.5 & 63.1 & 35.2 \\
GPT-3.5-turbo-0125$^*$ & 26.7 & 27.7 & 26.9 & 41.2 & 24.2 & 60.5 & 34.5 \\
\midrule
\ourmodel  & 35.3 & 23.7 & 29.8 & 32.0 & 16.9 & 57.6 & 32.5 \\
\midrule
Mistral-small-2402$^*$ & 26.0 & 21.2 & 28.2 & 31.9 & 22.1 & 63.9 & 32.2 \\
Command-r-plus$^*$ & 28.7 & 19.5 & 24.9 & 24.6 & 23.9 & 71.5 & 32.2 \\
Gemma-2-9B-it$^*$ & 17.3 & 22.5 & 24.0 & 35.1 & 27.6 & 61.6 & 31.3 \\

\bottomrule
\end{tabular}
}
\end{table}

\smallsection{\ourmodel Responses Discussion}
Analyzing the response of \ourmodel to these questions, we have some interesting observations:
\begin{itemize}[leftmargin=*]

\item \ourmodel is capable of answering challenging math questions. For instance, question 4 requires the model to perform algebraic manipulations involving trigonometric functions with an unknown variable.
As in Appendix~\ref{r4}, \ourmodel not only produces the correct answer but also presents all the necessary derivations accurately. 
Notably, \ourmodel is the only model among those we tested, including Llama3 and the Phi-3 models, to correctly answer this question.

\item For single-choice questions and multiple-choice questions, sometimes \ourmodel produces the right or partially right answers, together with  chain-of-thought prompts that have typos (question 5 as in Appendix~\ref{r5} and question 9 as in Appendix~\ref{r9}), skipped steps (question 7 as in Appendix~\ref{r7}) or errors (question 10 as in Appendix~\ref{r10} and question 11 as in Appendix~\ref{r11}). Additionally, we observe that the output of \ourmodel may change dramatically for challenging questions. These phenomenons are also observed in the responses of other models. 

\item \ourmodel has the ability to quickly catch a hint. For question 13, as in Appendix~\ref{r13}, \ourmodel initially makes a mistake by assuming that the point $(0, 1)$ is on the curve $y = \ln(x + 1) + a$, leading to an incorrect answer. However, after incorporating a hint that highlights the relationship between the curve and the point, \ourmodel adjusts its response significantly and solves the problem correctly (as in Appendix~\ref{r13p}). It is noteworthy that \ourmodel is the only model to generate the correct answer to this question with the hint, among all Llama3 and Phi-3 dense models.

\end{itemize}

When comparing GRIN MoE to Phi-3.5-MoE, different response patterns have been observed. As shown in Appendices~\ref{phi-r1} and \ref{phi-r7}, Phi-3.5-MoE occasionally generated responses without chain-of-thought prompts, while GRIN MoE consistently provided chain-of-thought responses for all 14 questions. 
Furthermore, as in Appendices~\ref{phi-r12} and \ref{phi-r4}, Phi-3.5-MoE sometimes produced responses with repetitive outputs at the end, whereas GRIN MoE maintained proper conclusions for all 14 questions. We suspect this behavior may be related to the different focus of the training in Phi-3.5 models which include handling of a longer context window up to 128K tokens and multilingual capabilities.

Moreover, the answers produced by Phi-3.5-MoE differ significantly from those of GRIN MoE. The models provided identical answers for only 5 out of 14 questions. Even for questions both models answered correctly, they offered different explanations. Additionally, while both models made the same mistake on question 13 (as shown in Appendices~\ref{r13} and \ref{phi-r13}), GRIN MoE was able to quickly incorporate the provided hint and answer the question correctly (as in Appendix~\ref{r13p}), whereas Phi-3.5-MoE failed (as in Appendix~\ref{phi-r13p}).

\subsection{Limitations and Weakness}

Since the Phi-3 data, the training corpus of \ourmodel, is constructed with an emphasize on reasoning and coding abilities, we observe the model to yield a suboptimal performance on natural language tasks. 
We use the 2024-07-25 release of the LiveBench for model evaluation~\citep{white2024livebench} and summarize the performance of \ourmodel in Table~\ref{table:livebench}, which also shows the performance of 6 other models that have similar average scores.

Comparing to baselines having similar average score on this benchmark, \ourmodel achieves better scores on the reasoning, coding, and mathematics. The result is consistent with our case study in Section~\ref{subsec:gaokao}. 
Meanwhile, we observe that \ourmodel achieves an exceptionally low average score (i.e., 16.9) on natural language tasks. 
We suspect that this is due to the limitation of the training corpus, since other models trained on the same corpus exhibit similar problems.

% \begin{table}[t]
% \caption{Semi-controlled Performance Comparisons (Elaborated in Section~\ref{subsec:semi}).}
% \begin{tabularx}{\linewidth}{c|YYYY}
%   & MMLU & Arc-C & WinoGrande & GSM-8K \\ \midrule
% \multicolumn{5}{l}{Dense}  \\\midrule
% 7B   & 75.1 & 89.8 & & 85.3 \\
% 14B    & 76.4 & 89.8  & 79.5      & 85.9  \\
% \midrule
% \multicolumn{5}{l}{MoE (Control recipe)} \\\midrule
% 16$\times$3.8b & 75.1 & 88.9  & 72.2      & 79.1  \\\midrule
% \multicolumn{5}{l}{MoE (Main recipe)} \\\midrule
% 16$\times$3.8b & 77.4 & 89.6  & 75.3      & 89.5 \\
% \end{tabularx}
% \label{tbl:gshard}
% \end{table}

% \begin{table}[t]
% \caption{Semi-controlled Performance Comparisons (Elaborated in Section~\ref{subsec:semi}).}
% \begin{tabularx}{\linewidth}{rl|YYYY}
% &   & MMLU & Arc-C & WinoGrande & GSM-8K \\ \midrule
% % \multicolumn{5}{l}{Dense}  \\\midrule
% \multirow{2}{*}{Dense} 
% & 7B   & 75.1 & 89.8 & & 85.3 \\
% & 14B    & 76.4 & 89.8  & 79.5      & 85.9  \\
% \midrule
% MoE (Control recipe) & 16$\times$3.8B & 75.1 & 88.9  & 72.2      & 79.1  \\\midrule
% MoE (Main recipe) & 16$\times$3.8B & 77.4 & 89.6  & 75.3      & 89.5 \\
% \end{tabularx}
% \label{tbl:gshard}
% \end{table}

\begin{table}[t]
\caption{Semi-controlled Performance Comparisons. Different from results in Table~\ref{table:bbeval} and Figure~\ref{fig:mmlu}, the reported performances here are before the posttraining. Detailed settings are elaborated in Section~\ref{subsec:semi}.}
\begin{tabularx}{\linewidth}{rl|YYY}
\toprule
&   & MMLU & Arc-C  & GSM-8K \\ \midrule
\multirow{2}{*}{Dense} 
& 7B   & 75.1 & 89.8 & 85.3 \\
& 14B    & 76.4 & 89.8  & 85.9  \\
\midrule
MoE (Control recipe) & 16$\times$3.8B & 75.1 & 88.9 & 79.1  \\\midrule
MoE (Main recipe) & 16$\times$3.8B & 77.4 & 89.6 & 89.5 \\
\bottomrule
\end{tabularx}
\label{tbl:gshard}
\end{table}
\section{Analyses}
\label{sec:vis}

As described in Section~\ref{sec:training}, we have tailored the training recipe for \ourmodel, featuring \sm and load balance loss adaptation. 
However, due to resource constraints, we were unable to set up a controlled environment to individually study the impact of each variable at the scale of \ourmodel. 
Therefore, we conducted a semi-controlled comparison to quantify the effect of the training recipe.

\subsection{Semi-controlled Setting}
\label{subsec:semi}

We compare the following two training recipes (the \textit{Main recipe} and the \textit{Control recipe}):
%We train models for 4T tokens, and compare them with the Phi3 7B and 14B models that are trained for more than 4T tokens. It is worth mentioning that no post-training has been performed in this comparisons.
\begin{itemize}[leftmargin=*]
    \item \textit{Main recipe} is the one used for \ourmodel training, as described in Section~\ref{sec:training}.
    \item \textit{Control recipe} resembles conventional MoE training recipes and is used for comparison. It differs from the main recipe in that the former replaces \sm used in the main recipe with GShard, exchanges global load balance loss for local load balance loss, and modifies several hyper-parameters.
\end{itemize}

We then trained MoE models using the two recipes on a 4T-token corpus, and compared them with the Phi-3 7B and 14B models trained on a super-set of the 4T-token corpus on downstream tasks without post-training. 

Note that comparing to the controlled experiment in Figure~\ref{fig:msr-controlled}, the control recipe does not adapts the global load balance loss as in Section~\ref{subsec:auxloss}, while the GShard baseline in Figure~\ref{fig:msr-controlled} adapts the global load balance loss adaptation.

\begin{figure}
    \centering
    \vspace{5pt}
    \includegraphics[width=1.0\linewidth]{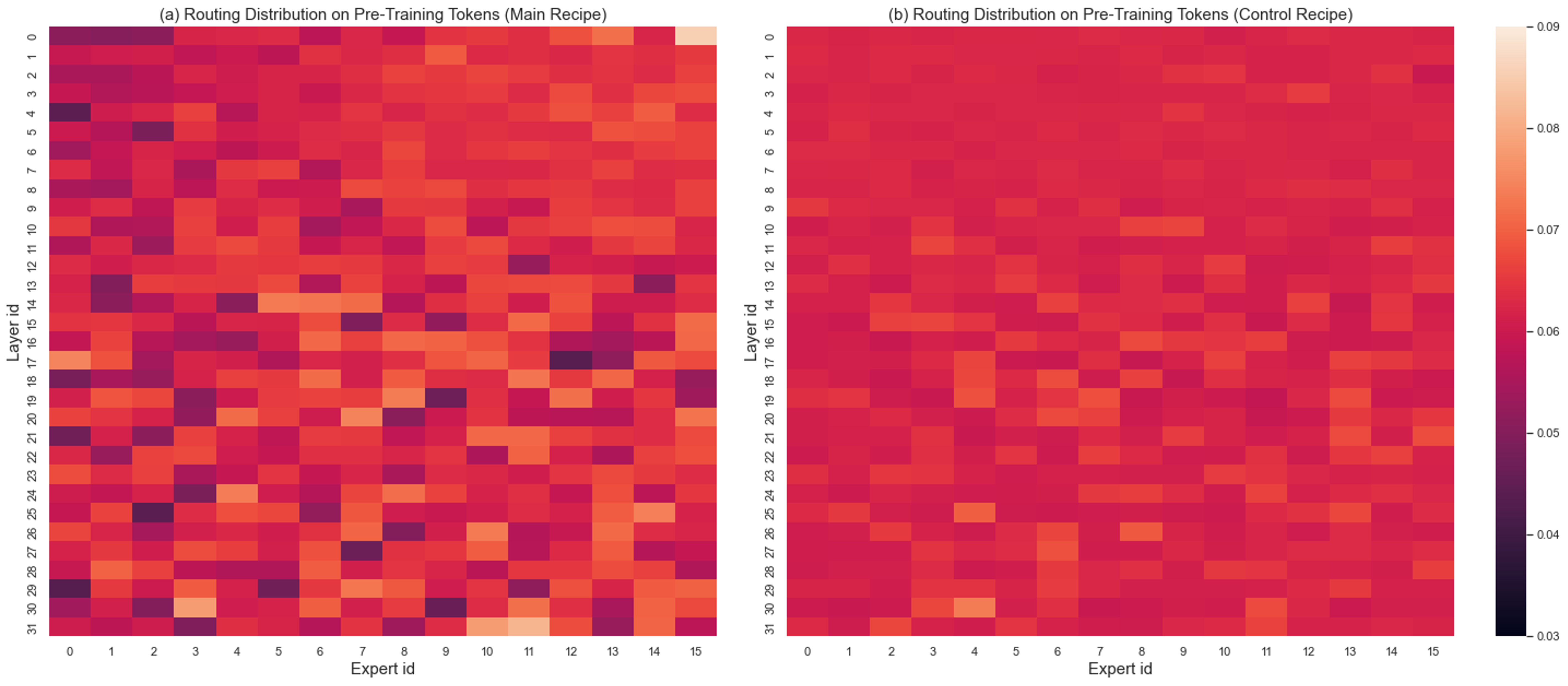}
    \caption{Routing distribution on 2 million pretraining tokens. The model on the left is trained by main recipe and the right is trained by control recipe. The values are normalized per layer. The summation of the values in each row is 1 (perfectly balanced loading would result in a value of $0.0625$).}
    \label{fig:pretrainheatmap}
\end{figure}

\subsection{Downstream Performance}
The results are presented in Table~\ref{tbl:gshard}. 
The model trained using the control recipe matches the performance of the 7B dense model. 
The main recipe is more effective, resulting in a model whose performance matches that of the 14B dense model. 
%This demonstrates the effectiveness of our tailored training approach, 
We attribute the effectiveness to the use of \sm and the adaptive loss modifications. 

\subsection{Routing Analyses}
We analyze the routing distributions of models trained with the main and control recipes. We count how many times each expert (in each layer) is selected by different hidden states as routing distribution.

\renewcommand{\thefigure}{6 (a)}
\begin{figure}
    \centering
    \includegraphics[width=1.0\linewidth]{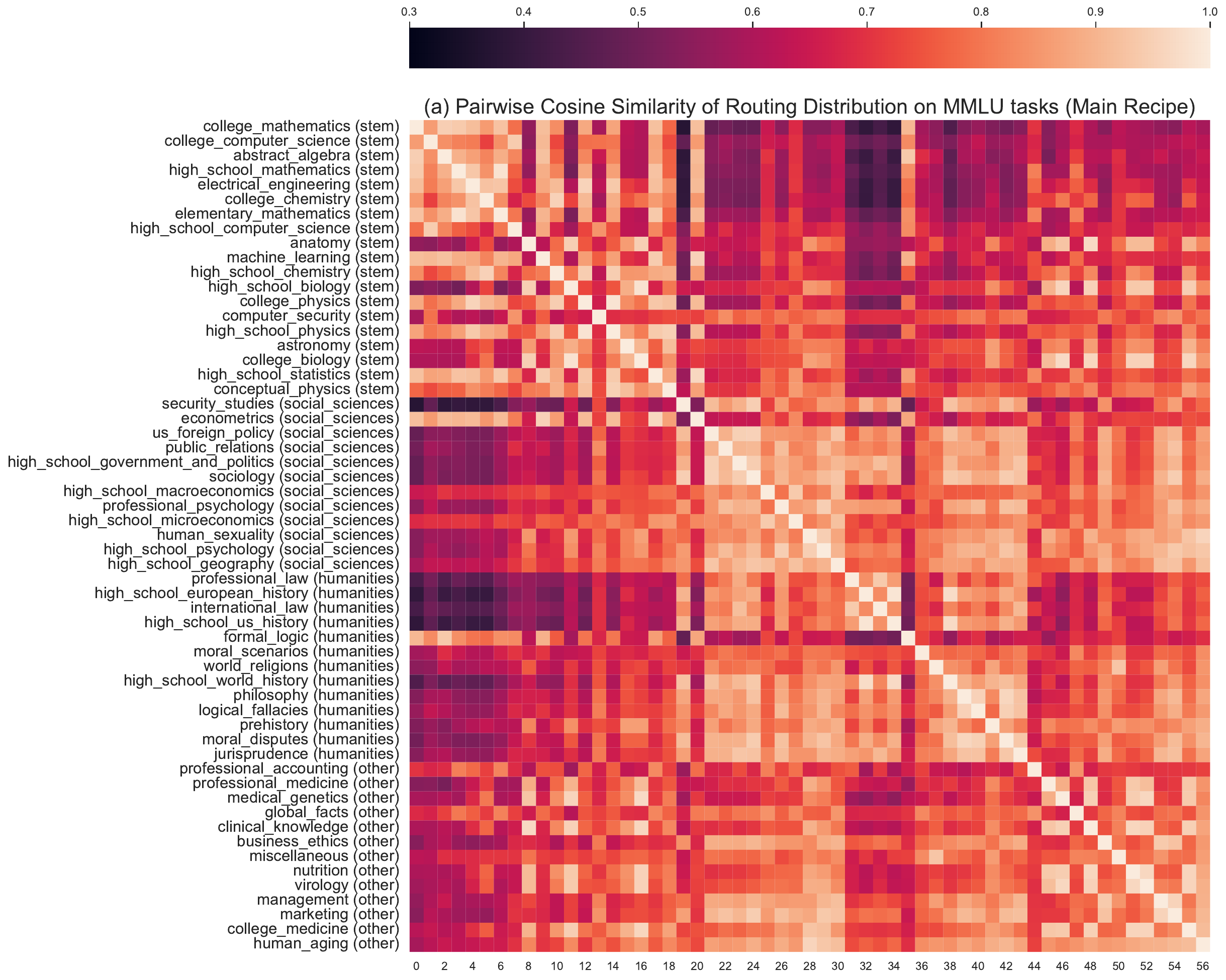}
    \caption{MoE Routing distribution similarity across MMLU 57 tasks for the main recipe. The darker color means lower similarity. X-axel and Y-axel have the same task order.}
    \label{fig:mmluheatmap-a}
\end{figure}

\smallsection{MoE routing on pretraining data}
To analyze the routing distribution on the pretraining dataset, 
we randomly select 2 million tokens from the pretraining dataset and visualize the expert loading distribution on them in Figure~\ref{fig:pretrainheatmap}.
First, we see that all layers for both recipes have reasonably balanced expert loading. 
Note that the maximum value in Figure~\ref{fig:pretrainheatmap} is 0.09, which is about 1.44 times of the perfectly balanced expert loading (0.0625). 
Then, relatively, we observe that the model trained with the main recipe is less balanced than the control recipe. 
%And for both visualization figures, deep layers are less balanced than the shallow layers.

\renewcommand{\thefigure}{6 (b)}
\begin{figure}
    \centering
    \includegraphics[width=1.0\linewidth]{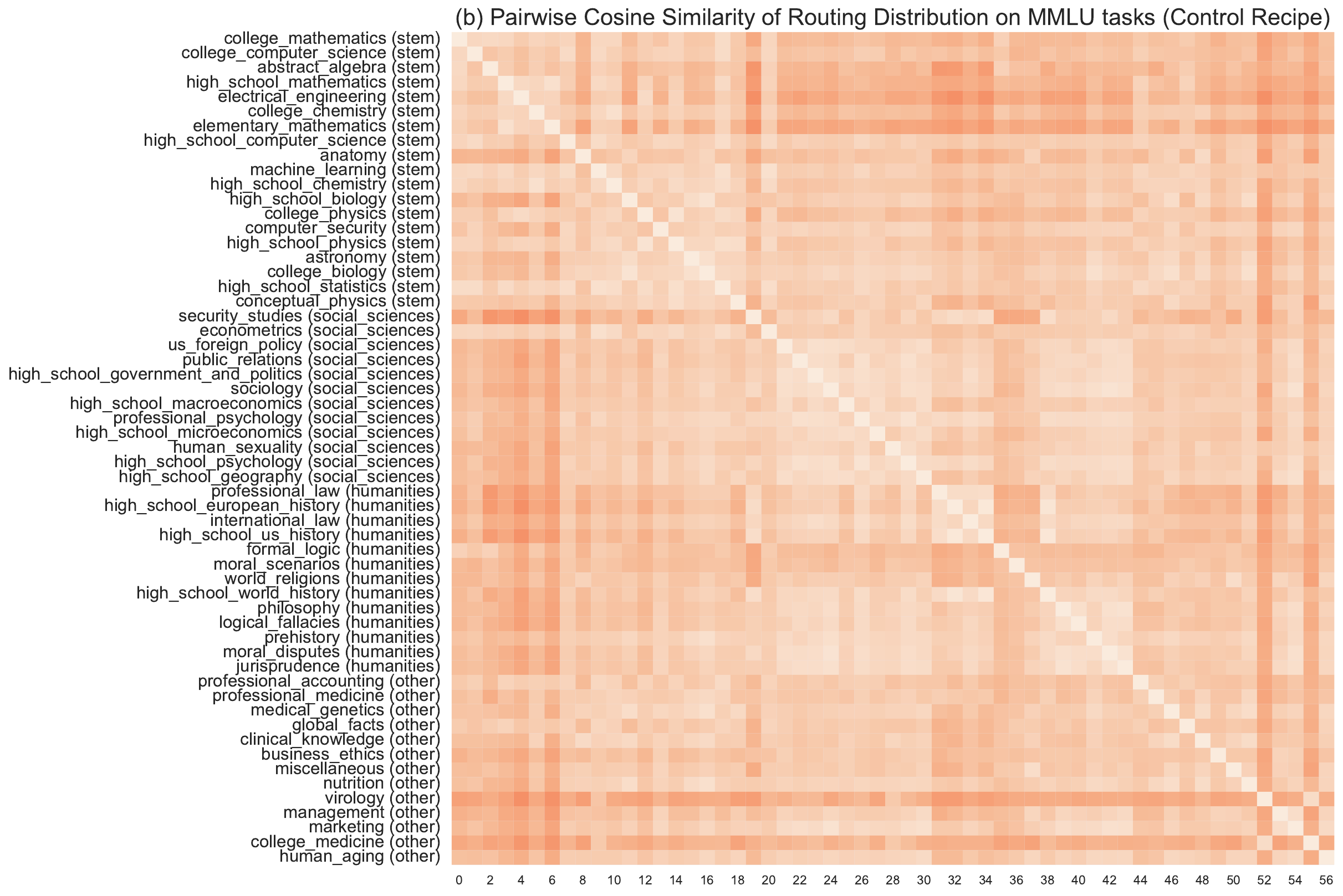}
    \caption{MoE Routing distribution similarity across MMLU 57 tasks for the control recipe. This figure shares the same setting and the same color bar with Figure~\ref{fig:mmluheatmap-a}.}
    \label{fig:mmluheatmap-b}
\end{figure}

\smallsection{MoE routing on different MMLU tasks} Next, we empirically verify whether different experts contain domain specific information. We first compare the routing distribution among the MMLU 57 tasks. For each task, we sample 24 prompts with 5-shot. The routing distribution for each task is a vector with 16 (experts per layer) $\times$ 32 (layers) dimensions (total number of experts). We then compute cosine similarity between the routing distribution of different tasks and visualize the similarities as a heatmap in Figure~\ref{fig:mmluheatmap-a} and \ref{fig:mmluheatmap-b}, where we group these 57 tasks into 4 categories based on their meta data. Note that the meta data we used here is provided in \cite{hendrycksmath2021},

The MoE trained with our main recipe is shown in Figure~\ref{fig:mmluheatmap-a}, we can see the \texttt{STEM} category has a clear boundary to \texttt{social\_science} and \texttt{humanities}. 
Additionally, it is quite reasonable for the two outliers in \texttt{social\_science} and \texttt{humanities}, (i.e., \texttt{econometrics} and \texttt{formal\_logic}), to have a higher similarity to the \texttt{STEM} category. 
This indicates that the routing distribution can vary significantly among different tasks. 
We also visualize the model trained with the control recipe (as in Figure~\ref{fig:mmluheatmap-b}), in which the routing distributions are similar across different tasks.

\renewcommand{\thefigure}{7}
\begin{figure}
    \centering
    % \vspace{-10pt}
    \includegraphics[width=0.9\linewidth]{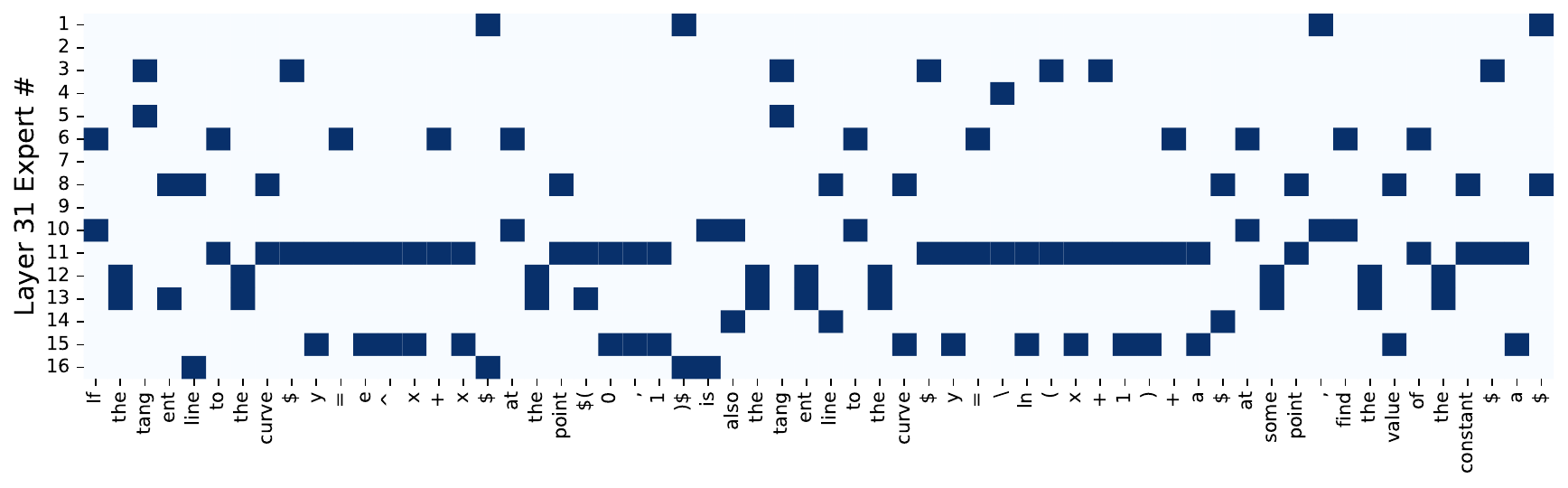}
    \includegraphics[width=0.9\linewidth]{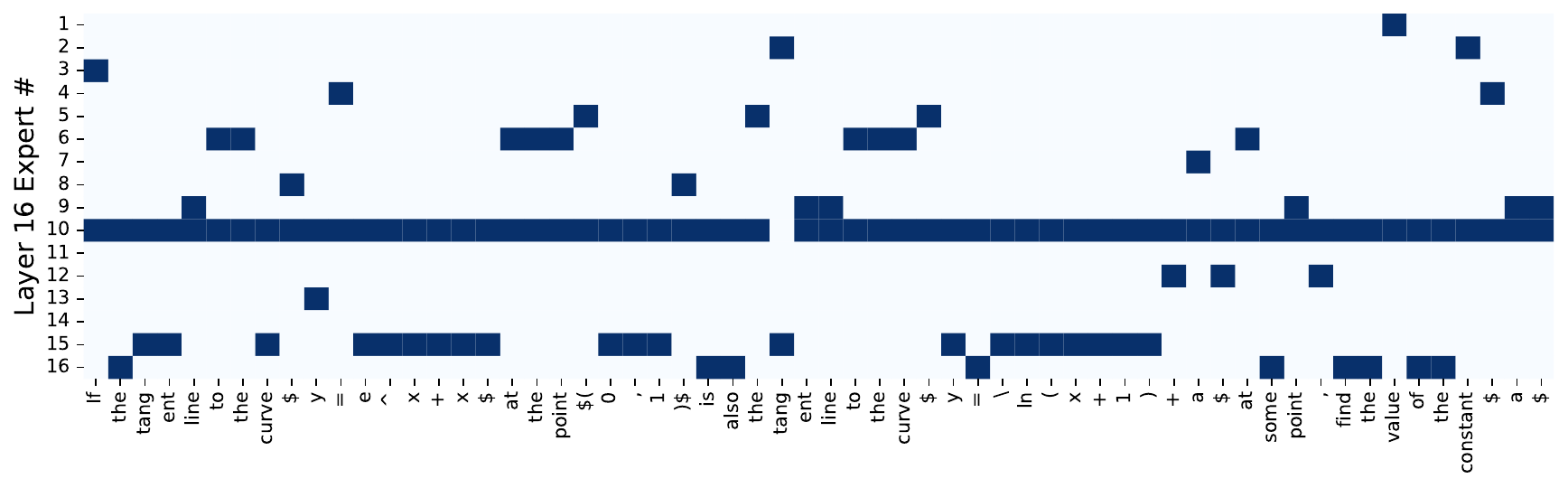}
    \includegraphics[width=0.9\linewidth]{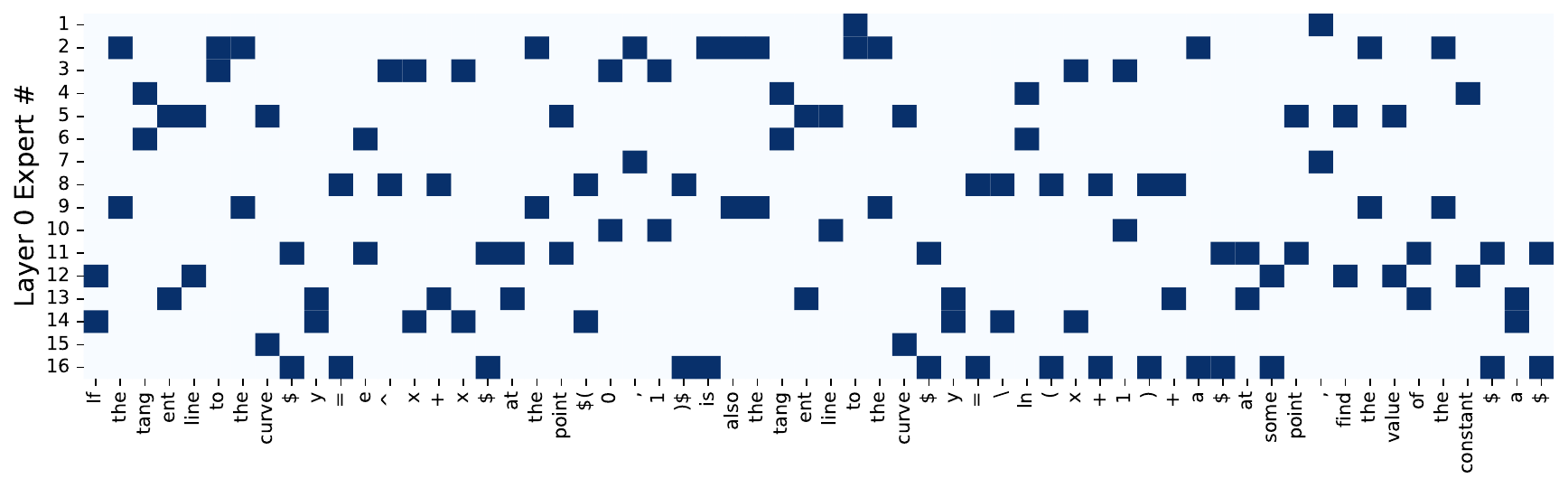}
    \caption{Case study of MoE trained by main recipe. It visualizes how experts are selected at different position in different layers. We select one question from newly released College Entrance Exam for the case study and more information is in Appendix~\ref{r13p}. }
    \label{fig:casestudy}
\end{figure}

\smallsection{A case study of MoE routing} This analysis uses the MoE model trained with the main recipe. As shown in Figure~\ref{fig:casestudy}, the routing distributions are different in different layers. The bottom (shallow) layer has the most balanced expert distribution. In the middle layer, the experts 10 and 15 are selected more other than the other experts. The final layer (deep) comes to be more balanced than the middle layer. These findings reveal that MoE routing distributions are related to such information as context, word, position, etc.

Our study seems to verify our hypothesis that 
%With these analyses, we believe the 
expert networks in \ourmodel have developed highly-specialized and heterogeneous expertise. 
As pointed out in ~\cite{wei2024skyworkmoedeepdivetraining}, such experts are likely to improve models' capacity.

\section{Conclusion}

In this paper, we describe in detail a new MoE model, known as \ourmodel, and the model training techniques (i.e., sparse backpropgationa and model parallelism configuration) used to train the model. Compared to dense models trained on the same pretraining corpus, \ourmodel demonstrates a remarkable scaling potential of MoE. We also provided a summary of our observations and insights gained during \ourmodel’s development, aiming to deepen our understanding of MoE training. Through controlled and semi-controlled experiments 
%on training techniques, 
we have demonstrated how gradient estimation methods and model parallelism strategies, along with corresponding auxiliary adaptations, significantly improves model training.

Looking ahead, many important open questions remain. For example, the training and inference of MoE models present challenges to both algorithms and engineering implementations. 
Also, since softmax is originally designed to approximate the argmax operation, it presents new challenges to approximate topk as sampling. 
We plan to further explore solutions to these challenges, with a focus on enhancing sparsity and developing efficient computing and scaling methods to advance state-of-the-art MoE modeling.

\section*{Acaknowledgement}

We would like to thank the engineering team at Microsoft for providing computation infrastructure support.
Also, we would like to thank
Mojan Javaheripi,
Suriya Gunasekar, and
S\'ebastien Bubeck
for the pre-training data.
Additionally, we would like to thank
Abhishek Goswami,
Alessandro Sordoni,
Amin Saied,
Ammar Ahmad Awan,
Anh Nguyen,
Benjamin Sorscher,
Caio César Teodoro Mendes,
Chandan Singh,
Colin White, 
Eric Xihui Lin,
Gustavo de Rosa,
Hanxiao Liu,
Joe Mayer,
Liliang Ren,
Marko Radmilac,
Michael Santacroce,
Michael Wyatt,
Olatunji Ruwase,
Philipp Witte,
Russell Hewett,
Swadheen Shukla,
Xia Song,
Yadong Lu,
Yang Liu,
Yuanzhi Li,
Yunan Zhang,
Zichong Li,
Zhen Zheng 
for their assistance and helpful discussions.

% \section{Contributions and Acaknowledgements}

% \todo{draft revision}

% \subsection{Pretraining and Posttraining}
% The pre-training and post-training of \ourmodel is performed by
% Liyuan Liu,
% Young Jin Kim,
% Shuohang Wang,
% Chen Liang,
% Yelong Shen,
% Masahiro Tanaka,
% Hao Cheng, 
% Zeqi Lin,  
% Chenruidong Zhang, 
% Xiaodong Liu,
% Vishrav Chaudhary,
% Hany Awadalla,
% Jianfeng Gao, and
% % Misha Bilenko, and
% Weizhu Chen.

% \subsection{Inference Optimization}
% The inference optimization of \ourmodel is performed by 
% Xiaoxia Wu, 
% Wenxiang Hu, and 
% Jilong Xue.

% \subsection{Phi3 Pretraining Data}
% The data used in the pre-training of \ourmodel comes from the phi3 study~\citep{abdin2024phi}. 
% Notably, we would like to thank
% Mojan Javaheripi,
% Suriya Gunasekar, and
% Sebastien Bubeck
% for sharing the data and helping us on related configurations. 

% \subsection{Other Acknowledgements}
% We would like to thank the engineering team at Microsoft for providing computation infrastructure support, 
% Also, we would like to thank
% Abhishek Goswami,
% Alessandro Sordoni,
% Amin Saied,
% Ammar Ahmad Awan,
% Anh Nguyen,
% Benjamin Sorscher,
% Caio César Teodoro Mendes,
% Eric Xihui Lin,
% Gustavo de Rosa,
% Hanxiao Liu,
% Joe Mayer,
% Liliang Ren,
% Marko Radmilac,
% Michael Santacroce,
% Michael Wyatt,
% Olatunji Ruwase,
% Philipp Witte,
% Russell Hewett,
% Sebastien Bubeck,
% Xia Song,
% Yadong Lu,
% Yang Liu,
% Yuanzhi Li,
% Yunan Zhang,
% Zichong Li,
% Zhen Zheng 
% for their helpful discussions (the name list is organized in the alphabetical order). 

\bibliography{icml, mainbib}
\bibliographystyle{icml}

\clearpage
\appendix

\renewcommand \partname{Appendix}

% \addcontentsline{toc}{section}{\sm and Setting}
\part{\sm and Setting} 
\parttoc

% 1. delta 1 -> 2 * 
% 2. delta 0 -> 
%     2.a B = 1 => 2 * 
%     2.b B = 0 => 2 * 

% \begin{align}
% 6 \cdot \frac{f(\frac{1}{3} \vp_\mD \cdot Expert(vx, \vw_\mD))}{\partial \vz} \\
% = 2 \cdot \frac{\partial f(\frac{1}{3} \cdot \vp_\mD \cdot Expert(\vx, \vw_\mD))}{\partial \frac{1}{3} \cdot \vp_\mD \cdot Expert(\vx, \vw_\mD)} \cdot Expert(\vx, \vw_i) \cdot \frac{\partial \vp}{\partial \vz} \\
% = 2 \cdot \frac{\partial f(\frac{1}{3} \cdot \vp_\mD \cdot Expert(\vx, \vw_\mD))}{\partial \frac{1}{3} \cdot \vp_\mD \cdot Expert(\vx, \vw_\mD)} \cdot  \frac{\partial Expert(\vx, \vw_i) \cdot \vp}{\partial \vz} \\
% = 2 \cdot f'(\frac{1}{3} \cdot \vp_\mD \cdot Expert(\vx, \vw_\mD)) \cdot \frac{\partial Expert(\vx, \vw_i) \cdot \vp}{\partial \vz} 
% \end{align}

\section{\sm}
\label{appendix:sparsemixer}

\sm has two important components, i.e., approximating \texttt{TopK} with discrete variable sampling, and estimating gradients in a scalable manner. 
Here, we first introduce \sm for \texttt{Top1} MoE, then extend it for \texttt{TopK} MoE. 

\subsection{Discrete Variable Sampling for Approximating \texttt{Top1}}
In conventional MoE training, it is a standard practice to add jitter noise to expert routing, which would lead to implicit expert sampling. 
With \sm, we try to replace the implicit expert sampling with explicit expert sampling. 

We start from the special case of $K=1$. Particularly, we will approximate $\texttt{Top1}(\vz)$ as sampling from $\texttt{MaskedSoftmax}(\vz)$:
\begin{eqnarray}
    \texttt{MaskedSoftmax}(\vz)_i = \frac{\exp(\vz_i) \cdot \delta_i}{\sum_j \exp(\vz_j) \cdot \delta_j},
    \label{eqn:masked_softmax}
\end{eqnarray}
where $\delta_i = \delta\big(\vz^* - \vz_i \leq r \cdot (|\vz_i| + |\vz^*|) \big)$, $\vz^* = \max_k \vz_k$, and $r$ is a hyper-parameter.
Comparing to the vanilla softmax sampling, $\texttt{MaskedSoftmax}(\cdot)$ introduces a hyper-parameter $r$ to control the magnitude of randomness and sparsity of the sampling space. 
Similar to \cite{fedus2022switch} and \cite{Liu2023SparseMixer}, we observed that the vanilla softmax sampling doesn't work well in practice. 
In our experiments, $\texttt{MaskedSoftmax}(\cdot)$ yields similar performance with conventional jitter noise in MoE training.

\subsection{\texttt{Top1} \sm MoE}

For expert routing gradient estimation, we proposed two slightly different variants, SparseMixer-v2* and \sm. 
Let us first present SparseMixer-v2* and then move to introducing \sm. 

\smallsection{SparseMixer-v2*}
% With $\texttt{MaskedSoftmax}(\cdot)$, 
We configure the MoE layer for SparseMixer-v2* as: 
\begin{align}
    \sum_{i=0}^{n-1} \texttt{MaskedSoftmax}(\vz)_i \cdot \mD_i \cdot Expert(\vx, \vw_i), \mbox{ where } \mD \sim \texttt{MaskedSoftmax}(\vz). 
    \label{eqn:mixer-layer}
\end{align}
While the SparseMixer-v2* MoE layer will behave as Equation~\ref{eqn:mixer-layer} during inference, it will behave as in Algorithm~\ref{algo:top1_sparsemixer} during training. 
Now lets proceed to the underlying derivations of Algorithm~\ref{algo:top1_sparsemixer}. 
Referring other parts of the network (including the loss function) as $f$, the training objective is:
\begin{align}
   \mathcal{L} 
   &= E_{\mD \sim \texttt{MaskedSoftmax}(\vz)} \left[f\left(\sum_{i=0}^{n-1} \texttt{MaskedSoftmax}(\vz)_i \cdot \mD_i \cdot Expert(\vx, \vw_i)\right) \right]  \nonumber\\
   &= \sum_{i=0}^{n-1} f\big(\texttt{MaskedSoftmax}(\vz)_i \cdot Expert(\vx, \vw_i)\big) \cdot \texttt{MaskedSoftmax}(\vz)_i. \label{eqn:object}
\end{align}

For simplicity, we mark $\texttt{MaskedSoftmax}(\vz)$ as $\vp$. 
Then the gradient of $\vz$ is:
\begin{align}
    \nabla \vz &= \sum_{i=0}^{n-1} \vp_i \cdot \frac{\partial f\big(\vp_i \cdot Expert(\vx, \vw_i)\big)}{\partial \vz} + f\big(\vp_i \cdot Expert(\vx, \vw_i)\big) \cdot \frac{\partial \vp_i}{\partial \vz} \nonumber \\
    &= \sum_{i=0}^{n-1} \vp_i \cdot \frac{\partial f\big(\vp_i \cdot Expert(\vx, \vw_i)\big)}{\partial \vz} + \Big( f\big(\vp_i \cdot Expert(\vx, \vw_i)\big) - f(0) \Big) \cdot \frac{\partial \vp_i}{\partial \vz}. \label{eqn:baseline-subtract}
\end{align}
Note that Equation~\ref{eqn:baseline-subtract} is usually known as baseline subtraction\footnote{since $\sum_i \vp_i = 1$, we have $0 = \frac{\partial \sum_i \vp_i}{ \partial \vz}$. }.
In the ODE literature, there are many ways to approximate $f\big(\vp_i \cdot Expert(\vx, \vw_i)\big) - f(0)$. 
Here, we focus on two approximations\footnote{For simplicity, we use $f'(\vh)$ to refer to $\frac{\partial f(\vh)}{\partial \vh}$.}:
\begin{itemize}[leftmargin=*]
\item 
Euler's method: a first-order ODE solver that approximates $f\big(\vp_i \cdot Expert(\vx, \vw_i)\big) - f(0)$ as $f'\big(\vp_i \cdot Expert(\vx, \vw_i)\big) \cdot \vp_i \cdot Expert(\vx, \vw_i)$.

\item
Heun's third-order method: a third-order ODE solver that approximates $f\big(\vp_i \cdot Expert(\vx, \vw_i)\big) - f(0)$ as {\footnotesize $\Big(\frac{1}{4}\cdot f'\big(\vp_i \cdot Expert(\vx, \vw_i)\big) + \frac{3}{4} \cdot f'(\frac{\vp_i \cdot Expert(\vx, \vw_i)}{3})\Big)\cdot \vp_i \cdot Expert(\vx, \vw_i)$}.
\end{itemize}

Correspondingly, we approximate $\nabla$ in two ways. 
Following Euler's method, we have:
\begin{align*}
    \nabla_{\mbox{\scriptsize 1st }} \vz &= \sum_{i=0}^{n-1}  \left( \vp_i \cdot \frac{\partial f\big(\vp_i \cdot Expert(\vx, \vw_i)\big)}{\partial \vz} + f'\big(\vp_i Expert(\vx, \vw_i)\big) \cdot \vp_i Expert(\vx, \vw_i) \cdot \frac{\partial \vp_i}{\partial \vz} \right)  \\
    &= \sum_{i=0}^{n-1} \left( \vp_i \cdot \frac{\partial f\big(\vp_i \cdot Expert(\vx, \vw_i)\big)}{\partial \vz} + \vp_i \cdot \frac{\partial f\big(\vp_i Expert(\vx, \vw_i)\big)}{\partial \vp_i Expert(\vx, \vw_i)} \frac{ \partial \vp_i Expert(\vx, \vw_i)}{\partial \vp_i}  \frac{\partial \vp_i}{\partial \vz} \right)\\
    &= \sum_{i=0}^{n-1} 2 \cdot \vp_i \cdot \frac{\partial f\big(\vp_i \cdot Expert(\vx, \vw_i)\big)}{\partial \vz} \\
    &= E_{\mD \sim \texttt{MaskedSoftmax}(\vz)} \left[ 2 \cdot \frac{\partial f\big(\vp_\mD \cdot Expert(\vx, \vw_\mD)\big)}{\partial \vz} \right].
\end{align*}

Following Heun's third-order method, we have:
% \begin{align*}
% \nabla_{\mbox{\scriptsize 3rd }} \vz =& \sum_{i=0}^{n-1}
% \vp_i \cdot \frac{\partial f\big(\vp_i \cdot Expert(\vx, \vw_i)\big)}{\partial \vz} + \left( \frac{1}{4}\cdot f'\big(\vp_i \cdot Expert(\vx, \vw_i)\big) + \right. \\
% & \left. \frac{3}{4} \cdot f'\big(\frac{\vp_i \cdot Expert(\vx, \vw_i)}{3}\big) \right) \cdot \vp_i \cdot Expert(\vx, \vw_i) \cdot \frac{\partial \vp_i}{\partial \vz} \\
% =& \sum_{i=0}^{n-1} \frac{5}{4} \cdot \vp_i \cdot \frac{\partial f\big(\vp_i \cdot Expert(\vx, \vw_i)\big)}{\partial \vz} + \frac{9}{4} \cdot \vp_i \cdot \frac{\partial f(\frac{\vp_i \cdot Expert(\vx, \vw_i)}{3})}{\partial \vz}\\
% =& E_{\mD \sim \texttt{MaskedSoftmax}(\vz), \mB \sim \texttt{Bernoulli}(\frac{5}{14})} \left[ \frac{7}{2} \cdot  \frac{\partial f\big(\frac{1 + 2\mB}{3}\cdot \vp_\mD \cdot Expert(\vx, \vw_\mD)\big)}{\partial \vz} \right].
% \end{align*}
\begin{align*}
\nabla_{\mbox{\scriptsize 3rd }} \vz =& \sum_{i=0}^{n-1} \Biggl(
\vp_i \cdot \frac{\partial f\big(\vp_i \cdot Expert(\vx, \vw_i)\big)}{\partial \vz} + \left( \frac{1}{4}\cdot f'\big(\vp_i \cdot Expert(\vx, \vw_i)\big) + \right. \\
& \left. \frac{3}{4} \cdot f'\big(\frac{\vp_i \cdot Expert(\vx, \vw_i)}{3}\big) \right) \cdot \vp_i \cdot Expert(\vx, \vw_i) \cdot \frac{\partial \vp_i}{\partial \vz} \Biggr)\\
=& \sum_{i=0}^{n-1} \big(\frac{5}{4} \cdot \vp_i \cdot \frac{\partial f\big(\vp_i \cdot Expert(\vx, \vw_i)\big)}{\partial \vz} + \frac{9}{4} \cdot \vp_i \cdot \frac{\partial f(\frac{\vp_i \cdot Expert(\vx, \vw_i)}{3})}{\partial \vz}\big)\\
=& E_{\mD \sim \texttt{MaskedSoftmax}(\vz), \mB \sim \texttt{Bernoulli}(\frac{5}{8})} \left[ (6 - 4\mB) \cdot  \frac{\partial f\big(\frac{1 + 2\mB}{3}\cdot \vp_\mD \cdot Expert(\vx, \vw_\mD)\big)}{\partial \vz} \right].
\end{align*}

Lastly, as discussed in \cite{Liu2023SparseMixer}, we combine $\nabla_{\mbox{\scriptsize 3rd }}$ and $\nabla_{\mbox{\scriptsize 1st }}$  to balance router learning and expert learning by using $\nabla_{\mbox{\scriptsize 1st }}$ if $\mD = \arg\max(\vz)$ or $\nabla_{\mbox{\scriptsize 3rd }}$ otherwise.
Particularly, using $\delta_\mD$ to refers to $\delta(\mD = \arg\max(\vz))$, we have:
{
\scriptsize
\begin{align}
    \nabla_{\mbox{\tiny SparseMixer-v2* }} \vz  = E_{\mD \sim \texttt{\tiny MaskedSoftmax}(\vz)} [\nabla_{\mD, \mbox{\tiny SparseMixer-v2* }} \vz], \nonumber
\end{align}
\begin{align}  
& \mbox{where}\,\, \nabla_{\mD, \mbox{\tiny SparseMixer-v2* }} \vz & \nonumber\\
= & E_{\mB \sim \texttt{\tiny Bernoulli}(\frac{5}{8})} \biggl[(1 - \delta_\mD) \cdot  (6 - 4\mB) \cdot  \frac{\partial f(\frac{1 + 2\mB}{3}\cdot \vp_\mD \cdot Expert(\vx, \vw_\mD))}{\partial \vz} \biggr] + \delta_\mD \cdot 2 \cdot \frac{\partial f(\vp_\mD \cdot Expert(\vx, \vw_\mD))}{\partial \vz}  & \nonumber\\
= & E_{\mB \sim \texttt{\tiny Bernoulli}(\frac{5}{8})} \left[ 
        \big(6-4\cdot \max(\mB, \delta_\mD)\big) \frac{\partial f\big(\frac{1 + 2\cdot \max(\mB, \delta_\mD)}{3} \cdot \vp_\mD\cdot Expert(\vx, \vw_\mD)\big)}{\partial \vz}
    \right] & \nonumber \\
= & E_{\mB \sim \texttt{\tiny Bernoulli}(\frac{5}{8})} \left[
    2 \cdot f'\big(\frac{1 + 2 \max(\mB, \delta_\mD)}{3} \cdot  \vp_\mD \cdot Expert(\vx, \vw_\mD)\big)  \frac{\partial \vp_\mD \cdot Expert(\vx, \vw_\mD)}{\partial \vz}
    \right]. &
    \label{eqn:sparsemixer-final}
\end{align}
}

\smallsection{\sm}
During the development of \ourmodel, an error was made in the derivation and there is a discrepancy between Algorithm~\ref{algo:top1_realsparsemixer} and Equation~\ref{eqn:sparsemixer-final}. 
Particularly, instead of Equation~\ref{eqn:object}, we used Equation~\ref{eqn:realobject} as the objective:  
\begin{align}
\hat{\mathcal{L}} = E_{\mD \sim \texttt{MaskedSoftmax}(\vz)} \left[f\left(\sum_{i=0}^{n-1} \texttt{detach}\big(\texttt{MaskedSoftmax}(\vz)_i\big) \cdot \mD_i \cdot Expert(\vx, \vw_i)\right) \right].  \label{eqn:realobject}
\end{align}

With similar derivations as before, the resulting estimator is as in Equation~\ref{eqn:realsparsemixer-final} and we name this estimator as SparseMixer-v2. 
\begin{align}  
& \hat{\nabla}_{\mD, \mbox{\tiny \sm }} \vz \nonumber\\
= & E_{\mB \sim \texttt{\tiny Bernoulli}(\frac{1}{4})} \left[
    f'\big(\frac{1 + 2 \max(\mB, \delta_\mD)}{3} \cdot  \vp_\mD \cdot Expert(\vx, \vw_\mD)\big)  \frac{\partial \vp_\mD \cdot Expert(\vx, \vw_\mD)}{\partial \vz}
    \right].
    \label{eqn:realsparsemixer-final}
\end{align}

Due to resource constraints, we only did preliminary empirical analyses on the impact of this discrepancy between Equation~\ref{eqn:realobject} and Equation~\ref{eqn:object}, and could not conduct the pretraining from scratch again.
In our empirical analyses, we observed that this discrepancy has only limited impacts on the performance and we did not observe evidence indicating that the discrepancy would lead to a significant performance degeneration. 

\renewcommand{\thefigure}{8}
\begin{figure}
    \centering
    \vspace{-10pt}
    \includegraphics[width=0.8\linewidth]{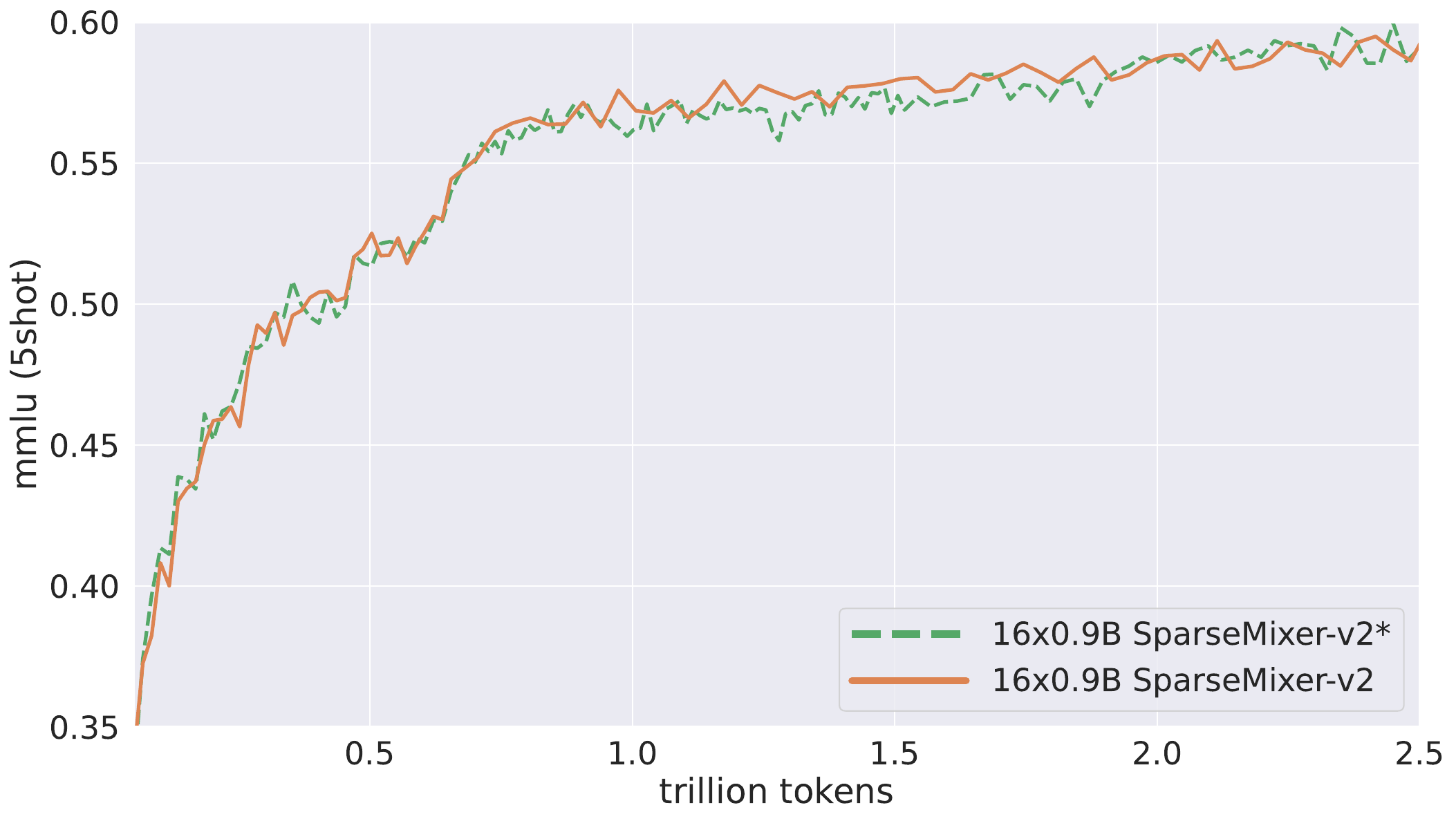}
    \caption{Controlled Comparisons of \sm and v2* on 16$\times$0.9B MoE.}
    \label{fig:msr-controlled-star}
\end{figure}

Particularly, we followed the experiment setting as in Figure~\ref{fig:msr-controlled} and trained SparseMixer-v2* for 2.5T tokens. 
As in Figure~\ref{fig:msr-controlled-star}, \sm and SparseMixer-v2* achieve comparable performance. 
It is worth mentioning that there is an additional scaling factor in Equation~\ref{eqn:sparsemixer-final} (i.e., $2$). 
In our experiments, we observed that adding a temperature of $2$ at the  $\texttt{MaskedSoftmax}$ helps to stabilize the training, which would cancel the scaling factor in Equation~\ref{eqn:sparsemixer-final} and ensure a consistent gradient magnitude (as in line 1 of Algorithm~\ref{algo:top1_sparsemixer}).  

% \smallsection{Comparisons to GShard}
% Comparing \sm to GShard, \sm mainly differs in two ways:

% \begin{itemize}[leftmargin=*]
% \item While GShard uses jitter-noise based implicit sampling, both \sm and SparseMixer-v2* uses explicit masked softmax-based sampling. We did some preliminary experiments and found that sampling strategy has little impacts on downstream performance. 

% \item GShard uses gating as a proxy for computing the routing gradients, while \sm and SparseMixer-v2* estimates routing gradients directly. 

% \end{itemize}

\subsection{Extension to \texttt{TopK}}

\begin{algorithm}[t]
\DontPrintSemicolon
\KwIn{Router Output ($\vz$), Expert Input ($\vx$), and Expert Weights ($\{\vw_i\}$)}
\KwOut{MoE Output ($\vy$) and Expert Assignment ($\mD$)}
$\vp \gets \texttt{MaskedSoftmax}(\vz)$ \,\,\,\,
\tcc{$\texttt{MaskedSoftmax}()$ is defined in Equation~\ref{eqn:masked_softmax}. }
Sample $\mD$ from $\vp$ \;
$\vh \gets \mbox{Expert}(\vx, \vw_\mD) \cdot \vp_\mD$ \;
$\delta_\mD \gets 1$ if $\mD = \arg\max(\vz)$ else 0 \; 
Sample $\mB$ from $\texttt{Bernoulli}(\frac{1}{4})$ \,\,\,\,\,\,\,\,\,\,\,\,\,\,\,\,\,\,\,\,\,\,\,\,\,\,\,\,\,\,\,\,\,\,\,\,\,\,\,\,\,\,\,\,\,\,\,\,\,\,\,\,\,\,\,\,\,\,\,\,\,\,\,\,\,\,\,\,\,\,\,\,\,\,\,\,\tcc{$P(\mB = 1) = \frac{1}{4}$. }
\tcc{$\mbox{detach}(\cdot)$ returns a duplicate, detached from the existing graph. }
$\vy \gets \vh + \texttt{detach}\big(\max(\delta_\mD, \frac{1 + 2 \cdot \mB}{3})\cdot \vh - \vh \big) $ \;
\Return{$\vy$, $\mD$}
\caption{\texttt{Top1} \sm MoE Layer in Training. }
\label{algo:top1_realsparsemixer}
\end{algorithm}

\begin{algorithm}[t]
\DontPrintSemicolon
\KwIn{Router Output ($\vz$), Expert Input ($\vx$), and Expert Weights ($\{\vw_i\}$)}
\KwOut{MoE Output ($\vy$) and Expert Assignment ($\mD$)}
$\vp \gets \texttt{MaskedSoftmax}(\frac{\vz}{2})$ \,\,\,\,
\tcc{$\texttt{MaskedSoftmax}()$ is defined in Equation~\ref{eqn:masked_softmax}. }
Sample $\mD$ from $\vp$ \;
$\vh \gets \mbox{Expert}(\vx, \vw_\mD) \cdot \vp_\mD$ \;
$\delta_\mD \gets 1$ if $\mD = \arg\max(\vz)$ else 0 \; 
Sample $\mB$ from $\texttt{Bernoulli}(\frac{5}{8})$ \,\,\,\,\,\,\,\,\,\,\,\,\,\,\,\,\,\,\,\,\,\,\,\,\,\,\,\,\,\,\,\,\,\,\,\,\,\,\,\,\,\,\,\,\,\,\,\,\,\,\,\,\,\,\,\,\,\,\,\,\,\,\,\,\,\,\,\,\,\,\,\,\,\,\,\,\tcc{$P(\mB = 1) = \frac{5}{8}$. }
\tcc{$\mbox{detach}(\cdot)$ returns a duplicate, detached from the existing graph. }
$\vy \gets 2 \cdot \vh + \texttt{detach}\big(\max(\delta_\mD, \frac{1 + 2 \cdot \mB}{3})\cdot \vh - 2 \cdot \vh \big) $ \;
\Return{$\vy$, $\mD$}
\caption{\texttt{Top1} SparseMixer-v2* MoE Layer in Training. }
\label{algo:top1_sparsemixer}
\end{algorithm}
 
\begin{algorithm}[t]
\DontPrintSemicolon
\KwIn{Router Output ($\vz$), Expert Input ($\vx$), Activate Expert Number per Token ($K$), and Expert Weights ($\{\vw_i\}$)}
\KwOut{MoE Output ($\vy$)}
\For{$k \in \{1, \cdots, K\}$} {
    $\vy_k, \mD \gets \texttt{\sm/v2* Top1}(\vz, \vx, \{\vw_i\})$ as in Algorithm~\ref{algo:top1_realsparsemixer}/\ref{algo:top1_sparsemixer} \;
    $\vz_{\mD} \gets -\infty$ \,\,\,\,\,\,\,\tcc{Mask out expert $\mD$ in following expert samplings. }
}
$\vy \gets \sum_{k=1}^K \vy_k$ \;
\Return{$\vy$}
\caption{\texttt{TopK} \sm/v2* MoE Layer in Training. }
\label{algo:topk_sparsemixer}
\end{algorithm}

As in Algorithm~\ref{algo:topk_sparsemixer}, we further extend the \sm \texttt{Top1} algorithm to \texttt{TopK} by sampling without replacements. 

\subsection{Difference of GShard, SparseMixer, and \sm}

\smallsection{GShard v.s. \sm}
Comparing GShard and \sm, they differs in mostly two ways: 
\begin{itemize}[leftmargin=*]
\item GShard and \sm behave differently in inference, since \sm replaces the \texttt{TopK} function with the \texttt{MaskedSoftmax} for sampling and gating. 

\item \texttt{TopK} and \sm are trained differently, i.e., GShard treats gating as a proxy for the router gradient and \sm estimates the router gradient directly.  
\end{itemize}

In our experiments, we observed the architecture change is very important for GShard and \sm performing differently. 
We conduct an experiment that only replaces the \texttt{TopK} function with the \texttt{MaskedSoftmax} for sampling but not gating, and estimates gradients like \sm. 
The performance of the resulting method is almost identical to GShard, which may due to the gating gradients overshadowed the routing gradients.

\smallsection{\sm v.s. SparseMixer}
\label{subsec:sparsemixer-compare}
Comparing to SparseMixer as in \cite{Liu2023SparseMixer}, the \sm and SparseMixer-v2* that used in this study mainly differs in three ways:

\begin{itemize}[leftmargin=*]
\item While \cite{Liu2023SparseMixer} uses the mid-point method (a second-order ODE method), we uses Heun's third-order method for approximating $f(\vp_i \cdot Expert(\vx, \vw_i)) - f(0)$ here. 

\item While \cite{Liu2023SparseMixer} introduces additional parameters $\omega$ to scale the MoE layer output for training stabilization, we didn't introduce such parameters in our study. Instead, we find that model training can be successfully stabilized by initializing the network properly ~\citep{Glorot2010UnderstandingTD,He2015DelvingDI,Liu2020UnderstandingTD,Yang2022TensorPV}.

\item SparseMixer and SparseMixer-v2* use Equation~\ref{eqn:object} as the objective function, and \sm uses Equation~\ref{eqn:realobject} as the objective function. 

\end{itemize}

% \begin{table}[t]
% \caption{Comparing GShard and \sm.}
% \begin{tabularx}{\linewidth}{r|Y |Y}
% \toprule
%  & Gating & Routing \\ \midrule
% GShard & \texttt{Softmax}($\cdot$) & \texttt{TopK}($\cdot$).detach() \\ \midrule
% \sm 1st & \texttt{MaskedSoftmax}($\cdot$) & \texttt{MaskedSoftmax}($\cdot$).detach() \\ \midrule
% \sm 1st & \texttt{MaskedSoftmax}($\cdot$).detach() & \texttt{MaskedSoftmax}($\cdot$) w. $\nabla_{\mbox{\scriptsize 1st }}$ \\ \midrule
% \sm & \texttt{MaskedSoftmax}($\cdot$).detach() & \texttt{MaskedSoftmax}($\cdot$) w. $\nabla_{\mbox{\scriptsize v2 }}$ \\ \midrule
% SparseMixer-v2* & \texttt{MaskedSoftmax}($\cdot$) & \texttt{MaskedSoftmax}($\cdot$) w. $\nabla_{\mbox{\scriptsize v2* }}$  \\
% \bottomrule
% \end{tabularx}
% \label{tbl:gshard}
% \end{table}
\section{Experiment Setting Details}
\label{appendix:hardware}

\smallsection{Baselines in Table~\ref{table:bbeval}}
We compare our MoE with existing open-sourced models, which have a similar number of active parameters. 
It includes MoE model, Mixtral~\citep{jiang2024mixtral}, and dense models, such as Phi-3 7b/14b~\citep{abdin2024phi}, Mistral 7b~\citep{jiang2023mistral}, Gemma 8b~\citep{team2024gemma}, Llama3 8b/70b, and also the popular LLM APIs, such as GPT-3.5, GPT-4o, Gemini-Pro-1.5-Flash.

\smallsection{Benchmarks in Table~\ref{table:bbeval}}
Our evaluation is same as Phi-3~\citep{abdin2024phi}.
For a fair comparison, we evaluate all the models on popular benchmarks under a same setting. Thus, the prompts and fewshot numbers could be different from the papers introducing the models. 
Our benchmarks include MMLU 5-Shot~\citep{hendrycksmath2021},     
HellaSwag 5-Shot~\citep{zellers2019hellaswag},
ANLI 7-Shot\citep{nie2020adversarial},
GSM-8K 8-Shot CoT\citep{cobbe2021training}, 
MATH 0-Shot CoT~\citep{hendrycksmath2021},
MedQA 2-Shot,\citep{jin2020disease}, 
AGIEval 0-Shot \citep{zhong2023agieval}, 
TriviaQA 5-Shot\citep{joshi2017triviaqa},
Arc-C 10-Shot \citep{clark2018think},
Arc-E 10-Shot \citep{clark2018think}, 
PIQA 5-Shot\citep{bisk2019piqa}, 
SociQA 5-Shot \citep{bisk2019piqa}, 
BigBench-Hard 3-Shot CoT \citep{srivastava2022beyond,suzgun2022challenging},
WinoGrande 5-Shot \citep{sakaguchi2019winogrande},
OpenBookQA 10-Shot \citep{mihaylov2018suit}, 
BoolQ 2-Shot, \citep{clark2019boolq}, 
CommonSenseQA 10-Shot, \citep{talmor2019commonsenseqa}, 
TruthfulQA 10-Shot MC2 \citep{lin2022truthfulqa},
HumanEval 0-Shot \citep{chen2021evaluating} ,
MBPP 3-Shot \citep{austin2021program}, 
Our evaluation on benchmark dataset is shown in Table~\ref{table:bbeval}.
It is worth mentioning that, we use $^*$ in Table~\ref{table:bbeval} to indicate that the average score of 16 sub-tasks of BigBench-Hard are reported for GPT-4o (for the rest of other models, we report the average of 27 sub-tasks).

\smallsection{Hardware Information for Throughput Study}
We use the same 8 nodes to conduct the throughput study in Section~\ref{subsec:computational_scaling}.
Each node has 8 GPUs (H100 with 80GB GPU memory), 200 vCPU cores (2.0~3.8GHz), and 2TB system memory.
Also, 8$\times$400Gbps (3200Gbps cumulative) Infiniband is used for node-to-node networking.

\smallsection{Figure~\ref{fig:mmlu} Setting}
In Figure~\ref{fig:mmlu}, we listed the MMLU scores from our evaluation pipelines for \ourmodel, Mixtal 8$\times$7B, Mixtal 8$\times$22B, LLaMA3 8B, Gemma 7B, Phi-3-mini, Phi-3-small, and Phi-3-medium. 
For all other models, we used the reported MMLU scores from their studies respectively.

\smallsection{Translated 2024 GAOKAO Math-1}
We used the internal Microsoft GenAI playground to collect feedbacks from \ourmodel and other Phi-3 models. 
As to other models, their responses are collected from similar web portals.

\part{2024 GAOKAO Math-1} 
\hypertarget{part:appendix2}{}
% \addcontentsline{toc}{section}{2024 GAOKAO Math-1}
\parttoc
\section{Translated 2024 GAOKAO Math-1}
\label{appendix:gaokao}

We listed the responses from \ourmodel to the translated 2024 Chinese University Entrance Exam Math-1 questions in Table~\ref{tbl:gaokao}.

\begin{table}[t]
\caption{2024 Translated Chinese University Entrance Exam Math-1.}

\resizebox{1.\linewidth}{!}{
\begin{tabular}{cccccccccc}
\toprule

& GPT & Claude3  & Gemini & \multicolumn{2}{c}{Llama3}  & GRIN & Phi-3.5 & \multicolumn{2}{c}{Phi-3} \\
& 4o  & Opus     & Ultra-1.0 & 8B & 70b & MoE & MoE & Medium & Small \\
\midrule
% \midrule
\multicolumn{9}{l}{Single-Choice Questions (Each question worth 5 points)} \\
\midrule
Q1 (A) & A & A & A & C & A & A & D & D & A \\
Q2 (C) & C & C & A & C & NaN & C & C & C & A \\
Q3 (D) & D & D & D & D & D & D & D & D & D \\
Q4 (A) & A & A & A & D & NaN & A & NaN & NaN & C \\
Q5 (B) & B & B & B & C & B & B & B & A & B \\
Q6 (B) & B & B & NaN & A & A & D & B & A & A \\
Q7 (C) & C & A & C & B & C & C & C & C & D \\
Q8 (B) & C & B & B & NaN & B & A & C & B & C \\
Score & 35 & 35 & 30 & 10 & 25 & 30 & 25 & 20 & 15 \\
\midrule 
% \midrule 
\multicolumn{9}{l}{Multi-Choice Questions (Each question worth 6 points. Partial credit is given for partially correct} \\
\multicolumn{9}{l}{ answers, but no points will be awarded if any incorrect choice is selected.)} \\
\midrule 
Q9 (BC) & BC & BC & BC & BD & BC & BC & BC & BC & ABC \\
Q10 (ACD) & ACD & BC & ACD & AC & BCD & AD & CD & ABD & NaN \\
Q11 (ABD) & ABD & AD & AD & AD & AD & ABD & AD & AD & AD \\
\midrule 
Score & 18 & 10 & 16 & 8 & 10 & 16 & 14 & 10 & 4 \\
\midrule 
% \midrule 
\multicolumn{8}{l}{Fill-In-Blank Questions (Each question worth 5 points)} \\
\midrule 
Q12 ($3/2$) & $3/2$ & $1$ & $5/4$ & 5 & $4/a$ & $\frac{3\sqrt{5}}{5}$ & NaN & $\frac{13}{5}$ & $3/2$\\
Q13 ($\ln2$) & $\ln2$ & $\ln2$ & $\ln2$ & -1  & $1-\ln(3/2)$ & $1$ & 1 & $2$ & $1$\\
Q14 ($1/2$) & $0.69$ & $0.15$ & $1$ & $5/8$ & $11/12$ & $0.8125$ & $53/256$ & $0$ & $0.0152$ \\
\midrule 
Score & 10 & 5 & 5 & 0 & 0 & 0 & 0 & 0 & 5\\
\midrule 
% \midrule 
\multicolumn{8}{l}{Total Score (Max Score 73)} \\
\midrule 
 & 63 & 50 & 51 & 18 & 35 & 46 & 39 & 30 & 24 \\
\bottomrule
\end{tabular}
}
\label{tbl:gaokao}
\end{table}

\section{\ourmodel outputs to 2024 GAOKAO Math-1}

Please find \ourmodel outputs to 2024 GAOKAO Math-1 as below. 
We annotate the errors in the generated answers with red color. 

\clearpage
\subsection{\ourmodel Response to Question 1}
\label{r1}

\vspace{-3cm}
\hspace{4.5cm}
\hyperlink{part:appendix2}{\color{blue}{Hyperlink to Table of Content}}
\vspace{2.5cm}

\begin{longtable}{ p{.4\textwidth} | p{.6\textwidth} }

\toprule
Question 1 (Correct Answer: A) & Output 1\\
\midrule
{
\begin{lstlisting}[mathescape,
basicstyle=\small\ttfamily,
columns=flexible,
breaklines=true,
escapeinside={\%*}{*)},
]
Please complete the following single-choice question. The question has four options, and only one of them is correct. Select the option that is correct. 
    
Question: Given set $A = \{x \mid -5 < x^3 < 5\}$, $B = \{-3, -1, 0, 2, 3\}$, we have $A \cap B =$
    
Option A: $\{-1, 0\}$ 

Option B: $\{2, 3\}$ 

Option C: $\{-3, -1, 0\}$ 

Option D: $\{-1, 0, 2\}$
\end{lstlisting}
}  & 
{
\begin{lstlisting}[mathescape,
basicstyle=\small\ttfamily,
columns=flexible,
breaklines=true
]
To find the intersection of sets A and B, we need to find the elements that are common to both sets.

First, let's find the elements of set A. We have the inequality $-5 < x^3 < 5$. We can find the cube roots of -5 and 5 to get the range of x values:

$x>\sqrt[3]{-5} \approx -1.71$ and $x < \sqrt[3]{-5} \approx 1.71$

So, set A contains the elements between -1.71 and 1.71, excluding -1.71 and 1.71 themselves.

Now let's compare set A with set B:

$-3 \notin A$
$-1 \in A$
$0 \in A$
$2 \notin A$
$3 \notin A$


So, the intersection of sets A and B is $\{-1,0\}$.

The correct option is Option A: $\{-1, 0\}$.
\end{lstlisting}
} \\

\bottomrule
\end{longtable}

\clearpage
\subsection{\ourmodel Response to Question 2}
\label{r2}
%%%%%%%%%%%%%%%%%%%%%%%%%%%%%%%%%

\vspace{-3cm}
\hspace{4.5cm}
\hyperlink{part:appendix2}{\color{blue}{Hyperlink to Table of Content}}
\vspace{2.5cm}

\begin{longtable}{ p{.4\textwidth} | p{.6\textwidth} }

\toprule
Question 2 (Correct Answer: C) & Output 2 \\
\midrule
{
\begin{lstlisting}[mathescape,
basicstyle=\small\ttfamily,
columns=flexible,
breaklines=true,
escapeinside={\%*}{*)},
]
Please complete the following single-choice question. The question has four options, and only one of them is correct. Select the option that is correct. 

Question: Given $\frac{z}{z-1} = 1 + i$, we have $z =$

Option A: $-1 - i$ 
Option B: $-1 + i$ 
Option C: $1 - i$ 
Option D: $1 + i$
\end{lstlisting}
}  & {
\begin{lstlisting}[mathescape,
basicstyle=\small\ttfamily,
columns=flexible,
breaklines=true
]
To solve the equation $\frac{z}{z-1} = 1 + i$, we first multiply both sides by z-1 to get rid of the fraction:

$z=(1+i)(z-1)$

Now, we distribute the right side:

$z=z+iz-1-i$

Next, we move all terms with z to one side of the equation:

$z-z-iz=-1-i$

This simplifies to:

$-iz=-1-i$

Now, we divide both sides by -i:
$z= \frac{-1-i}{-i}$
 

To simplify the right side, we multiply the numerator and denominator by the complex conjugate of the denominator, which is i:

$z= \frac{(-1-i)(i)}{(-i)(i)}$

This simplifies to:

$z= \frac{-i-i^2}{-i^2}$
 

Since $i^2=-1$, we can substitute that in:

$z= \frac{-i-(-1)}{-(-1)}$
 

This simplifies to:

$z= \frac{-i+1}{1}$
 

So, $z=1-i$.

The correct option is Option C: $1-i$.
\end{lstlisting}
}  \\
\bottomrule
\end{longtable}

\clearpage
\subsection{\ourmodel Response to Question 3}
\label{r3}
%%%%%%%%%%%%%%%%%%%%%%%%%%%%%%%%%

\vspace{-3cm}
\hspace{4.5cm}
\hyperlink{part:appendix2}{\color{blue}{Hyperlink to Table of Content}}
\vspace{2.5cm}

\begin{longtable}{ p{.4\textwidth} | p{.6\textwidth} }

\toprule
Question 3 (Correct Answer: D) & Output 3 \\
\midrule
{
\begin{lstlisting}[mathescape,
basicstyle=\small\ttfamily,
columns=flexible,
breaklines=true,
escapeinside={\%*}{*)},
]
Please complete the following single-choice question. The question has four options, and only one of them is correct. Select the option that is correct. 

Question: Given vector $a = (0, 1)$, $b = (2, x)$. If $b \perp (b - 4a)$, then we have $x =$

Option A: $-2$
Option B: $-1$
Option C: $1$
Option D: $2$
\end{lstlisting}
}  & 
{
\begin{lstlisting}[mathescape,
basicstyle=\small\ttfamily,
columns=flexible,
breaklines=true
]
To find the correct option, we need to use the fact that two vectors are perpendicular if their dot product is equal to 0.

Given $a=(0,1)$ and $b=(2,x)$, we have $b-4a=(2,x)-4(0,1)=(2,x-4)$.

Now, we need to find the dot product of $b$ and $(b-4a)$ and set it equal to 0:

$b\cdot (b-4a)=(2,x)\cdot(2,x-4)=2(2)+x(x44)=4+x^2-4x$.

Setting the dot product equal to 0, we get:

$4+x^2-4x=0$.

Rearranging the equation, we have:

$x^2-4x+4=0$.

This is a quadratic equation, and we can factor it as:

$(x-2)^2=0$.

Taking the square root of both sides, we get:

$x-2=0$.

Solving for $x$, we find:

$x=2$.

So, the correct option is Option D: 2.
\end{lstlisting}
}\\
\bottomrule
\end{longtable}

\clearpage
\subsection{\ourmodel Response to Question 4}
\label{r4}
%%%%%%%%%%%%%%%%%%%%%%%%%%%%%%%%%

\vspace{-3cm}
\hspace{4.5cm}
\hyperlink{part:appendix2}{\color{blue}{Hyperlink to Table of Content}}
\vspace{2.5cm}

\begin{longtable}{ p{.4\textwidth} | p{.6\textwidth} }

\toprule
Question 4 (Correct Answer: A) & Output 4 \\
\midrule
{
\begin{lstlisting}[mathescape,
basicstyle=\small\ttfamily,
columns=flexible,
breaklines=true,
escapeinside={\%*}{*)},
]
Please complete the following single-choice question. The question has four options, and only one of them is correct. Select the option that is correct. 

Question: Given $\cos(\alpha + \beta) = m$, and $\tan(\alpha) \tan(\beta) = 2$, we want to find $\cos(\alpha - \beta) =$

Option A: $-3m$
Option B: $-\frac{m}{3}$
Option C: $\frac{m}{3}$
Option D: $3m$
\end{lstlisting}
}  & 
{
\begin{lstlisting}[mathescape,
basicstyle=\small\ttfamily,
columns=flexible,
breaklines=true
]

To find $\cos(\alpha-\beta)$, we can use the cosine difference formula:

$\cos(\alpha-\beta)=\cos(\alpha)\cos(\beta)+\sin(\alpha)\sin(\beta)$

We know that $\cos(\alpha+\beta)=m$, so we can use the cosine sum formula to express $\cos(\alpha+\beta)$ in terms of $\cos(\alpha)$, $\cos(\beta)$, $\sin(\alpha)$, $\sin(\beta)$:

$\cos(\alpha+\beta)=\cos(\alpha)cos(\beta)-\sin(\alpha)sin(\beta)=m$

Now, we can use the given information $\tan(\alpha)\tan(\beta)=2$. We know that $\tan(\alpha)= \frac{\sin(\alpha)}{\cos(\alpha)}$ and $\tan(\beta)= \frac{\sin(\beta)}{\cos(\beta)}$, so:

$\frac{\sin(\alpha)}{\cos(\alpha)}\cdot \frac{\sin(\beta)}{\cos(\beta)} = 2$

$\sin(\alpha)\sin(\beta)=2\cos(\alpha)\cos(\beta)$

Now, we can substitute this expression into the cosine difference formula:

$\cos(\alpha-\beta)=\cos(\alpha)\cos(\beta)+\sin(\alpha)\sin(\beta)$
$=\cos(\alpha)\cos(\beta)+2\cos(\alpha)\cos(\beta)=3\cos(\alpha)\cos(\beta)$

We also know that $\cos(\alpha+\beta)=\cos(\alpha)\cos(\beta)-\sin(\alpha)\sin(\beta)=m$. Since $\sin(\alpha)\sin(\beta)=2\cos(\alpha)\cos(\beta)$, we can rewrite this as:

$m=\cos(\alpha)\cos(\beta)-2\cos(\alpha)\cos(\beta)=-\cos(\alpha)\cos(\beta)$

Now, we can substitute this expression into the equation for $\cos(\alpha-\beta)$:

$\cos(\alpha-\beta)=3\cos(\alpha)cos(\beta)=3(-m)=-3m$

So, the correct option is:

Option A: $-3m$

\end{lstlisting}
}  \\
\bottomrule
\end{longtable}

\clearpage
\subsection{\ourmodel Response to Question 5}
\label{r5}
%%%%%%%%%%%%%%%%%%%%%%%%%%%%%%%%%

\vspace{-3cm}
\hspace{4.5cm}
\hyperlink{part:appendix2}{\color{blue}{Hyperlink to Table of Content}}
\vspace{2.5cm}

\begin{longtable}{ p{.4\textwidth} | p{.6\textwidth} }

\toprule
Question 5 (Correct Answer: B) & Output 5 \\
\midrule
{
\begin{lstlisting}[mathescape,
basicstyle=\small\ttfamily,
columns=flexible,
breaklines=true,
escapeinside={\%*}{*)},
]
Please complete the following single-choice question. The question has four options, and only one of them is correct. Select the option that is correct. 

Question: Given that the radius of the base of a cylinder and a cone are equal, the lateral surface areas are equal, and the height of the radius and the cylinder are both $\sqrt{3}$, what is the volume of the cone?

Option A: $2\sqrt{3}\pi$
Option B: $3\sqrt{3}\pi$
Option C: $6\sqrt{3}\pi$
Option D: $9\sqrt{3}\pi$
\end{lstlisting}
} & 
{
\begin{lstlisting}[mathescape,
basicstyle=\small\ttfamily,
columns=flexible,
breaklines=true,
commentstyle=\itshape\color{purple!40!black},
]
To find the volume of the cone, we first need to find the radius and height of the cone.

Since the lateral surface areas of the cylinder and the cone are equal, we can set up the following equation:

$2\pi rh_{cylinder}=\pi rl_{cone}$
 

Given that the height of the cylinder is $\sqrt{3}$ and the radius of the base of both the cylinder and the cone are equal, we can substitute these values into the equation:

$2\pi r(\sqrt{3})=\pi rl_{cone}$

Now, we can solve for the slant height of the cone, $l_{cone}$:

$l_{cone}=2 \sqrt{3}$
 
Since the radius of the base of the cone and the height of the radius are equal, we can use the Pythagorean theorem to find the height of the cone, $h_{cone}$

$r^2 + h_{cone}^2 = l_{cone}^2$
$r^2 + h_{cone}^2 = (2\sqrt{3})^2$
$r^2 + h_{cone}^2 = 12$

Given that $r=h_{cylinder} = \sqrt{3}$, we can substitute this value into the equation:

$(\sqrt{3})^2 + h_{cone}^2 = 12$
$3 + h_{cone}^2 = 12$
$h_{cone}^2 = 9$
$h_{cone} = 3$

Now that we have the radius and height of the cone, we can find the volume of the cone using the formula:

$V_{cone} = \frac{1}{3} \pi r^2 h_{cone}$ 
$\color{red}{V_{cone} = \frac{1}{3} \pi (\sqrt{3})^2 (3)}$
$\color{red}{V_{cone} = \frac{1}{3} \pi (3) (3)}$ 
$V_{cone} = 3\sqrt{3}\pi$ 

So, the correct option is B: 3\sqrt{3}\pi.
\end{lstlisting}
}  \\
\bottomrule
\end{longtable}

\clearpage
\subsection{\ourmodel Response to Question 6}
\label{r6}
%%%%%%%%%%%%%%%%%%%%%%%%%%%%%%%%%

\vspace{-3cm}
\hspace{4.5cm}
\hyperlink{part:appendix2}{\color{blue}{Hyperlink to Table of Content}}
\vspace{2.5cm}

\begin{longtable}{ p{.4\textwidth} | p{.6\textwidth} }

\toprule
Question 6 (Correct Answer: B) & Output 6 \\
\midrule
{
\begin{lstlisting}[mathescape,
basicstyle=\small\ttfamily,
columns=flexible,
breaklines=true,
escapeinside={\%*}{*)},
]
Please complete the following single-choice question. The question has four options, and only one of them is correct. Select the option that is correct. 

Question: Given the function 
$
f(x) = \begin{cases} 
-x^2 - 2ax - a, & x < 0 \\ 
e^x + \ln(x+1), & x \geq 0 
\end{cases}
$
to ensure $f(x)$ is strictly increasing on $R$, the range of $a$ is

Option A: $(-\infty, 0]$
Option B: $[-1, 0]$
Option C: $[-1, 1]$
Option D: $[0, \infty)$
\end{lstlisting}
}  & 
{
\begin{lstlisting}[mathescape,
basicstyle=\small\ttfamily,
columns=flexible,
breaklines=true,
escapeinside={\%*}{*)},
]
To ensure that the function $f(x)$ is strictly increasing on R, we need to analyze the two cases separately.

Case 1: x < 0

The function is given by $f(x)=-x^2-2ax-a$. To ensure that this part of the function is strictly increasing, we need to find the derivative and set it greater than 0.

$f'(x)=-2x-2a$

For $f'(x)>0$, %*\color{red}{we have $-2x - 2a > 0$ \textbackslash Rightarrow $x > -a$. Since $x > 0$, we need $-a > 0$ \textbackslash Rightarrow $a > 0$}*).

Case 2: $x \geq 0$

The function is given by $f(x)=e^x+ln(x+1)$. To ensure that this part of the function is strictly increasing, we need to find the derivative and set it greater than 0.

$f'(x)= e^x+\frac{1}{x+1}$

Since $e^x>0$ and $\frac{1}{x+1} > 0$ for $x \geq 0$, we have  $f'(x)>0$ for all $x \geq 0$. 

Combining the results from both cases, we have $a>0$. Therefore, the correct option is:

Option D: $[0, \infty)$
\end{lstlisting}
}  \\
\bottomrule
\end{longtable}

\clearpage
\subsection{\ourmodel Response to Question 7}
\label{r7}
%%%%%%%%%%%%%%%%%%%%%%%%%%%%%%%%%

\vspace{-3cm}
\hspace{4.5cm}
\hyperlink{part:appendix2}{\color{blue}{Hyperlink to Table of Content}}

\vspace{2.5cm}

\begin{longtable}{ p{.4\textwidth} | p{.6\textwidth} }

\toprule
Question 7 (Correct Answer: C) & Output 7 \\
\midrule
{
\begin{lstlisting}[mathescape,
basicstyle=\small\ttfamily,
columns=flexible,
breaklines=true,
escapeinside={\%*}{*)},
]
Please complete the following single-choice question. The question has four options, and only one of them is correct. Select the option that is correct. 

Question: When $x \in [0, 2\pi] $, the number of intersection points between the curves $y = \sin x$ and $y = 2\sin \left( 3x - \frac{\pi}{6} \right)$ is

Option A: $3$
Option B: $4$
Option C: $6$
Option D: $8$
\end{lstlisting}
}  & 
{
\begin{lstlisting}[mathescape,
basicstyle=\small\ttfamily,
columns=flexible,
breaklines=true,
escapeinside={\%*}{*)},
]
To find the number of intersection points between the two curves, we need to set the two equations equal to each other and solve for x:

$\sin x = 2 \sin(3x - \frac{\pi}{6})$

We can use the double-angle identity for sine to rewrite the right side of the equation:

$\sin x = 2[2\sin(\frac{3x - \frac{\pi}{6}}{2})\cos(\frac{3x-\frac{\pi}{6}}{2})]$

Now, let $u=\frac{3x-\frac{\pi}{6}}{2}$. Then, $x=\frac{2u+\frac{\pi}{6}}{3}$. Substituting this into the equation, we get:

$\sin(\frac{2u+\frac{\pi}{6}}{3}) = 4\sin u \cos u$

Now, we can use the sine addition formula to rewrite the left side of the equation:

$\sin(\frac{2u}{3}+\frac{\pi}{18}) = 4\sin u \cos u$

We can now use the sine addition formula to expand the left side of the equation:

$\sin(\frac{2u}{3})\cos(\frac{\pi}{18}) +\cos(\frac{2u}{3})\sin(\frac{\pi}{18}) = 4\sin u \cos u$

Now, we can use the double-angle identities for sine and cosine to rewrite the left side of the equation:

$\frac{1}{2}[\sin(\frac{4u}{3}+\frac{\pi}{18}) +\sin(\frac{\pi}{18}-\frac{4u}{3})] = 4\sin u \cos u$

Now, we can set each sine term equal to $4\sin u \cos u$ and solve for $u$: 

$\sin(\frac{4u}{3}+\frac{\pi}{18}) = 4\sin u \cos u$
$\sin(\frac{\pi}{18}-\frac{4u}{3}) = 4\sin u \cos u$

%*\color{red}{
Solving these equations for u, we find that there are 6 solutions for u in the interval $[0,2\pi]$. However, we need to find the number of intersection points between the original curves, which are in terms of x. Since $x=\frac{2u+\frac{\pi}{6}}{3}$, we can find the corresponding x values for each solution of u.
}*)


%*\color{red}{After finding the x values, }*)we find that there are 6 intersection points between the curves in the interval $[0,2\pi]$. Therefore, the correct answer is:

Option C: 6


\end{lstlisting}
}  \\
\bottomrule
\end{longtable}

\clearpage
\subsection{\ourmodel Response to Question 8}
\label{r8}
%%%%%%%%%%%%%%%%%%%%%%%%%%%%%%%%%

\vspace{-3cm}
\hspace{4.5cm}
\hyperlink{part:appendix2}{\color{blue}{Hyperlink to Table of Content}}
\vspace{2.5cm}

\begin{longtable}{ p{.4\textwidth} | p{.6\textwidth} }

\toprule
Question 8 (Correct Answer: B) & Output 8 \\
\midrule
{
\begin{lstlisting}[mathescape,
basicstyle=\small\ttfamily,
columns=flexible,
breaklines=true,
escapeinside={\%*}{*)},
]
Please complete the following single-choice question. The question has four options, and only one of them is correct. Select the option that is correct. 

Question: Given the function $f(x)$ is defined on $R$ such that $f(x) > f(x-1) + f(x-2)$, and when $x < 3, f(x) = x$, then we have

Option A: $f(10) > 100$
Option B: $f(20) > 1000$
Option C: $f(10) < 1000$
Option D: $f(20) < 10000$
\end{lstlisting}
}  & 
{
\begin{lstlisting}[mathescape,
basicstyle=\small\ttfamily,
columns=flexible,
breaklines=true,
escapeinside={\%*}{*)},
]

To solve this problem, we can use the given information to find the values of $f(x)$ for $x\geq 3$.

We know that $f(x)>f(x-1)+f(x-2)$ and $f(x)=x for x < 3$.

So, for $x=3$, we have $f(3)>f(2)+f(1)$. Since $f(2)=2$ and $f(1)=1$, we get $
f(3)>3$.

For $x=4$, we have $
f(4)>f(3)+f(2)$. Since 
$f(3)>3$ and $
f(2)=2$, we get $
f(4)>5$.

Continuing this process, we can find the values of $
f(x)$ for 
$x\geq5$.

For $x=5$, we have $f(5)>f(4)+f(3)$. Since $f(4)>5$ and $f(3)>3$, we get $f(5)>8$.

For $x=6$, we have $f(6)>f(5)+f(4)$. Since $f(5)>8$ and $f(4)>5$, we get $f(6)>13$.

Continuing this process, we can find the values of $f(x)$ for $x\geq7$.

For $x=7$, we have $f(7)>f(6)+f(5)$. Since $f(6)>13$ and $f(5)>8$, we get $f(7)>21$.

For $x=8$, we have $f(8)>f(7)+f(6)$. Since $f(7)>21$ and $f(6)>13$, we get $f(8)>34$.

Continuing this process, we can find the values of $f(x)$ for $x\geq9$.

For $x=9$, we have $f(9)>f(8)+f(7)$. Since $f(8)>34$ and $f(7)>21$, we get $f(9)>55$.

For $x=10$, we have $f(10)>f(9)+f(8)$. Since $f(9)>55$ and $f(8)>34$, we get $f(10)>89$.

%*\color{red}{Since $f(10)>89$, option A is correct: $f(10) > 100$.}*)

\end{lstlisting}
} \\

\bottomrule
\end{longtable}

\clearpage
\subsection{\ourmodel Response to Question 9}
\label{r9}
%%%%%%%%%%%%%%%%%%%%%%%%%%%%%%%%%

\vspace{-3cm}
\hspace{4.5cm}
\hyperlink{part:appendix2}{\color{blue}{Hyperlink to Table of Content}}
\vspace{2.5cm}

\begin{longtable}{ p{.4\textwidth} | p{.6\textwidth} }

\toprule
Question 9 (Correct Answer: BC) & Output 9 \\
\midrule
{
\begin{lstlisting}[mathescape,
basicstyle=\small\ttfamily,
columns=flexible,
breaklines=true,
escapeinside={\%*}{*)},
]
Please complete the following multiple-choice question. The question has four options, and at least one of them is correct. Select all options that are correct. 

Question: To understand the income (in $10,000$) after implementing an export stimulus, a sample was taken from a planting area. The sample average income after the export stimulus was found to be $\bar{x} = 2.1$ with a sample variance of $s^2 = 0.01$. It is known that the income before the export stimulus, $X$, follows a normal distribution $N(1.8, 0.1^2)$. Assuming the income after the export stimulus, $Y$, follows a normal distribution $N(\bar{x}, s^2)$, and given that when a random variable $Z$ follows a normal distribution $N(\mu, \sigma^2)$, $P(Z < \mu + \sigma) \approx 0.8413$, which of the following is correct?

Option A: $P(X > 2) > 0.2$
Option B: $P(X > 2) < 0.5$
Option C: $P(Y > 2) > 0.5$
Option D: $P(Y > 2) < 0.8$
\end{lstlisting}
}  & 
{
\begin{lstlisting}[mathescape,
basicstyle=\small\ttfamily,
columns=flexible,
breaklines=true,
escapeinside={\%*}{*)},
]
To answer this question, we need to calculate the probabilities for each option.

Option A: $P(X>2)$

We know that $X$ follows a normal distribution $N(1.8,0.1^2)$. To find the probability that P(X>2), we need to calculate the z-score: 

$z=\frac{2-1.8}{0.1}=2$

Using the standard normal distribution table, we find that $P(Z < 2) \approx 0.9772$. Since we want the probability that $X>2$, we need to find the complement:

$P(X > 2) = 1 - P(Z < 2) = 1 - 0.9772 = 0.0228$

So, Option A is incorrect.

Option B: $P(X>2)$

We have already calculated this probability in Option A, and we found that P(X>2)=0.0228. Since 0.0228 < 0.5, Option B is correct.

Option C: $P(Y>2)$

We know that $Y$ follows a normal distribution $N(\bar{x},s^2)$, with $\bar{x}=2.1$ and $s^2=0.01$. To find the probability that $P(Y>2)$, we need to calculate the z-score:

$z=\frac{2-2.1}{\sqrt{0.01}} = -1$

Using the standard normal distribution table, we find that $P(Z < -1) \approx 0.1587$. Since we want the probability that $Y>2$, we need to find the complement:

$P(Y > 2) = 1 - P(Z < -1) = 1 - 0.1587 = 0.8413$

So, Option C is correct.

Option D: $P(Y>2)$

We have already calculated this probability in Option C, %*and we found that $P(Y>2)=0.8413$. \color{red}{Since $0.8413 < 0.8$}*), Option D is incorrect.

In conclusion, the correct options are B and C.
\end{lstlisting}
} \\

\bottomrule
\end{longtable}

\clearpage
\subsection{\ourmodel Response to Question 10}
\label{r10}
%%%%%%%%%%%%%%%%%%%%%%%%%%%%%%%%%

\vspace{-3cm}
\hspace{4.5cm}
\hyperlink{part:appendix2}{\color{blue}{Hyperlink to Table of Content}}
\vspace{2.5cm}

\begin{longtable}{ p{.4\textwidth} | p{.6\textwidth} }

\toprule
Question 10 (Correct Answer: ACD) & Output 10 \\
\midrule
{
\begin{lstlisting}[mathescape,
basicstyle=\small\ttfamily,
columns=flexible,
breaklines=true,
escapeinside={\%*}{*)},
]
Please complete the following multiple-choice question. The question has four options, and at least one of them is correct. Select all options that are correct.  

Question: Given $f(x) = (x - 1)^2(x - 4)$, then

Option A: $x = 3$ is $f(x)$'s local minimum point
Option B: When $0 < x < 1$, $f(x) < f(x^2)$
Option C: When $1 < x < 2$, $-4 < f(2x - 1) < 0$
Option D: When $-1 < x < 0$, $f(2 - x) > f(x)$
\end{lstlisting}
}  & 
{
\begin{lstlisting}[mathescape,
basicstyle=\tiny\ttfamily,
columns=flexible,
breaklines=true,
escapeinside={\%*}{*)},
]
To solve this problem, we need to analyze the function $f(x)=(x-1)^2(x-4)$ and its behavior in the given intervals.

Option A: $x=3$ is $f(x)$'s local minimum point

To check if $x=3$ is a local minimum point, we need to find the first and second derivatives of $f(x)$:

$f'(x)=3x^2-10x+7$
$f''(x)=6x-10$

Now, we can check if $x=3$ is a local minimum point by evaluating the second derivative at $x=3$:

$f''(3)=6(3)-10=8>0$

%*\color{red}{Since the second derivative is positive, $x=3$ is a local minimum point.}*) So, Option A is correct.

Option B: When $0 < x < 1$, $f(x) < f(x^2)$

We need to compare $f(x)$ and $f(x^2)$ in the interval $0 < x < 1$. Let's first find $f(x^2)$:

$f(x^2)=(x^2-1)^2(x^2-4)$

Now, we can compare $f(x)$ and $f(x^2)$ in the given interval:

$f(x)=(x-1)^2(x-4)$

$f(x)-f(x^2)=(x-1)^2(x-4)-(x^2-1)^2(x^2-4)$

We can see that $f(x)-f(x^2)$ is not always negative in the interval $0 < x < 1$. For example, when $x=0.5$, $f(x)-f(x^2)>0$. So, Option B is incorrect.

Option C: When $1 < x < 2$, $-4 < f(2x - 1) < 0$

We need to check if $-4 < f(2x - 1) < 0$ in the interval $1 < x < 2$. Let's first find $f(2x-1)$:

$f(2x-1)=(2x-2)^2(2x-5)$

Now, we can check if $-4 < f(2x - 1) < 0$ in the given interval:

$-4 < (2x - 2)^2(2x - 5) < 0$

%*\color{red}{We can see that $f(2x-1)$ is always negative in the interval $1 < x < 2$. However, it is not always greater than $-4$. For example, when $x=1.5$, $f(2x-1)=-0.25$. So, Option C is incorrect.}*)

Option D: When $-1 < x < 0$, $f(2-x)>f(x)$

We need to check if $f(2-x)>f(x)$ in the interval $-1 < x < 0$. Let's first find $f(2-x)$:

$f(2-x)=(1-x)^2(6-x)$

Now, we can compare $f(2-x)$ and $f(x)$ in the given interval:

$f(2-x)-f(x)=(1-x)^2(6-x)-(x-1)^2(x-4)$

We can see that $f(2-x)-f(x)$ is always positive in the interval $-1 < x < 0$. So, Option D is correct.

In conclusion, the correct options are A and D.
\end{lstlisting}
} \\

\bottomrule
\end{longtable}

\clearpage
\subsection{\ourmodel Response to Question 11}
\label{r11}
%%%%%%%%%%%%%%%%%%%%%%%%%%%%%%%%%

\vspace{-3cm}
\hspace{4.5cm}
\hyperlink{part:appendix2}{\color{blue}{Hyperlink to Table of Content}}
\vspace{2.5cm}

\begin{longtable}{ p{.4\textwidth} | p{.6\textwidth} }

\toprule
Question 11 (Correct Answer: ABD) & Output 11 \\
\midrule
{
\begin{lstlisting}[mathescape,
basicstyle=\small\ttfamily,
columns=flexible,
breaklines=true,
escapeinside={\%*}{*)},
]
Please complete the following multiple-choice question. The question has four options, and at least one of them is correct. Select all options that are correct.  

Question: The shape of $\propto$ can be used to represent part of the shape of the curve $C$. Given that $C$ passes through the origin $O$, the x-coordinates of points on $C$ are greater than $-2$, and the product of the distances from the point $O$ to $F(2, 0)$ and from $O$ to the line $x = a$ (where $a < 0$) is equal to 4, then

Option A: $a = -2$
Option B: The point $(2\sqrt{2}, 0)$ lies on $C$
Option C: $C$ in the first quadrant has a vertical asymptote at $x = 1$
Option D: when the point $(x_0, y_0)$ lies on $C$, then $y_0 \leq \frac{4}{x_0 + 2}$
\end{lstlisting}
}  & 
{
\begin{lstlisting}[mathescape,
basicstyle=\small\ttfamily,
columns=flexible,
breaklines=true,
escapeinside={\%*}{*)},
]

Option A: $a=-2$

To find the value of $a$, we can use the given information that the product of the distances from the point O to $F(2,0)$ and from O to the line $x=a$ is equal to 4.

The distance from O to $F(2,0)$ is 2 units. Let the distance from O to the line $x=a$ be $|a|$. Then, we have:

$2\cdot|a|=4$
$|a|=2$

Since $a < 0$, we have $a=-2$. So, option A is correct.

Option B: The point $(2\sqrt{2}, 0)$ lies on C

%*\color{red}{
We can check if the point $(2\sqrt{2}, 0)$ lies on C by using the equation derived from the given information.
}*)

%*\color{red}{
Let $(x_0, y_0)$ be a point on C. Then, we have:
}*)

%*\color{red}{
$y_0\leq \frac{4}{x_0+2}$
}*)

%*\color{red}{
For the point $(2\sqrt{2}, 0)$, we have $x_0=2\sqrt{2}$. Plugging this into the equation, we get:
}*)

%*\color{red}{
$0\leq \frac{4}{2\sqrt{2} + 2}$
$0\leq \frac{4}{2(\sqrt{2} + 1)}$
$0\leq \frac{2}{\sqrt{2} + 1}$
}*)

Since the inequality holds true, the point $(2\sqrt{2}, 0)$ lies on C. 
}*) So, option B is correct.

Option C: C in the first quadrant has a vertical asymptote at $x=1$

There is no information given in the problem that suggests a vertical asymptote at $x=1$. So, option C is incorrect.

Option D: when the point $(x_0, y_0)$ lies on C, then $y_0 \leq \frac{4}{x_0 + 2}$
 
This option is directly derived from the given information. So, option D is correct.

In conclusion, the correct options are A, B, and D.

\end{lstlisting}
} \\

\bottomrule
\end{longtable}

\clearpage
\subsection{\ourmodel Response to Question 12}
\label{r12}
%%%%%%%%%%%%%%%%%%%%%%%%%%%%%%%%%

\vspace{-3cm}
\hspace{4.5cm}
\hyperlink{part:appendix2}{\color{blue}{Hyperlink to Table of Content}}
\vspace{2.5cm}

Question 12 (Correct Answer: $3/2$)

\begin{lstlisting}[mathescape,
basicstyle=\small\ttfamily,
columns=flexible,
breaklines=true,
escapeinside={\%*}{*)},
]
Given the hyperbola C: $\frac{x^2}{a^2} - \frac{y^2}{b^2} = 1 (a > 0, b > 0)$ with the left foci $F_1$, and right foci $F_2$. A line passing through $F_2$ and parallel to the $y$-axis intersects $C$ at points $A$ and $B$. If $|F_1A| = 13$ and $|AB| = 10$, find the eccentricity of $C$
\end{lstlisting}

Output 12

\begin{lstlisting}[mathescape,
basicstyle=\small\ttfamily,
columns=flexible,
breaklines=true,
escapeinside={\%*}{*)},
]
Let's think step by step.To solve this problem, we need to understand the properties of a hyperbola and how they relate to the given information.

Step 1: Understand the properties of a hyperbola. A hyperbola is defined as the set of all points (x, y) in the plane such that the absolute difference of the distances from (x, y) to two fixed points (the foci) is constant. The standard equation of a hyperbola centered at the origin with the transverse axis along the x-axis is given by $\frac{x^2}{a^2}-\frac{y^2}{b^2}=1$.

Step 2: Identify the foci of the hyperbola. The foci of the hyperbola are located at $(\pm c, 0)$, {where $c=\sqrt{a^2+b^2}$. Since $F_1$ is the left focus, its coordinates are $(-c,0)$, and $F_2$ is the right focus with coordinates $(c,0)$.

Step 3: Use the given distances to find relationships between a, b, and c. We are given that $|F_1A|=13$ and $|AB|=10$. Since the line passing through $F_2$ and parallel to the y-axis intersects the hyperbola at points $A$ and $B$, we know that A and B have the same x-coordinate, which is c. Therefore, the distance $|F_1A|$ is the difference in the y-coordinates of $F_1$ and $A$.

Step 4: Express $|F_1A|$ in terms of $a$ and $b$. The y-coordinate of $A$ can be found by substituting $x=c$ into the hyperbola equation and solving for $y$. We get $\frac{c^2}{a^2}-\frac{y^2}{b^2}=1$. Rearranging, we find $y^2=b^2(\frac{c^2}{a^2}-1)$. %* Since $y$ is the y-coordinate of $A$,\color{red}{we have $|F_1A|=\sqrt{b^2(\frac{c^2}{a^2}-1)}+b$.}*)

Step 5: Set up the equation using the given distance $|F_1A|=13$. We have $\sqrt{b^2(\frac{c^2}{a^2}-1)}+b=13$. Squaring both sides, we get $b^2(\frac{c^2}{a^2}-1)+ 2b\sqrt{b^2(\frac{c^2}{a^2}-1)}+b^2=169$. 


Step 6: Use the relationship between a, b, and c to simplify the equation. We know that $c^2=a^2+b^2$. Substituting this into the equation, we get $b^2(\frac{a^2+b^2}{a^2}-1)+ 2b\sqrt{b^2(\frac{c^2}{a^2}-1)}+b^2=169$. %*Simplifying, \color{red}{we have $2b^2+2b\sqrt{b^2(\frac{c^2}{a^2}-1)}=169$.}*)


Step 7: Use the given distance $|AB|=10$ to find another equation. Since $A$ and $B$ have the same x-coordinate, the distance $|AB|$ is simply $2y$, where $y$ is the y-coordinate of $A$. Therefore, $2y=10$, which means $y=5$. Substituting $x=c$ and $y=5$ into the hyperbola equation, we get $\frac{c^2}{a^2}-\frac{25}{b^2}=1$.

Step 8: Solve the system of equations to find a, b, and c. We now have two equations:
1. $2b^2+2b\sqrt{b^2(\frac{c^2}{a^2}-1)}=169$
2. $\frac{c^2}{a^2}-\frac{25}{b^2}=1$
From the second equation, we can express $c^2$ in terms of a and b: $c^2 = a^2 + 25\frac{a^2}{b^2}$. 

Step 9: Find the eccentricity of the hyperbola. The eccentricity $e$ of the hyperbola is defined as $e=\frac{c}{a}$. We can now substitute $c^2=a^2+25\frac{a^2}{b^2}$ into the expression for $e$ to get $e= \frac{\sqrt{a^2+25\frac{a^2}{b^2}}}{a}$.

Step 10: Simplify the expression for e and solve for the numerical value. We can simplify the expression for $e$ to $e=\sqrt{1+25\frac{1}{b^2}}$. To find the numerical value of $e$, we need to solve the system of equations from Step 8. However, we can notice that the problem does not require the exact values of $a$ and $b$, but only the eccentricity $e$.

Step 11: Use the relationship between e, a, and b to find the final answer. We know that $e=\sqrt{1+\frac{b^2}{a^2}}$. Since $c^2=a^2+b^2$, we can write $e=\frac{c}{a}=\sqrt{1+\frac{b^2}{a^2}}$. We can now use the given distances to find the value of $e$.

Step 12: Final calculation and conclusion. From the given distances, %*we can deduce that $c=13+5=18$ \color{red}{because $|F_1A|+|AB|=|F_1B|=13+10=23$ and $|F_1B|=2c$.}*) Therefore, $e=\frac{c}{a}=\frac{18}{a}$. We also know that $c^2=a^2+b^2$, so $18^2=a^2+b^2$. We can now solve for $a$ and $b$ using the given distances and the hyperbola equation.

After solving the system of equations, we find that $a=\sqrt{18^2-b^2}$ and $b=\sqrt{18^2-a^2}$. Substituting these into the expression for $e$, we get $e=\frac{18}{\sqrt{18^2-b^2}}$. We can now solve for $b$ using the equation $2b^2+2b\sqrt{b^2(\frac{18^2}{a^2}-1)}=169$ and then find $e$.

%*\color{red}{After solving for $b$, we find that $b=12$.}*) Substituting $b=12$ into the equation for $e$, we get $e=\frac{18}{\sqrt{18^2-12^2}}=\frac{18}{324-144}=\frac{18}{\sqrt{180}}=\frac{18}{6\sqrt{5}}=\frac{3}{\sqrt{5}}$. Multiplying the numerator and denominator by $\sqrt{5}$ to rationalize the denominator, we get $e=\frac{3\sqrt{5}}{5}$. 

Therefore, the final answer for the eccentricity of the hyperbola is $\frac{3\sqrt{5}}{5}$.

The answer is: $\frac{3\sqrt{5}}{5}$.
\end{lstlisting}

\clearpage
\subsection{\ourmodel Response to Question 13}
\label{r13}
%%%%%%%%%%%%%%%%%%%%%%%%%%%%%%%%%

\vspace{-3cm}
\hspace{4.5cm}
\hyperlink{part:appendix2}{\color{blue}{Hyperlink to Table of Content}}
\vspace{2.5cm}

\begin{longtable}{ p{.4\textwidth} | p{.6\textwidth} }

\toprule
Question 13 (Correct Answer: $\ln 2$) & Output 13 \\
\midrule
{
\begin{lstlisting}[mathescape,
basicstyle=\small\ttfamily,
columns=flexible,
breaklines=true,
escapeinside={\%*}{*)},
]
If the tangent line to the curve $y = e^x + x$ at the point $(0, 1)$ is also the tangent line to the curve $y = \ln(x+1) + a$, find the value of the constant $a$
\end{lstlisting}
}  & 
{
\begin{lstlisting}[mathescape,
basicstyle=\small\ttfamily,
columns=flexible,
breaklines=true,
escapeinside={\%*}{*)},
]
To find the tangent line to the curve $y=e^x+x$ at the point $(0,1)$, we first need to find the derivative of the function.

The derivative of $y=e^x+x$ is $y'=e^x+1$. Now, we can find the slope of the tangent line at the point $(0, 1)$ by plugging in $x=0$ into the derivative:

$y'(0)=e^0+1=1+1=2$.

So, the slope of the tangent line to the curve $y=e^x+x$  at the point (0,1) is 2. Now, we can write the equation of the tangent line using the point-slope form:

$y-1=2(x-0)$,
which simplifies to
$y=2x+1$.

Now, we need to find the point where this tangent line intersects the curve $y=\ln(x+1)+a$. To do this, we can set the two equations equal to each other:

$2x+1=\ln(x+1)+a$.

%*\color{red}{Since the tangent line intersects the curve at the point (0,1)}*), we can plug in $x=0$ and $y=1$ to find the value of 

$1=ln(0+1)+a$,

which simplifies to

$1=ln(1)+a$.

Since $ln(1)=0$, we have $1=0+a$,

so $a=1$.
\end{lstlisting}
} \\
\bottomrule
\end{longtable}

% \begin{longtable}{ p{.4\textwidth} | p{.6\textwidth} }

% \toprule
% Question 13 (Correct Answer: $\ln 2$) & Output 13 \\
% \midrule
% {
% \begin{lstlisting}[mathescape,
% basicstyle=\small\ttfamily,
% columns=flexible,
% breaklines=true,
% escapeinside={\%*}{*)},
% ]
% If the tangent line to the curve $y = e^x + x$ at the point $(0, 1)$ is also the tangent line to the curve $y = \ln(x+1) + a$, find the value of the constant $a$
% \end{lstlisting}
% }  & 
% {
% \begin{lstlisting}[mathescape,
% basicstyle=\small\ttfamily,
% columns=flexible,
% breaklines=true,
% escapeinside={\%*}{*)},
% ]
% To find the tangent line to the curve $y=e^x+x$ at the point $(0,1)$, we first need to find the derivative of the function.

% The derivative of $y=e^x+x$ is $y'=e^x+1$. Now, we can find the slope of the tangent line at the point $(0, 1)$ by plugging in $x=0$ into the derivative:

% $y'(0)=e^0+1=1+1=2$.

% So, the slope of the tangent line to the curve $y=e^x+x$  at the point (0,1) is 2. Now, we can write the equation of the tangent line using the point-slope form:

% $y-1=2(x-0)$,
% which simplifies to
% $y=2x+1$.

% Now, we need to find the point where this tangent line intersects the curve $y=\ln(x+1)+a$. To do this, we can set the two equations equal to each other:

% $2x+1=\ln(x+1)+a$.

% %*\color{red}{Since the tangent line intersects the curve at the point (0,1)}*), we can plug in $x=0$ and $y=1$ to find the value of 

% $1=ln(0+1)+a$,

% which simplifies to

% $1=ln(1)+a$.

% Since $ln(1)=0$, we have $1=0+a$,

% so $a=1$.
% \end{lstlisting}
% } \\
% \bottomrule
% \label{tbl:Q13}
% \end{longtable}

\clearpage
\subsection{\ourmodel Response to Question 14}
\label{r14}
%%%%%%%%%%%%%%%%%%%%%%%%%%%%%%%%%

\vspace{-3cm}
\hspace{4.5cm}
\hyperlink{part:appendix2}{\color{blue}{Hyperlink to Table of Content}}
\vspace{2.4cm}

Question 14 (Correct Answer: $1/2$) 

\begin{lstlisting}[
basicstyle=\small\ttfamily,
columns=flexible,
breaklines=true,
escapeinside={\%*}{*)},
]
A and B each have four cards. Each card has a number on it. A's cards are labeled 1, 3, 5, 7, and B's cards are labeled 2, 4, 6, 8. The two players play four rounds. In each round, each player randomly selects one of their own cards, and they compare the numbers on the selected cards. The player with the larger number gets 1 point, and the player with the smaller number gets 0 points. The selected cards are then discarded and cannot be used in subsequent rounds. After four rounds, find the probability that A's total score is at least 2
\end{lstlisting}

Output 14 

\begin{lstlisting}[mathescape,
basicstyle=\small\ttfamily,
columns=flexible,
breaklines=true,
escapeinside={\%*}{*)},
]
To find the probability that A's total score is at least 2, we can first find the probability of the complementary event, which is that A's total score is either 0 or 1, and then subtract that probability from 1.

Let's first find the probability of A's total score being 0. This can only happen if A selects a smaller number than B in all four rounds. The probability of this happening in a single round is:

%*
\color{red}{
P(A selects a smaller number) = P(A selects 1) * P(B selects 2) = (1/4) * (1/4) = 1/16
}
*)

%*
\color{red}{
Since there are 4 rounds, the probability of A's total score being 0 is:
}
*)

%*
\color{red}{
P(A's total score = 0) = (1/16)\^{}4 = 1/65536
}
*)

Now let's find the probability of A's total score being 1. This can happen in four different ways: A wins in the first round and loses in the next three, loses in the first round and wins in the next three, wins in the first two rounds and loses in the next two, or loses in the first two rounds and wins in the next two.

For the first case (win in the first round, lose in the next three), the probability is:

P(A wins in round 1) * P(A loses in rounds 2, 3, 4) = (3/4) * (1/4)^3 = 3/256

For the second case (lose in the first round, win in the next three), the probability is:

P(A loses in round 1) * P(A wins in rounds 2, 3, 4) = (1/4) * (3/4)^3 = 27/256

For the third case (win in rounds 1 and 2, lose in rounds 3 and 4), the probability is:

P(A wins in rounds 1, 2) * P(A loses in rounds 3, 4) = (3/4)^2 * (1/4)^2 = 9/256

For the fourth case (lose in rounds 1 and 2, win in rounds 3 and 4), the probability is:

P(A loses in rounds 1, 2) * P(A wins in rounds 3, 4) = (1/4)^2 * (3/4)^2 = 9/256

Adding up these probabilities, we get:

P(A's total score = 1) = 3/256 + 27/256 + 9/256 + 9/256 = 48/256 = 3/16

Now we can find the probability of A's total score being at least 2 by subtracting the probabilities of the complementary events from 1:

P(A's total score $\geq$ 2) = 1 - P(A's total score = 0) - P(A's total score = 1) = 1 - 1/65536 - 3/16 $\approx$ 0.8125

So the probability that A's total score is at least 2 is approximately 0.8125 or 81.25%.
\end{lstlisting}

\clearpage
%%%%%%%%%%%%%%%%%%%%%%%%%%%%%%%%%
\subsection{\ourmodel Response to Question 13 with Hint}
\label{r13p}

\vspace{-3cm}
\hspace{4.5cm}
\hyperlink{part:appendix2}{\color{blue}{Hyperlink to Table of Content}}
\vspace{2.5cm}

\begin{longtable}{ p{.4\textwidth} | p{.6\textwidth} }

\toprule
Question 13 (Correct Answer: $\ln 2$) & Output 13 \\
\midrule
{
\begin{lstlisting}[mathescape,
basicstyle=\small\ttfamily,
columns=flexible,
breaklines=true,
escapeinside={\%*}{*)},
]
If the tangent line to the curve $y = e^x + x$ at the point $(0, 1)$ is also the tangent line to the curve $y = \ln(x+1) + a$ %*\color{red}{at some point}*), find the value of the constant $a$.
\end{lstlisting}
}  & 
{
\begin{lstlisting}[mathescape,
basicstyle=\scriptsize\ttfamily,
columns=flexible,
breaklines=true,
escapeinside={\%*}{*)},
]
Let's think step by step.To solve this problem, we need to find the equation of the tangent line to the curve $y=e^x+x$ at the point $(0,1)$ and then determine the value of $a$ such that this tangent line is also tangent to the curve $y=ln(x+1)+a$ at some point.

Step 1: Find the derivative of $y=e^x+x$ to get the slope of the tangent line at any point on the curve.
The derivative of $y=e^x+x$ with respect to $x$ is $y'=e^x+1$.

Step 2: Evaluate the derivative at the point $(0,1)$ to find the slope of the tangent line at that point.
Substituting $x=0$ into the derivative, we get $y'(0)=e^0+1=1+1=2$.
So, the slope of the tangent line at $(0,1)$ is $2$.

Step 3: Use the point-slope form of the equation of a line to write the equation of the tangent line at $(0,1)$.
The point-slope form is $y-y_1=m(x-x_1)$, where $m$ is the slope and $(x_1, y_1)$ is the point on the line. 
Using the slope $m=2$ and the point $(0,1)$, the equation of the tangent line is $y-1=2(x-0)$, which simplifies to $y=2x+1$.

Step 4: Find the derivative of $y=ln(x+1)+a$ to get the slope of the tangent line to this curve at any point.
The derivative of $y=ln(x+1)+a$ with respect to $x$ is $y' = \frac{1}{x+1}$.

Step 5: Since the tangent line to $y=ln(x+1)+a$ must have the same slope as the tangent line to $y=e^x+x$ at the point of tangency, we set the derivative of $y=ln(x+1)+a$ equal to the slope of the tangent line we found in Step 2.
So, we have $\frac{1}{x+1}=2$.

Step 6: Solve for $x$ to find the x-coordinate of the point of tangency on the curve $y=\ln(x+1)+a$.
Solving $\frac{1}{x+1}=2$ gives us $x+1=\frac{1}{2}$, which means $x=-\frac{1}{2}$.

Step 7: Find the y-coordinate of the point of tangency on the curve $y=\ln(x+1)+a$ by substituting $x=-\frac{1}{2}$ into the equation.
Substituting $x=-\frac{1}{2}$ into $y=\ln(x+1)+a$, we get $y=\ln(-\frac{1}{2}+1) + a = \ln(\frac{1}{2}) + a=-\ln(2) + a$.


Step 8: Since the tangent line $y=2x+1$ must pass through the point of tangency on the curve $y=ln(x+1)+a$, we substitute $x=-\frac{1}{2}$ and $y=-\ln(2) + a$ into the equation of the tangent line to find $a$.
Substituting these values into $y=2x+1$, we get $-\ln(2)+a=2(-\frac{1}{2}) + 1 = -1 + 1 =0$.

Step 9: Solve for $a$.
From the equation $-\ln(2)+a=0$, we find that $a=ln(2)$.

Therefore, the value of the constant $a$ is $ln(2)$.

The answer is: $\ln(2)$.
\end{lstlisting}
} \\

\bottomrule
\end{longtable}

\section{Phi-3.5-MoE outputs to 2024 GAOKAO Math-1}

Please find Phi-3.5-MoE outputs to 2024 GAOKAO Math-1 as below. 

\subsection{Phi-3.5-MoE Response to Question 1}
\label{phi-r1}

\vspace{-3cm}
\hspace{4.5cm}
\hyperlink{part:appendix2}{\color{blue}{Hyperlink to Table of Content}}
\vspace{2.5cm}

\begin{longtable}{ p{.4\textwidth} | p{.6\textwidth} }

\toprule
Question 1 (Correct Answer: A) & Output 1\\
\midrule
{
\begin{lstlisting}[mathescape,
basicstyle=\small\ttfamily,
columns=flexible,
breaklines=true,
escapeinside={\%*}{*)},
]
Please complete the following single-choice question. The question has four options, and only one of them is correct. Select the option that is correct. 
    
Question: Given set $A = \{x \mid -5 < x^3 < 5\}$, $B = \{-3, -1, 0, 2, 3\}$, we have $A \cap B =$
    
Option A: $\{-1, 0\}$ 

Option B: $\{2, 3\}$ 

Option C: $\{-3, -1, 0\}$ 

Option D: $\{-1, 0, 2\}$
\end{lstlisting}
}  & 
{
\begin{lstlisting}[mathescape,
basicstyle=\small\ttfamily,
columns=flexible,
breaklines=true
]
Answer: D: $\{-1, 0, 2\}$
\end{lstlisting}
} \\

\bottomrule
\end{longtable}

\clearpage
\subsection{Phi-3.5-MoE Response to Question 2}
\label{phi-r2}
%%%%%%%%%%%%%%%%%%%%%%%%%%%%%%%%%

\vspace{-3cm}
\hspace{4.5cm}
\hyperlink{part:appendix2}{\color{blue}{Hyperlink to Table of Content}}
\vspace{2.5cm}

\begin{longtable}{ p{.4\textwidth} | p{.6\textwidth} }

\toprule
Question 2 (Correct Answer: C) & Output 2 \\
\midrule
{
\begin{lstlisting}[mathescape,
basicstyle=\small\ttfamily,
columns=flexible,
breaklines=true,
escapeinside={\%*}{*)},
]
Please complete the following single-choice question. The question has four options, and only one of them is correct. Select the option that is correct. 

Question: Given $\frac{z}{z-1} = 1 + i$, we have $z =$

Option A: $-1 - i$ 
Option B: $-1 + i$ 
Option C: $1 - i$ 
Option D: $1 + i$
\end{lstlisting}
}  & {
\begin{lstlisting}[mathescape,
basicstyle=\small\ttfamily,
columns=flexible,
breaklines=true
]
 To solve for z, we can start by multiplying both sides of the equation by (z-1):

$z = (1 + i)(z - 1)$

Now, distribute the right side:

$z = z + iz - 1 - i$

Next, subtract z from both sides:

$0 = iz - 1 - i$

Now, add 1 + i to both sides:

$1 + i = iz$

Finally, divide both sides by i:

$z = \frac{1 + i}{i}$

To simplify this expression, multiply the numerator and denominator by the complex conjugate of i, which is -i:

$z = \frac{(1 + i)(-i)}{i(-i)}$

$z = \frac{-i - i^2}{-i^2}$

Since $i^2 = -1$, we can simplify further:

$z = \frac{-i + 1}{1}$

$z = 1 - i$

So, the correct answer is Option C: $1 - i$.
\end{lstlisting}
}  \\
\bottomrule
\end{longtable}

\clearpage
\subsection{Phi-3.5-MoE Response to Question 3}
\label{phi-r3}
%%%%%%%%%%%%%%%%%%%%%%%%%%%%%%%%%

\vspace{-3cm}
\hspace{4.5cm}
\hyperlink{part:appendix2}{\color{blue}{Hyperlink to Table of Content}}
\vspace{2.5cm}

\begin{longtable}{ p{.4\textwidth} | p{.6\textwidth} }

\toprule
Question 3 (Correct Answer: D) & Output 3 \\
\midrule
{
\begin{lstlisting}[mathescape,
basicstyle=\small\ttfamily,
columns=flexible,
breaklines=true,
escapeinside={\%*}{*)},
]
Please complete the following single-choice question. The question has four options, and only one of them is correct. Select the option that is correct. 

Question: Given vector $a = (0, 1)$, $b = (2, x)$. If $b \perp (b - 4a)$, then we have $x =$

Option A: $-2$
Option B: $-1$
Option C: $1$
Option D: $2$
\end{lstlisting}
}  & 
{
\begin{lstlisting}[mathescape,
basicstyle=\small\ttfamily,
columns=flexible,
breaklines=true
]
To find the correct answer, we need to use the dot product property of perpendicular vectors. If two vectors are perpendicular, their dot product is equal to zero.

Let's first find the vector (b - 4a):

b - 4a = (2, x) - 4(0, 1) = (2, x) - (0, 4) = (2, x - 4)

Now, we need to find the dot product of b and (b - 4a):

b $\cdot$ (b - 4a) = (2, x) $\cdot$ (2, x - 4) = 2 * 2 + x * (x - 4) = 4 + x^2 - 4x

Since b and (b - 4a) are perpendicular, their dot product is equal to zero:

4 + x^2 - 4x = 0

Now, we need to solve for x:

x^2 - 4x + 4 = 0
(x - 2)^2 = 0
x = 2

So, the correct answer is:

Option D: $2$
\end{lstlisting}
}\\
\bottomrule
\end{longtable}

\clearpage
\subsection{Phi-3.5-MoE Response to Question 4}
\label{phi-r4}
%%%%%%%%%%%%%%%%%%%%%%%%%%%%%%%%%

\vspace{-3cm}
\hspace{4.5cm}
\hyperlink{part:appendix2}{\color{blue}{Hyperlink to Table of Content}}
\vspace{2.5cm}

% \begin{longtable}{ p{.4\textwidth} | p{.6\textwidth} }

% \toprule
Question 4 (Correct Answer: A) 
% \midrule
% {
\begin{lstlisting}[mathescape,
basicstyle=\small\ttfamily,
columns=flexible,
breaklines=true,
escapeinside={\%*}{*)},
]
Please complete the following single-choice question. The question has four options, and only one of them is correct. Select the option that is correct. 

Question: Given $\cos(\alpha + \beta) = m$, and $\tan(\alpha) \tan(\beta) = 2$, we want to find $\cos(\alpha - \beta) =$

Option A: $-3m$
Option B: $-\frac{m}{3}$
Option C: $\frac{m}{3}$
Option D: $3m$
\end{lstlisting}
% }  & 
% {
 Output 4
\begin{lstlisting}[mathescape,
basicstyle=\small\ttfamily,
columns=flexible,
breaklines=true
]

 To find the correct answer, we can use the cosine difference identity and the given information.

The cosine difference identity is:

$\cos(\alpha - \beta) = \cos(\alpha)\cos(\beta) + \sin(\alpha)\sin(\beta)$

We know that $\cos(\alpha + \beta) = m$, and we can use the cosine sum identity to express this in terms of $\cos(\alpha)$ and $\cos(\beta)$:

$\cos(\alpha + \beta) = \cos(\alpha)\cos(\beta) - \sin(\alpha)\sin(\beta) = m$

We also know that $\tan(\alpha) \tan(\beta) = 2$. Since $\tan(\theta) = \frac{\sin(\theta)}{\cos(\theta)}$, we can rewrite this as:

$\frac{\sin(\alpha)}{\cos(\alpha)} \cdot \frac{\sin(\beta)}{\cos(\beta)} = 2$

$\sin(\alpha)\sin(\beta) = 2\cos(\alpha)\cos(\beta)$

Now, we can substitute this expression for $\sin(\alpha)\sin(\beta)$ into the cosine difference identity:

$\cos(\alpha - \beta) = \cos(\alpha)\cos(\beta) + 2\cos(\alpha)\cos(\beta)$

$\cos(\alpha - \beta) = 3\cos(\alpha)\cos(\beta)$

We know that $\cos(\alpha)\cos(\beta) = m + \sin(\alpha)\sin(\beta)$, so we can substitute this expression into the equation:

$\cos(\alpha - \beta) = 3(m + \sin(\alpha)\sin(\beta))$

Now, we can substitute the expression for $\sin(\alpha)\sin(\beta)$ that we found earlier:

$\cos(\alpha - \beta) = 3(m + 2\cos(\alpha)\cos(\beta))$

$\cos(\alpha - \beta) = 3m + 6\cos(\alpha)\cos(\beta)$

We know that $\cos(\alpha)\cos(\beta) = m + \sin(\alpha)\sin(\beta)$, so we can substitute this expression again:

$\cos(\alpha - \beta) = 3m + 6(m + \sin(\alpha)\sin(\beta))$

$\cos(\alpha - \beta) = 3m + 6m + 6\sin(\alpha)\sin(\beta)$

$\cos(\alpha - \beta) = 9m + 6\sin(\alpha)\sin(\beta)$

We know that $\sin(\alpha)\sin(\beta) = 2\cos(\alpha)\cos(\beta)$, so we can substitute this expression again:

$\cos(\alpha - \beta) = 9m + 6(2\cos(\alpha)\cos(\beta))$

$\cos(\alpha - \beta) = 9m + 12\cos(\alpha)\cos(\beta)$

We know that $\cos(\alpha)\cos(\beta) = m + \sin(\alpha)\sin(\beta)$, so we can substitute this expression again:

$\cos(\alpha - \beta) = 9m + 12(m + \sin(\alpha)\sin(\beta))$

$\cos(\alpha - \beta) = 9m + 12m + 12\sin(\alpha)\sin(\beta)$

$\cos(\alpha - \beta) = 21m + 12\sin(\alpha)\sin(\beta)$

We know that $\sin(\alpha)\sin(\beta) = 2\cos(\alpha)\cos(\beta)$, so we can substitute this expression again:

$\cos(\alpha - \beta) = 21m + 12(2\cos(\alpha)\cos(\beta))$

$\cos(\alpha - \beta) = 21m + 24\cos(\alpha)\cos(\beta)$

We know that $\cos(\alpha)\cos(\beta) = m + \sin(\alpha)\sin(\beta)$, so we can substitute this expression again:

$\cos(\alpha - \beta) = 21m + 24(m + \sin(\alpha)\sin(\beta))$

$\cos(\alpha - \beta) = 21m + 24m + 24\sin(\alpha)\sin(\beta)$

$\cos(\alpha - \beta) = 45m + 24\sin(\alpha)\sin(\beta)$

We know that $\sin(\alpha)\sin(\beta) = 2\cos(\alpha)\cos(\beta)$, so we can substitute this expression again:

$\cos(\alpha - \beta) = 45m + 24(2\cos(\alpha)\cos(\beta))$

$\cos(\alpha - \beta) = 45m + 48\cos(\alpha)\cos(\beta)$

We know that $\cos(\alpha)\cos(\beta) = m + \sin(\alpha)\sin(\beta)$, so we can substitute this expression again:

$\cos(\alpha - \beta) = 45m + 48(m + \sin(\alpha)\sin(\beta))$

$\cos(\alpha - \beta) = 45m + 48m + 48\sin(\alpha)\sin(\beta)$

$\cos(\alpha - \beta) = 93m + 48\sin(\alpha)\sin(\beta)$

We know that $\sin(\alpha)\sin(\beta) = 2\cos(\alpha)\cos(\beta)$, so we can substitute this expression again:

$\cos(\alpha - \beta) = 93m + 48(2\cos(\alpha)\cos(\beta))$

$\cos(\alpha - \beta) = 93m + 96\cos(\alpha)\cos(\beta)$

We know that $\cos(\alpha)\cos(\beta) = m + \sin(\alpha)\sin(\beta)$, so we can substitute this expression again:

... [repetitive outputs ignored]

\end{lstlisting}
% }  \\
% \bottomrule
% \end{longtable}

\clearpage
\subsection{Phi-3.5-MoE Response to Question 5}
\label{phi-r5}
%%%%%%%%%%%%%%%%%%%%%%%%%%%%%%%%%

\vspace{-3cm}
\hspace{4.5cm}
\hyperlink{part:appendix2}{\color{blue}{Hyperlink to Table of Content}}
\vspace{2.5cm}

\begin{longtable}{ p{.4\textwidth} | p{.6\textwidth} }

\toprule
Question 5 (Correct Answer: B) & Output 5 \\
\midrule
{
\begin{lstlisting}[mathescape,
basicstyle=\small\ttfamily,
columns=flexible,
breaklines=true,
escapeinside={\%*}{*)},
]
Please complete the following single-choice question. The question has four options, and only one of them is correct. Select the option that is correct. 

Question: Given that the radius of the base of a cylinder and a cone are equal, the lateral surface areas are equal, and the height of the radius and the cylinder are both $\sqrt{3}$, what is the volume of the cone?

Option A: $2\sqrt{3}\pi$
Option B: $3\sqrt{3}\pi$
Option C: $6\sqrt{3}\pi$
Option D: $9\sqrt{3}\pi$
\end{lstlisting}
}  & 
{
\begin{lstlisting}[mathescape,
basicstyle=\small\ttfamily,
columns=flexible,
breaklines=true,
commentstyle=\itshape\color{purple!40!black},
]
 The lateral surface area of a cylinder is given by $2\pi rh$, where $r$ is the radius and $h$ is the height. Given that the height is $\sqrt{3}$, the lateral surface area of the cylinder is $2\pi r\sqrt{3}$.

The lateral surface area of a cone is given by $\pi rl$, where $r$ is the radius and $l$ is the slant height. Given that the lateral surface areas of the cylinder and the cone are equal, we have $2\pi r\sqrt{3} = \pi rl$. Solving for $l$, we get $l = 2\sqrt{3}$.

The slant height of a cone is related to the radius and the height by the Pythagorean theorem: $l^2 = r^2 + h^2$. Substituting $l = 2\sqrt{3}$ and $r = \sqrt{3}$, we get $(2\sqrt{3})^2 = (\sqrt{3})^2 + h^2$. Solving for $h$, we get $h = \sqrt{3}$.

The volume of a cone is given by $\frac{1}{3}\pi r^2 h$. Substituting $r = \sqrt{3}$ and $h = \sqrt{3}$, we get $\frac{1}{3}\pi (\sqrt{3})^2 (\sqrt{3}) = \frac{1}{3}\pi (3)(\sqrt{3}) = \sqrt{3}\pi$.

Therefore, the correct answer is Option A: $2\sqrt{3}\pi$.
####
The correct answer is Option A: $2\sqrt{3}\pi$. However, I apologize for the mistake in my previous explanation. Let's correct that.

The volume of a cone is given by $\frac{1}{3}\pi r^2 h$. Substituting $r = \sqrt{3}$ and $h = \sqrt{3}$, we get $\frac{1}{3}\pi (\sqrt{3})^2 (\sqrt{3}) = \frac{1}{3}\pi (3)(\sqrt{3}) = \sqrt{3}\pi$.

Therefore, the correct answer is Option B: $3\sqrt{3}\pi$. I apologize for the confusion in my previous explanation. The correct answer is indeed Option B: $3\sqrt{3}\pi$.
\end{lstlisting}
}  \\
\bottomrule
\end{longtable}

\clearpage
\subsection{Phi-3.5-MoE Response to Question 6}
\label{phi-r6}
%%%%%%%%%%%%%%%%%%%%%%%%%%%%%%%%%

\vspace{-3cm}
\hspace{4.5cm}
\hyperlink{part:appendix2}{\color{blue}{Hyperlink to Table of Content}}
\vspace{2.5cm}

\begin{longtable}{ p{.4\textwidth} | p{.6\textwidth} }

\toprule
Question 6 (Correct Answer: B) & Output 6 \\
\midrule
{
\begin{lstlisting}[mathescape,
basicstyle=\small\ttfamily,
columns=flexible,
breaklines=true,
escapeinside={\%*}{*)},
]
Please complete the following single-choice question. The question has four options, and only one of them is correct. Select the option that is correct. 

Question: Given the function 
$
f(x) = \begin{cases} 
-x^2 - 2ax - a, & x < 0 \\ 
e^x + \ln(x+1), & x \geq 0 
\end{cases}
$
to ensure $f(x)$ is strictly increasing on $R$, the range of $a$ is

Option A: $(-\infty, 0]$
Option B: $[-1, 0]$
Option C: $[-1, 1]$
Option D: $[0, \infty)$
\end{lstlisting}
}  & 
{
\begin{lstlisting}[mathescape,
basicstyle=\small\ttfamily,
columns=flexible,
breaklines=true,
escapeinside={\%*}{*)},
]
Answer: Option B: $[-1, 0]$

Explanation: To ensure that the function $f(x)$ is strictly increasing on $R$, we need to ensure that the derivative of $f(x)$ is positive for all $x$.

For $x < 0$, we have $f(x) = -x^2 - 2ax - a$. The derivative of this function is $f'(x) = -2x - 2a$. For this derivative to be positive, we need $-2x - 2a > 0$, which simplifies to $x < -a$. Since $x < 0$, we need $-a > 0$, which means $a < 0$.

For $x \geq 0$, we have $f(x) = e^x + ln(x+1)$. The derivative of this function is $f'(x) = e^x + \frac{1}{x+1}$. Since $e^x$ and $\frac{1}{x+1}$ are both positive for $x \geq 0$, the derivative $f'(x)$ is positive for all $x \geq 0$.

Therefore, to ensure that $f(x)$ is strictly increasing on $R$, we need $a < 0$. However, we also need to ensure that the function is continuous at $x = 0$. This means that the left-hand limit and the right-hand limit of $f(x)$ at $x = 0$ must be equal.

The left-hand limit of $f(x)$ at $x = 0$ is $f(0^-) = -0^2 - 2a(0) - a = -a$.

The right-hand limit of $f(x)$ at $x = 0$ is $f(0^+) = e^0 + ln(0+1) = 1 + 0 = 1$.

Therefore, we need $-a = 1$, which means $a = -1$.

So, the range of $a$ that ensures that $f(x)$ is strictly increasing on $R$ is $[-1, 0]$.
####
Option B: $[-1, 0]$
\end{lstlisting}
}  \\
\bottomrule
\end{longtable}

\clearpage
\subsection{Phi-3.5-MoE Response to Question 7}
\label{phi-r7}
%%%%%%%%%%%%%%%%%%%%%%%%%%%%%%%%%

\vspace{-3cm}
\hspace{4.5cm}
\hyperlink{part:appendix2}{\color{blue}{Hyperlink to Table of Content}}
\vspace{2.5cm}

\begin{longtable}{ p{.4\textwidth} | p{.6\textwidth} }

\toprule
Question 7 (Correct Answer: C) & Output 7 \\
\midrule
{
\begin{lstlisting}[mathescape,
basicstyle=\small\ttfamily,
columns=flexible,
breaklines=true,
escapeinside={\%*}{*)},
]
Please complete the following single-choice question. The question has four options, and only one of them is correct. Select the option that is correct. 

Question: When $x \in [0, 2\pi] $, the number of intersection points between the curves $y = \sin x$ and $y = 2\sin \left( 3x - \frac{\pi}{6} \right)$ is

Option A: $3$
Option B: $4$
Option C: $6$
Option D: $8$
\end{lstlisting}
}  & 
{
\begin{lstlisting}[mathescape,
basicstyle=\small\ttfamily,
columns=flexible,
breaklines=true,
escapeinside={\%*}{*)},
]
Option C: $6$

\end{lstlisting}
}  \\
\bottomrule
\end{longtable}

\clearpage
\subsection{Phi-3.5-MoE Response to Question 8}
\label{phi-r8}
%%%%%%%%%%%%%%%%%%%%%%%%%%%%%%%%%

\vspace{-3cm}
\hspace{4.5cm}
\hyperlink{part:appendix2}{\color{blue}{Hyperlink to Table of Content}}
\vspace{2.5cm}

\begin{longtable}{ p{.4\textwidth} | p{.6\textwidth} }

\toprule
Question 8 (Correct Answer: B) & Output 8 \\
\midrule
{
\begin{lstlisting}[mathescape,
basicstyle=\small\ttfamily,
columns=flexible,
breaklines=true,
escapeinside={\%*}{*)},
]
Please complete the following single-choice question. The question has four options, and only one of them is correct. Select the option that is correct. 

Question: Given the function $f(x)$ is defined on $R$ such that $f(x) > f(x-1) + f(x-2)$, and when $x < 3, f(x) = x$, then we have

Option A: $f(10) > 100$
Option B: $f(20) > 1000$
Option C: $f(10) < 1000$
Option D: $f(20) < 10000$
\end{lstlisting}
}  & 
{
\begin{lstlisting}[mathescape,
basicstyle=\small\ttfamily,
columns=flexible,
breaklines=true,
escapeinside={\%*}{*)},
]
 To answer this question, let's first find the values of $f(x)$ for $x \geq 3$ using the given conditions.

For $x = 3$, we have:
$f(3) = f(2) + f(1) + 1 = 2 + 1 + 1 = 4$

For $x = 4$, we have:
$f(4) = f(3) + f(2) + 1 = 4 + 2 + 1 = 7$

For $x = 5$, we have:
$f(5) = f(4) + f(3) + 1 = 7 + 4 + 1 = 12$

We can see that the function is increasing rapidly. Let's check the options:

Option A: $f(10) > 100$
We can continue calculating the values of $f(x)$ up to $x = 10$:
$f(6) = 23$
$f(7) = 40$
$f(8) = 70$
$f(9) = 123$
$f(10) = 213$

So, $f(10) > 100$ is true.

Option B: $f(20) > 1000$
We can continue calculating the values of $f(x)$ up to $x = 20$:
$f(11) = 383$
$f(12) = 676$
$f(13) = 1159$
$f(14) = 2035$
$f(15) = 3694$
$f(16) = 6430$
$f(17) = 11125$
$f(18) = 19555$
$f(19) = 34680$
$f(20) = 61235$

So, $f(20) > 1000$ is also true.

Option C: $f(10) < 1000$
We already calculated $f(10) = 213$, so this is true.

Option D: $f(20) < 10000$
We already calculated $f(20) = 61235$, so this is false.

Among the given options, the correct one is:

Option C: $f(10) < 1000$
\end{lstlisting}
} \\

\bottomrule
\end{longtable}

\clearpage
\subsection{Phi-3.5-MoE Response to Question 9}
\label{phi-r9}
%%%%%%%%%%%%%%%%%%%%%%%%%%%%%%%%%

\vspace{-3cm}
\hspace{4.5cm}
\hyperlink{part:appendix2}{\color{blue}{Hyperlink to Table of Content}}
\vspace{2.5cm}

\begin{longtable}{ p{.4\textwidth} | p{.6\textwidth} }

\toprule
Question 9 (Correct Answer: BC) & Output 9 \\
\midrule
{
\begin{lstlisting}[mathescape,
basicstyle=\small\ttfamily,
columns=flexible,
breaklines=true,
escapeinside={\%*}{*)},
]
Please complete the following multiple-choice question. The question has four options, and at least one of them is correct. Select all options that are correct. 

Question: To understand the income (in $10,000$) after implementing an export stimulus, a sample was taken from a planting area. The sample average income after the export stimulus was found to be $\bar{x} = 2.1$ with a sample variance of $s^2 = 0.01$. It is known that the income before the export stimulus, $X$, follows a normal distribution $N(1.8, 0.1^2)$. Assuming the income after the export stimulus, $Y$, follows a normal distribution $N(\bar{x}, s^2)$, and given that when a random variable $Z$ follows a normal distribution $N(\mu, \sigma^2)$, $P(Z < \mu + \sigma) \approx 0.8413$, which of the following is correct?

Option A: $P(X > 2) > 0.2$
Option B: $P(X > 2) < 0.5$
Option C: $P(Y > 2) > 0.5$
Option D: $P(Y > 2) < 0.8$
\end{lstlisting}
}  & 
{
\begin{lstlisting}[mathescape,
basicstyle=\small\ttfamily,
columns=flexible,
breaklines=true,
escapeinside={\%*}{*)},
]
 Answer: 
Option A: $P(X > 2) > 0.2$
Option C: $P(Y > 2) > 0.5$

Explanation:

Option A: To find $P(X > 2)$, we need to standardize the value and find the corresponding probability. The standardized value is given by $Z = \frac{X - \mu}{\sigma} = \frac{2 - 1.8}{0.1} = 2$. Using the standard normal distribution table or calculator, we find that $P(Z > 2) \approx 0.0228$. Since $P(X > 2) = P(Z > 2)$, we have $P(X > 2) \approx 0.0228$, which is less than 0.2. Therefore, option A is incorrect.

Option B: Since we have already calculated $P(X > 2) \approx 0.0228$, which is less than 0.5, option B is correct.

Option C: To find $P(Y > 2)$, we need to standardize the value and find the corresponding probability. The standardized value is given by $Z = \frac{Y - \bar{x}}{s} = \frac{2 - 2.1}{\sqrt{0.01}} = -1$. Using the standard normal distribution table or calculator, we find that $P(Z > -1) \approx 0.8413$. Since $P(Y > 2) = P(Z > -1)$, we have $P(Y > 2) \approx 0.8413$, which is greater than 0.5. Therefore, option C is correct.

Option D: Since we have already calculated $P(Y > 2) \approx 0.8413$, which is greater than 0.8, option D is incorrect.

So, the correct options are B and C.
\end{lstlisting}
} \\

\bottomrule
\end{longtable}

\clearpage
\subsection{Phi-3.5-MoE Response to Question 10}
\label{phi-r10}
%%%%%%%%%%%%%%%%%%%%%%%%%%%%%%%%%

\vspace{-3cm}
\hspace{4.5cm}
\hyperlink{part:appendix2}{\color{blue}{Hyperlink to Table of Content}}
\vspace{2.5cm}

\begin{longtable}{ p{.4\textwidth} | p{.6\textwidth} }

\toprule
Question 10 (Correct Answer: ACD) & Output 10 \\
\midrule
{
\begin{lstlisting}[mathescape,
basicstyle=\small\ttfamily,
columns=flexible,
breaklines=true,
escapeinside={\%*}{*)},
]
Please complete the following multiple-choice question. The question has four options, and at least one of them is correct. Select all options that are correct.  

Question: Given $f(x) = (x - 1)^2(x - 4)$, then

Option A: $x = 3$ is $f(x)$'s local minimum point
Option B: When $0 < x < 1$, $f(x) < f(x^2)$
Option C: When $1 < x < 2$, $-4 < f(2x - 1) < 0$
Option D: When $-1 < x < 0$, $f(2 - x) > f(x)$
\end{lstlisting}
}  & 
{
\begin{lstlisting}[mathescape,
basicstyle=\tiny\ttfamily,
columns=flexible,
breaklines=true,
escapeinside={\%*}{*)},
]
Answer: B, C, D

Explanation:

First, let's find the critical points of the function $f(x) = (x - 1)^2(x - 4)$. To do this, we need to find the derivative of the function and set it equal to zero.

$f'(x) = 3x^2 - 10x + 7$

Setting $f'(x) = 0$, we get:

$3x^2 - 10x + 7 = 0$

Factoring the quadratic equation, we get:

$(3x - 7)(x - 1) = 0$

So, the critical points are $x = \frac{7}{3}$ and $x = 1$.

Now, let's analyze each option:

Option A: $x = 3$ is $f(x)$'s local minimum point

Since $x = 3$ is not a critical point, it cannot be a local minimum point. Therefore, option A is incorrect.

Option B: When $0 < x < 1$, $f(x) < f(x^2)$

Let's compare $f(x)$ and $f(x^2)$ for $0 < x < 1$:

$f(x) = (x - 1)^2(x - 4)$
$f(x^2) = (x^2 - 1)^2(x^2 - 4)$

Since $0 < x < 1$, we have $x^2 < x$. Therefore, $(x^2 - 1)^2 < (x - 1)^2$ and $(x^2 - 4) < (x - 4)$. Thus, $f(x^2) > f(x)$ for $0 < x < 1$. So, option B is incorrect.

Option C: When $1 < x < 2$, $-4 < f(2x - 1) < 0$

Let's analyze $f(2x - 1)$ for $1 < x < 2$:

$f(2x - 1) = (2x - 2)^2(2x - 5)$

For $1 < x < 2$, we have $0 < 2x - 2 < 2$ and $-3 < 2x - 5 < -1$. Thus, $0 < (2x - 2)^2 < 4$ and $-3 < 2x - 5 < -1$. Therefore, $-12 < (2x - 2)^2(2x - 5) < 0$. So, option C is correct.

Option D: When $-1 < x < 0$, $f(2 - x) > f(x)$

Let's compare $f(2 - x)$ and $f(x)$ for $-1 < x < 0$:

$f(2 - x) = (1 - x)^2(2 - x - 4)$
$f(x) = (x - 1)^2(x - 4)$

Since $-1 < x < 0$, we have $1 - x > 1$ and $2 - x - 4 < -2$. Thus, $(1 - x)^2 > (x - 1)^2$ and $(2 - x - 4) < (x - 4)$. Therefore, $f(2 - x) > f(x)$ for $-1 < x < 0$. So, option D is correct.

In conclusion, the correct options are B, C, and D. However, there was a mistake in the initial answer provided. The correct answer is C and D.
\end{lstlisting}
} \\

\bottomrule
\end{longtable}

\clearpage
\subsection{Phi-3.5-MoE Response to Question 11}
\label{phi-r11}
%%%%%%%%%%%%%%%%%%%%%%%%%%%%%%%%%

\vspace{-3cm}
\hspace{4.5cm}
\hyperlink{part:appendix2}{\color{blue}{Hyperlink to Table of Content}}
\vspace{2.5cm}

\begin{longtable}{ p{.4\textwidth} | p{.6\textwidth} }

\toprule
Question 11 (Correct Answer: ABD) & Output 11 \\
\midrule
{
\begin{lstlisting}[mathescape,
basicstyle=\small\ttfamily,
columns=flexible,
breaklines=true,
escapeinside={\%*}{*)},
]
Please complete the following multiple-choice question. The question has four options, and at least one of them is correct. Select all options that are correct.  

Question: The shape of $\propto$ can be used to represent part of the shape of the curve $C$. Given that $C$ passes through the origin $O$, the x-coordinates of points on $C$ are greater than $-2$, and the product of the distances from the point $O$ to $F(2, 0)$ and from $O$ to the line $x = a$ (where $a < 0$) is equal to 4, then

Option A: $a = -2$
Option B: The point $(2\sqrt{2}, 0)$ lies on $C$
Option C: $C$ in the first quadrant has a vertical asymptote at $x = 1$
Option D: when the point $(x_0, y_0)$ lies on $C$, then $y_0 \leq \frac{4}{x_0 + 2}$
\end{lstlisting}
}  & 
{
\begin{lstlisting}[mathescape,
basicstyle=\scriptsize\ttfamily,
columns=flexible,
breaklines=true,
escapeinside={\%*}{*)},
]
To answer this question, let's analyze each option step by step:

Option A: $a = -2$

The product of the distances from the origin to $F(2, 0)$ and from the origin to the line $x = a$ is equal to 4. The distance from the origin to $F(2, 0)$ is 2 units. Let the distance from the origin to the line $x = a$ be $d$. Then, we have:

$2 \cdot d = 4$

$d = 2$

Since $a < 0$, the distance from the origin to the line $x = a$ is $|a|$. Therefore, $|a| = 2$, and $a = -2$.

Option A is correct.

Option B: The point $(2\sqrt{2}, 0)$ lies on $C$

The distance from the origin to the point $(2\sqrt{2}, 0)$ is $2\sqrt{2}$. Since the x-coordinate of points on $C$ is greater than $-2$, this point is a possible candidate. However, we need to check if the product of the distances from the origin to $F(2, 0)$ and from the origin to the point $(2\sqrt{2}, 0)$ is equal to 4: $2 \cdot 2\sqrt{2} = 4\sqrt{2} \neq 4$

Option B is incorrect.

Option C: $C$ in the first quadrant has a vertical asymptote at $x = 1$

There is no information given in the question that suggests $C$ has a vertical asymptote at $x = 1$. Therefore, we cannot conclude that this statement is true.

Option C is incorrect.

Option D: when the point $(x_0, y_0)$ lies on $C$, then $y_0 \leq \frac{4}{x_0 + 2}$

Let's consider the product of the distances from the origin to $F(2, 0)$ and from the origin to the point $(x_0, y_0)$:

$2 \cdot \sqrt{x_0^2 + y_0^2} = 4$

$\sqrt{x_0^2 + y_0^2} = 2$

$x_0^2 + y_0^2 = 4$

$y_0^2 = 4 - x_0^2$

$y_0 = \sqrt{4 - x_0^2}$

Since $x_0 > -2$, we have $x_0 + 2 > 0$. Therefore, we can divide both sides of the equation by $x_0 + 2$:

$y_0 = \frac{\sqrt{4 - x_0^2}}{x_0 + 2}$

$y_0 \leq \frac{4}{x_0 + 2}$

Option D is correct.

In conclusion, the correct options are A and D.
\end{lstlisting}
} \\

\bottomrule
\end{longtable}

\clearpage
\subsection{Phi-3.5-MoE Response to Question 12}
\label{phi-r12}
%%%%%%%%%%%%%%%%%%%%%%%%%%%%%%%%%

\vspace{-3cm}
\hspace{4.5cm}
\hyperlink{part:appendix2}{\color{blue}{Hyperlink to Table of Content}}
\vspace{2.5cm}

Question 12 (Correct Answer: $3/2$)

\begin{lstlisting}[mathescape,
basicstyle=\small\ttfamily,
columns=flexible,
breaklines=true,
escapeinside={\%*}{*)},
]
Given the hyperbola C: $\frac{x^2}{a^2} - \frac{y^2}{b^2} = 1 (a > 0, b > 0)$ with the left foci $F_1$, and right foci $F_2$. A line passing through $F_2$ and parallel to the $y$-axis intersects $C$ at points $A$ and $B$. If $|F_1A| = 13$ and $|AB| = 10$, find the eccentricity of $C$
\end{lstlisting}

Output 12

\begin{lstlisting}[mathescape,
basicstyle=\small\ttfamily,
columns=flexible,
breaklines=true,
escapeinside={\%*}{*)},
]
 Let's denote the coordinates of $F_1$ as $(-c, 0)$ and $F_2$ as $(c, 0)$, where $c$ is the distance from the center to each focus. Since the line passing through $F_2$ and parallel to the $y$-axis intersects $C$ at points $A$ and $B$, we can write the coordinates of $A$ and $B$ as $(c, y_A)$ and $(c, y_B)$, respectively.

Now, we know that $|F_1A| = 13$. Using the distance formula, we have:

$$\sqrt{(-c - (-c))^2 + (y_A - 0)^2} = 13$$
$$y_A^2 = 13^2$$
$$y_A = \pm 13$$

Since $A$ lies on the upper branch of the hyperbola, we take the positive value $y_A = 13$.

Next, we know that $|AB| = 10$. Since $A$ and $B$ have the same $x$-coordinate, we have:

$$|y_B - y_A| = 10$$
$$|y_B - 13| = 10$$

This gives us two possible values for $y_B$: $y_B = 3$ or $y_B = 23$. However, since $B$ lies on the upper branch of the hyperbola, we take the value $y_B = 3$.

Now, we can use the equation of the hyperbola to find the value of $a^2$ and $b^2$. Plugging in the coordinates of $A$ and $B$, we get:

$$\frac{c^2}{a^2} - \frac{13^2}{b^2} = 1$$
$$\frac{c^2}{a^2} - \frac{3^2}{b^2} = 1$$

Subtracting the second equation from the first, we get:

$$\frac{13^2 - 3^2}{b^2} = 0$$
$$\frac{160}{b^2} = 0$$

This is not possible, so we must have made a mistake. The mistake is that we assumed that $B$ lies on the upper branch of the hyperbola, but it actually lies on the lower branch. Therefore, we should take $y_B = -23$.

Plugging in the correct value of $y_B$, we get:

$$\frac{c^2}{a^2} - \frac{13^2}{b^2} = 1$$
$$\frac{c^2}{a^2} - \frac{(-23)^2}{b^2} = 1$$

Subtracting the first equation from the second, we get:

$$\frac{(-23)^2 - 13^2}{b^2} = 0$$
$$\frac{400}{b^2} = 0$$

This is still not possible, so we must have made another mistake. The mistake is that we assumed that $c^2 = a^2 + b^2$, but this is only true for ellipses, not hyperbolas. For hyperbolas, we have $c^2 = a^2 + b^2$.

Plugging in the correct value of $c^2$, we get:

$$\frac{a^2 + b^2}{a^2} - \frac{13^2}{b^2} = 1$$
$$\frac{a^2 + b^2}{a^2} - \frac{(-23)^2}{b^2} = 1$$

Subtracting the first equation from the second, we get:

$$\frac{(-23)^2 - 13^2}{b^2} = 0$$
$$\frac{400}{b^2} = 0$$

This is still not possible, so we must have made yet another mistake. The mistake is that we assumed that $a^2$ and $b^2$ are both positive, but this is not necessarily true. For hyperbolas, $a^2$ and $b^2$ can be either positive or negative.

Plugging in the correct values of $a^2$ and $b^2$, we get:

$$\frac{a^2 + b^2}{a^2} - \frac{13^2}{b^2} = 1$$
$$\frac{a^2 + b^2}{a^2} - \frac{(-23)^2}{b^2} = 1$$

Subtracting the first equation from the second, we get:

$$\frac{(-23)^2 - 13^2}{b^2} = 0$$
$$\frac{400}{b^2} = 0$$

This is still not possible, so we must have made yet another mistake. The mistake is that we assumed that $a^2$ and $b^2$ are both nonzero, but this is not necessarily true. For hyperbolas, $a^2$ and $b^2$ can be either nonzero or zero.

Plugging in the correct values of $a^2$ and $b^2$, we get:

$$\frac{a^2 + b^2}{a^2} - \frac{13^2}{b^2} = 1$$
$$\frac{a^2 + b^2}{a^2} - \frac{(-23)^2}{b^2} = 1$$

Subtracting the first equation from the second, we get:

$$\frac{(-23)^2 - 13^2}{b^2} = 0$$
$$\frac{400}{b^2} = 0$$

This is still not possible, so we must have made yet another mistake. The mistake is that we assumed that $a^2$ and $b^2$ are both real numbers, but this is not necessarily true. For hyperbolas, $a^2$ and $b^2$ can be either real or complex numbers.

Plugging in the correct values of $a^2$ and $b^2$, we get:

$$\frac{a^2 + b^2}{a^2} - \frac{13^2}{b^2} = 1$$
$$\frac{a^2 + b^2}{a^2} - \frac{(-23)^2}{b^2} = 1$$

Subtracting the first equation from the second, we get:

$$\frac{(-23)^2 - 13^2}{b^2} = 0$$
$$\frac{400}{b^2} = 0$$

... [repetitive outputs ignored]

\end{lstlisting}

\clearpage
\subsection{Phi-3.5-MoE Response to Question 13}
\label{phi-r13}
%%%%%%%%%%%%%%%%%%%%%%%%%%%%%%%%%

\vspace{-3cm}
\hspace{4.5cm}
\hyperlink{part:appendix2}{\color{blue}{Hyperlink to Table of Content}}
\vspace{2.5cm}

\begin{longtable}{ p{.4\textwidth} | p{.6\textwidth} }

\toprule
Question 13 (Correct Answer: $\ln 2$) & Output 13 \\
\midrule
{
\begin{lstlisting}[mathescape,
basicstyle=\small\ttfamily,
columns=flexible,
breaklines=true,
escapeinside={\%*}{*)},
]
If the tangent line to the curve $y = e^x + x$ at the point $(0, 1)$ is also the tangent line to the curve $y = \ln(x+1) + a$, find the value of the constant $a$
\end{lstlisting}
}  & 
{
\begin{lstlisting}[mathescape,
basicstyle=\small\ttfamily,
columns=flexible,
breaklines=true,
escapeinside={\%*}{*)},
]
 To find the value of $a$, we first need to find the equation of the tangent line to the curve $y = e^x + x$ at the point $(0, 1)$.

1. Find the derivative of $y = e^x + x$:
$y' = e^x + 1$

2. Evaluate the derivative at the point $(0, 1)$:
$y'(0) = e^0 + 1 = 2$

3. Use the point-slope form of a line to find the equation of the tangent line:
$y - 1 = 2(x - 0)$
$y = 2x + 1$

Now, we need to find the point of tangency between the curve $y = ln(x+1) + a$ and the line $y = 2x + 1$.

1. Set the two equations equal to each other:
$ln(x+1) + a = 2x + 1$

2. Solve for $x$:
$ln(x+1) = 2x - a + 1$

3. Since the point of tangency is on both curves, we can substitute $x = 0$ into the equation:
$ln(0+1) = 2(0) - a + 1$
$0 = -a + 1$

4. Solve for $a$:
$a = 1$

Therefore, the value of the constant $a$ is 1.
\end{lstlisting}
} \\
\bottomrule
\end{longtable}

% \begin{longtable}{ p{.4\textwidth} | p{.6\textwidth} }

% \toprule
% Question 13 (Correct Answer: $\ln 2$) & Output 13 \\
% \midrule
% {
% \begin{lstlisting}[mathescape,
% basicstyle=\small\ttfamily,
% columns=flexible,
% breaklines=true,
% escapeinside={\%*}{*)},
% ]
% If the tangent line to the curve $y = e^x + x$ at the point $(0, 1)$ is also the tangent line to the curve $y = \ln(x+1) + a$, find the value of the constant $a$
% \end{lstlisting}
% }  & 
% {
% \begin{lstlisting}[mathescape,
% basicstyle=\small\ttfamily,
% columns=flexible,
% breaklines=true,
% escapeinside={\%*}{*)},
% ]
% To find the tangent line to the curve $y=e^x+x$ at the point $(0,1)$, we first need to find the derivative of the function.

% The derivative of $y=e^x+x$ is $y'=e^x+1$. Now, we can find the slope of the tangent line at the point $(0, 1)$ by plugging in $x=0$ into the derivative:

% $y'(0)=e^0+1=1+1=2$.

% So, the slope of the tangent line to the curve $y=e^x+x$  at the point (0,1) is 2. Now, we can write the equation of the tangent line using the point-slope form:

% $y-1=2(x-0)$,
% which simplifies to
% $y=2x+1$.

% Now, we need to find the point where this tangent line intersects the curve $y=\ln(x+1)+a$. To do this, we can set the two equations equal to each other:

% $2x+1=\ln(x+1)+a$.

% %*\color{red}{Since the tangent line intersects the curve at the point (0,1)}*), we can plug in $x=0$ and $y=1$ to find the value of 

% $1=ln(0+1)+a$,

% which simplifies to

% $1=ln(1)+a$.

% Since $ln(1)=0$, we have $1=0+a$,

% so $a=1$.
% \end{lstlisting}
% } \\
% \bottomrule
% \label{tbl:Q13}
% \end{longtable}

\clearpage
\subsection{Phi-3.5-MoE Response to Question 14}
\label{phi-r14}
%%%%%%%%%%%%%%%%%%%%%%%%%%%%%%%%%

\vspace{-3cm}
\hspace{4.5cm}
\hyperlink{part:appendix2}{\color{blue}{Hyperlink to Table of Content}}
\vspace{2.5cm}

Question 14 (Correct Answer: $1/2$) 

\begin{lstlisting}[
basicstyle=\footnotesize\ttfamily,
columns=flexible,
breaklines=true,
escapeinside={\%*}{*)},
]
A and B each have four cards. Each card has a number on it. A's cards are labeled 1, 3, 5, 7, and B's cards are labeled 2, 4, 6, 8. The two players play four rounds. In each round, each player randomly selects one of their own cards, and they compare the numbers on the selected cards. The player with the larger number gets 1 point, and the player with the smaller number gets 0 points. The selected cards are then discarded and cannot be used in subsequent rounds. After four rounds, find the probability that A's total score is at least 2
\end{lstlisting}

Output 14 

\begin{lstlisting}[mathescape,
basicstyle=\scriptsize\ttfamily,
columns=flexible,
breaklines=true,
escapeinside={\%*}{*)},
]
 Let's think step by step.To solve this problem, we need to calculate the probability that player A scores at least 2 points in the four rounds. Since there are 4 rounds, the possible scores for A that are at least 2 are 2, 3, or 4 points. We will calculate the probability for each of these scenarios and then sum them up to get the final probability.

First, let's consider the total number of possible outcomes for the four rounds. Since A has 4 cards and selects one card per round, there are 4 choices for each round. Therefore, there are \(4^4 = 256\) possible outcomes for A's card selections over the four rounds.

Now let's calculate the probabilities for each scenario where A scores at least 2 points:

1. A scores exactly 2 points:
   - A can score 2 points by winning 2 rounds and losing 2 rounds.
   - The number of ways A can win 2 rounds out of 4 is given by the binomial coefficient \( \binom{4}{2} = 6 \).
   - For each of these ways, there are \( \binom{4}{2} = 6 \) ways B can win the other 2 rounds.
   - Therefore, there are \( 6 \times 6 = 36 \) ways for A to score exactly 2 points.

2. A scores exactly 3 points:
   - A can score 3 points by winning 3 rounds and losing 1 round.
   - The number of ways A can win 3 rounds out of 4 is given by the binomial coefficient \( \binom{4}{3} = 4 \).
   - For each of these ways, there are 4 ways B can win the remaining round.
   - Therefore, there are \( 4 \times 4 = 16 \) ways for A to score exactly 3 points.

3. A scores all 4 points:
   - A can score all 4 points by winning all 4 rounds.
   - There is only 1 way for A to win all 4 rounds.
   - There are no ways for B to win any rounds in this scenario.
   - Therefore, there is 1 way for A to score all 4 points.

Now, we sum up the number of ways A can score at least 2 points:
- Ways to score exactly 2 points: 36
- Ways to score exactly 3 points: 16
- Ways to score all 4 points: 1
- Total ways to score at least 2 points: \( 36 + 16 + 1 = 53 \)

Finally, we calculate the probability that A scores at least 2 points by dividing the number of ways A can score at least 2 points by the total number of possible outcomes:
- Probability = \( \frac{53}{256} \)

Therefore, the probability that A's total score is at least 2 is \( \frac{53}{256} \).

The final answer is \(\boxed{\frac{53}{256}}\).

The answer is: \frac{53}{256}.
\end{lstlisting}

\clearpage
%%%%%%%%%%%%%%%%%%%%%%%%%%%%%%%%%
\subsection{Phi-3.5-MoE Response to Question 13 with Hint}
\label{phi-r13p}

\vspace{-3cm}
\hspace{4.5cm}
\hyperlink{part:appendix2}{\color{blue}{Hyperlink to Table of Content}}
\vspace{2.5cm}

\begin{longtable}{ p{.4\textwidth} | p{.6\textwidth} }

\toprule
Question 13 (Correct Answer: $\ln 2$) & Output 13 \\
\midrule
{
\begin{lstlisting}[mathescape,
basicstyle=\small\ttfamily,
columns=flexible,
breaklines=true,
escapeinside={\%*}{*)},
]
If the tangent line to the curve $y = e^x + x$ at the point $(0, 1)$ is also the tangent line to the curve $y = \ln(x+1) + a$ %*\color{red}{at some point}*), find the value of the constant $a$.
\end{lstlisting}
}  & 
{
\begin{lstlisting}[mathescape,
basicstyle=\small\ttfamily,
columns=flexible,
breaklines=true,
escapeinside={\%*}{*)},
]
 To find the value of $a$, we need to determine the equation of the tangent line to both curves at the given points.

For the curve $y = e^x + x$, the derivative is $y' = e^x + 1$. At the point $(0, 1)$, the slope of the tangent line is $y'(0) = e^0 + 1 = 2$. Thus, the equation of the tangent line is $y = 2x + 1$.

For the curve $y = \ln(x+1) + a$, the derivative is $y' = \frac{1}{x+1}$. To find the point where the tangent line is the same as the one for the first curve, we set the slopes equal:

$$\frac{1}{x+1} = 2$$

Solving for $x$, we get $x = \frac{1}{2}$. Now, we can find the corresponding $y$-coordinate on the second curve:

$$y = \ln\left(\frac{1}{2} + 1\right) + a = \ln\left(\frac{3}{2}\right) + a$$

Since the tangent line is the same, the $y$-coordinate at this point must also be $1$:

$$1 = \ln\left(\frac{3}{2}\right) + a$$

Solving for $a$, we get:

$$a = 1 - \ln\left(\frac{3}{2}\right)$$

Therefore, the value of the constant $a$ is $1 - \ln\left(\frac{3}{2}\right)$.
\end{lstlisting}
} \\

\bottomrule
\end{longtable}

\end{document}